\documentclass{article}
\PassOptionsToPackage{numbers,sort&compress}{natbib}
\usepackage[final]{neurips_2025}

\usepackage[utf8]{inputenc} 
\usepackage[T1]{fontenc}    
\usepackage[colorlinks=true,citecolor=blue,linkcolor=black]{hyperref}
\usepackage{url}            
\usepackage{booktabs}       
\usepackage{amsfonts}       
\usepackage{nicefrac}       
\usepackage{microtype}      
\usepackage{xcolor}         

\usepackage{amsmath}
\usepackage{mathtools, nccmath}
\usepackage{array}
\usepackage{siunitx}
\usepackage{multirow}
\usepackage{makecell}  
\usepackage{subcaption}
\usepackage{xcolor}
\usepackage{colortbl}
\usepackage{tcolorbox}
\usepackage{enumitem}
\usepackage{CJKutf8}
\usepackage{moresize}
\usepackage{setspace}
\definecolor{noisyblue}{RGB}{0, 102, 153}
\definecolor{naturalgreen}{RGB}{75, 107, 75}
\usepackage{tikz}
\usepackage[acronym]{glossaries}
\newacronym{qat}{QAT}{Quantization Aware Training}
\newacronym{ptq}{PTQ}{Post-Training Quantization}
\newacronym{hwa}{HWA}{Hardware Aware}
\newacronym{ai}{AI}{Artificial Intelligence}
\newacronym{flop}{FLOP}{Floating Point Operation}
\newacronym{top}{TOP}{Tera Operation}
\newacronym{llm}{LLM}{Large Language Model}
\newacronym{nlp}{NLP}{Natural Language Processing}
\newacronym{cim}{CiM}{Compute-in-Memory}
\newacronym{aimc}{AIMC}{Analog in-Memory Computing}
\newacronym{imc}{IMC}{In-Memory Computing}
\newacronym{dimc}{DIMC}{Digital in-Memory Computing}
\newacronym{dpu}{DPU}{Digital Processing Unit}
\newacronym{hbm}{HBM}{High Bandwidth Memory}
\newacronym{mha}{MHA}{Multi Head Attention}
\newacronym{mvm}{MVM}{Matrix-Vector Multiplication}
\newacronym{ir}{IR}{Intermediate Representation}
\newacronym{ott}{OTT}{One-Tier-at-a-Time}
\newacronym{moe}{MoE}{Mixture of Experts}
\newacronym{mlp}{MLP}{Multi-Layer Perceptron}
\newacronym{pcm}{PCM}{Phase Change Memory}
\newacronym{reram}{ReRAM}{Resisitive RAM}
\newacronym{adc}{ADC}{Analog to Digital Converter}
\newacronym{dac}{DAC}{Digital to Analog Converter}
\newacronym{mfu}{MFU}{Model FLOPs Utilization}
\newacronym{nvm}{NVM}{Non-Volatile Memory}
\newacronym{beol}{BEOL}{Back End of Line}
\newacronym{pwm}{PWM}{Pulse Width Modulation}
\newacronym{snr}{SNR}{Signal-to-Noise Ratio}
\newacronym{ssm}{SSM}{State-Space Model}
\newacronym{simd}{SIMD}{Single Instruction Multiple Data}
\newacronym{rtn}{RTN}{Round-to-Nearest}

\definecolor{shotcolor100}{RGB}{130,207,255}
\definecolor{promptcolor100}{RGB}{255,126,182}
\colorlet{shotcolor}{shotcolor100!80}
\colorlet{promptcolor}{promptcolor100!80}

\newtcolorbox{shotbox}[1][Few shot examples]{
  colback=shotcolor,
  colframe=shotcolor,
  arc=3mm,
  width=\textwidth,
  boxrule=0.5pt,
  title=#1,
  fonttitle=\sffamily,
  fontupper=\sffamily 
}

\newtcolorbox{finalbox}[1][Prompt example]{
  colback=promptcolor,
  colframe=promptcolor,
  arc=3mm,
  width=\textwidth,
  boxrule=0.5pt,
  title=#1,
  fonttitle=\sffamily,
  fontupper=\sffamily 
}

\usepackage{listings}
\newcommand{\pycode}[1]{\lstinline[basicstyle=\ttfamily\color{black}]|#1|}
\DeclarePairedDelimiter{\nint}\lfloor\rceil

\title{Analog Foundation Models}

\author{%
    Julian Büchel$^{1,2}$, Iason Chalas$^{1,2}$, Giovanni Acampa$^{1,2}$, An Chen$^{3}$ \\
    \textbf{Omobayode Fagbohungbe}$^{4}$, \textbf{Sidney Tsai}$^{3}$, \textbf{Kaoutar El Maghraoui}$^{4}$ \\
    \textbf{Manuel Le Gallo}$^{1}$, \textbf{Abbas Rahimi}$^{1}$, \textbf{Abu Sebastian}$^{1}$ \vspace{5pt} \\
    $^{1}$IBM Research – Zurich, $^{2}$ETH Zürich \\
    $^{3}$IBM Research – Almaden, $^{4}$IBM Thomas J. Watson Research Center \vspace{5pt} \\
    \small \texttt{\{jub,anu,abr,ase\}@zurich.ibm.com} \\
    \small \texttt{\{iason.chalas,giovanni.acampa,omobayode.fagbohungbe\}@ibm.com} \\
    \small \texttt{\{chenan,htsai,kelmaghr\}@us.ibm.com}
}

\begin{document}
\maketitle
\addtocontents{toc}{\protect\iffalse}

\section*{Abstract}
Analog in-memory computing (AIMC) is a promising compute paradigm to improve speed and power efficiency of neural network inference beyond the limits of conventional von Neumann-based architectures. However, AIMC introduces fundamental challenges such as noisy computations and strict constraints on input and output quantization. Because of these constraints and imprecisions, off-the-shelf LLMs are not able to achieve 4-bit-level performance when deployed on AIMC-based hardware. While researchers previously investigated recovering this accuracy gap on small, mostly vision-based models, a generic method applicable to LLMs pre-trained on trillions of tokens does not yet exist. In this work, we introduce a general and scalable method to robustly adapt LLMs for execution on noisy, low-precision analog hardware. Our approach enables state-of-the-art models — including \textit{Phi-3-mini-4k-instruct} and \textit{Llama-3.2-1B-Instruct} — to retain performance comparable to 4-bit weight, 8-bit activation baselines, despite the presence of analog noise and quantization constraints. Additionally, we show that as a byproduct of our training methodology, analog foundation models can be quantized for inference on low-precision digital hardware. Finally, we show that our models also benefit from test-time compute scaling, showing better scaling behavior than models trained with 4-bit weight and 8-bit static input quantization. Our work bridges the gap between high-capacity LLMs and efficient analog hardware, offering a path toward energy-efficient foundation models. Code is available at
\url{https://github.com/IBM/analog-foundation-models}.

\section{Introduction}
Fueled by constant hardware~\cite{nvidia-hopper,amd-mi325x,tpu} and software~\cite{attention-is-all-you-need,switch-transformer} innovation, today's \glspl{llm} have billions~\cite{llama3}, or even trillions~\cite{llama4} of parameters and support context windows with millions of tokens~\cite{gemini,llama4}. However, this scale comes at the cost of increased energy consumption when training, and especially serving these models. While hardware used for training has been largely dominated by high-throughput GPUs~\cite{nvidia-hopper,amd-mi325x} or TPUs, novel specialized digital chips for inference have recently begun to emerge~\cite{cerebras,groq,north-pole,furiosa}. However, these chips often either do not scale in weight density~\cite{cerebras,groq,north-pole} or remain limited by the von Neumann bottleneck~\cite{furiosa}.

\begin{figure}
    \centering
    \includegraphics{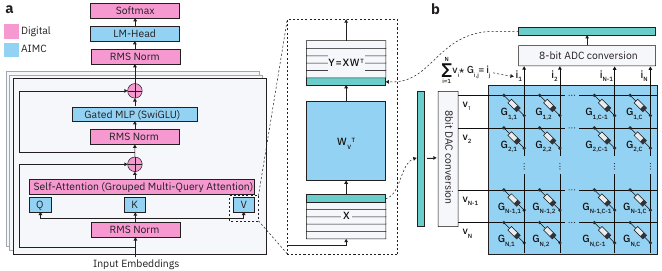}
    \caption{\textbf{a.} Linear layers in the transformer architecture can be offloaded to \gls{aimc} cores. \textbf{b.} In one implementation, given a digital weight matrix $\mathbf{W}$, each \gls{aimc} core calculates $\mathbf{y}=\mathbf{x} \mathbf{W}^T$ in the analog domain using voltages $\mathbf{v}$ and conductances $\mathbf{G}$, where $G_{i,j} \propto W_{i,j}$.}
    \label{fig:figure1}
\end{figure}

A promising alternative to fully-digital architectures is in-memory computing, where computations are performed in-place on stored data. \gls{aimc} addresses both compute efficiency and data movement by performing fully-parallel \glspl{mvm} in the analog domain~\cite{burr-advances,abu-nano} on stored weight matrices, without having to move the weight data to an external processor (see figure \ref{fig:figure1}).
While in-memory computing using SRAM to store weights has recently been shown to achieve impressive throughput and power efficiency~\cite{metis,3nm-sram}, it is fundamentally limited by the bit-cell size of SRAM, which has plateaued over the years~\cite{death-sram}. In order to store larger neural networks fully on-chip, a more scalable memory technology must be used, which is why researchers explored \gls{aimc} with dense \gls{nvm} such as embedded flash~\cite{mythic}, \gls{pcm}~\cite{hermes,ares}, ReRAM~\cite{reram-nature,rram-science-tsmc}, or MRAM~\cite{mram}. To avoid being bound by the planar dimension, researchers have further explored the use of 3D \gls{nvm} technologies such as 3D NAND~\cite{3dnand,3dnand-genome}.
Because the computation is performed in-place within high-density 3D \gls{nvm}, such architectures are suitable for hosting billion- and even trillion-parameter models with a small footprint. This can yield up to three orders of magnitude higher energy efficiency compared to state-of-the-art GPUs when running \gls{moe}-based \glspl{llm}~\cite{buechel-nat-comp}.

Computations performed with \gls{aimc} are significantly faster and more energy efficient, but are imprecise due to various device/circuit nonidealities. These include input and output quantization as a result of digital-to-analog and analog-to-digital conversion, as well as device programming noise~\cite{nandakumar-prog}, read noise~\cite{nandakumar-pcm}, temperature-dependent conductance variations~\cite{boybat-temp-sens}, IR-drop~\cite{accurate-modelling,anchen-ir}, or conductance drift~\cite{zipoli-drift,le-gallo-drift}. Similar to \gls{qat}, methods have been devised to make neural networks more robust to analog noise~\cite{murray-hwa,joshi-hwa,buechel-hwa,rasch-hwa,yu-nature-comm} so that they are amenable for deployment on \gls{aimc} accelerators. The efficacy of these methods has been successfully validated on recently developed \gls{aimc} chips~\cite{hermes,ares,reram-nature} for small CNNs and RNNs with less than 50 million parameters. 
However, the robustness of \glspl{llm} to analog noise remains uncertain~\cite{macronix-iedm}, highlighting the need for innovative approaches to improve their resilience without repeating the entire training pipeline.
In particular, a key challenge is that, unlike quantization, the noise profile of an \gls{aimc} chip is non-deterministic, cannot be efficiently modeled during training, and varies across chips – necessitating a generic training method that can generalize across the diverse noise characteristics of multiple \gls{aimc} devices.

In this paper, we answer the question of whether off-the-shelf \glspl{llm} are robust to noise extracted from existing \gls{aimc} hardware, and we present a detailed data generation and training strategy to turn \glspl{llm} into analog foundation models without access to the original pre-training data. Using scalable \gls{hwa} training techniques, we train two analog foundation models on less than $1\%$ of the total number of pre-training tokens and demonstrate that analog noise induces a drop comparable to 4-bit per-channel weight and 8-bit static input quantization on models trained with state-of-the-art \gls{ptq} and \gls{qat} methods. Our key contributions are as follows:
\begin{itemize}
    \item We demonstrate a detailed and easy-to-follow data generation and training pipeline to turn \glspl{llm} into analog foundation models;
    \item Using these techniques, we train analog versions of \textit{Phi-3-mini-4k-instruct}~\cite{phi3} and \textit{Llama-3.2-1B-Instruct}~\cite{llama3}, and show that when hardware-realistic noise is applied, these models attain accuracy levels comparable to quantized models using 4-bit per-channel weight and 8-bit static input quantization;
    \item We evaluate our analog foundation models on a variety of benchmarks, testing various capabilities such as reasoning, factual knowledge, instruction following, and safety;
    \item As a byproduct of our training pipeline, analog foundation models can also be quantized for inference on low-precision digital hardware, obtaining competitive performance compared to models trained with state-of-the-art quantization algorithms;
    \item Finally, we demonstrate that our models benefit equally from increased test-time compute when compared to quantized models using 4-bit per-channel weight and 8-bit static input quantization, even showing improvements of up to $4.74\%$ on the MATH-500 benchmark.
\end{itemize}

\section{Background and Related Work}
When deploying a model to \gls{aimc} hardware, the weights of the neural network are programmed into \gls{nvm} devices which are arranged in a grid-like structure (see figure \ref{fig:figure1}b).
Then, quantized digital inputs get converted into voltages using \glspl{dac} and are applied to the rows of the grid. Currents representing the dot product between the voltage vector and the weights at each column then flow into \glspl{adc}, where they are finally converted back into the digital domain. The result is then send to either another \gls{aimc} core, a \gls{dpu}, or a RISC-V in a heterogeneous multi-core architecture~\cite{boybat-iedm,jain-tvlsi}.

Three key attributes affect the accuracy of \gls{aimc}: weight noise, input \gls{dac} quantization and output \gls{adc} quantization. Weight noise primarily comes from temporal and device-to-device variations of the \gls{nvm} conductance that encodes a weight \cite{abu-nano} (see figure \ref{fig:figure1}b). Static weight noise (does not get resampled during inference) is mainly due to programming noise, which is the conductance error from the target weight that remains after a device has been programmed with an iterative read-write-verify programming scheme~\cite{nandakumar-prog}. Dynamic weight noise sources whose magnitude vary as a function of time include read noise~\cite{nandakumar-pcm} – due to analog 1/f noise of devices and circuits – and temporal conductance drift, which is the decrease in device conductance over time that is notably observed in \gls{pcm} devices \cite{le-gallo-drift}.
Input \gls{dac}  quantization is equivalent to quantizing the input activations of each layer. However, in contrast to many reduced-precision training papers that assume that the quantization ranges can be dynamically recomputed per-token~\cite{llm-qat,spinquant}, \gls{aimc} typically uses static ranges \cite{rasch-hwa,verma-pmlr}. 
Computing the range of a signal dynamically in hardware is very expensive, and therefore will negatively impact the overall efficiency of an  \gls{aimc} accelerator~\cite{gholami2022survey}. Similarly, \gls{adc} quantization which quantizes the pre-activation outputs of each layer, must also use static ranges. But although \gls{dac} ranges can be configured per-layer, \gls{adc} ranges must be identical across layers, as they reflect the fixed dynamic range and precision of the \glspl{adc}~\cite{verma-pmlr}. 
Although \gls{aimc} hardware advancements can improve these nonidealities, increasing the precision of analog circuits comes at an exponential cost in power and area~\cite{murmann2021}. Therefore, they cannot be completely eliminated and must be mitigated by on-chip calibration methods and by adapting the pre-trained neural network with \gls{hwa} training. On-chip calibration methods typically involve calibrating the \glspl{adc} to reduce non-linearity and \gls{adc}-to-\gls{adc} variability, as well as determining per-output scales and offsets to correct for systematic linear nonidealities in each output channel~\cite{hermes}. However, calibration alone is insufficient to achieve high on-chip accuracy because it cannot compensate for the stochastic nature of programming noise, or temporal variations such as read noise and drift. Therefore, adapting the neural network through \gls{hwa} training is essential for accurate on-chip deployment. Calibration and network adaptation are complementary: calibration reduces deterministic hardware variations, while \gls{hwa} training makes the network robust to the remaining stochastic and nonlinear effects.

Previous works that study \gls{hwa} training for \gls{aimc}-based hardware are limited to CNNs~\cite{rasch-hwa-2019,kariyappa-hwa,joshi-hwa}, RNNs~\cite{rasch-hwa}, LSTMs~\cite{sidney-hwa,rasch-hwa,kariyappa-hwa}, GANs~\cite{yang-hwa} and small encoder-only transformers~\cite{rasch-hwa,spoon-hwa}. Consequently, the studied tasks are easy and the number of training samples small. Additionally, unlike in modern \glspl{llm}, one always has full access to the training data and can easily repeat the full training on a single GPU, with the exception of the encoder-only transformers, where \gls{hwa} training is employed at the finetuning stage. As we show in appendix \ref{appendix:roberta}, performing \gls{hwa} training only during the finetuning stage leads to suboptimal results, and does not necessarily apply to \glspl{llm} as they do not require task-specific finetuning.

In \citet{verma-pmlr}, the authors mainly focus on the accuracy impact of \glspl{adc} with a fixed range and resolution. This static output quantization leads to accuracy degradations, which the authors address by reshaping the activation- and weight-distribution, and by using low-rank adaptation to increase the model's robustness to output quantization. However, additional nonidealities such as weight noise are neglected, and \glspl{llm} only up to 1.7B parameters are investigated solely on the WikiText benchmark. 
In contrast, our work globally focuses on the robustness to weight noise, input, and output quantization of pre-trained \glspl{llm} with more than 3 billion parameters, evaluated on various standard \gls{llm} benchmarks. In fact, we show that – instead of using involved methods such as the ones presented in \citet{verma-pmlr} – training with simple straight-through estimation suffices to tolerate globally static output quantization with minimal loss in accuracy.

\gls{qat} can be used to enhance the robustness of neural networks to quantization errors by training with the quantization operators in the forward pass. In the backward pass, these operators are ignored and the gradients simply passed through~\cite{straight-through}. In the context of \glspl{llm}, a particular challenge that arises when applying \gls{qat} is the unavailability of the training data. This particular problem is solved by LLM-QAT~\cite{llm-qat}, where the authors propose to use the \gls{llm} itself to sample training data, which is then used to train the quantized model via distillation from the original model. As we show in our results, \gls{qat} does not yield satisfiable robustness to noise commonly found in \gls{aimc} hardware. This is due to the fact that by quantizing the weights in the forward pass, the model overfits to the deterministic quantization noise and does not generalize to other types of noise.

While \gls{qat} allows for more harsh modifications to the forward pass, it has the downside of requiring resources and infrastructure for training \glspl{llm}. \gls{ptq} methods solve this by often relying on only local optimizations~\cite{gptq,awq} and other cheap modifications to the network architecture~\cite{quarot,spinquant}. For example, SpinQuant~\cite{spinquant} learns optimal rotations that are applied to activation vectors in order to remove outliers~\cite{quarot} and uses GPT-Q~\cite{gptq} for the weight quantization. Similar to \gls{qat}, we show that \gls{ptq} methods do not yield any robustness to noise found in \gls{aimc} hardware. Furthermore, we find that \gls{ptq} methods tend to degrade accuracy when static ranges for activation quantization are calibrated in a post-training method. This further limits the applicability of \gls{ptq} methods, as dynamic per-token quantization introduces non-negligible computational overhead in dedicated accelerators.

\section{Methods}
In our experiments, we make the following hardware assumptions. We assume a heterogeneous accelerator that is capable of executing \glspl{mvm} with static weights in analog. Other digital operations such as activation functions and the attention computation are executed in FP16 using dedicated digital units. Activations streaming into an analog core are quantized to 8 bits using learnable input ranges that are fixed during inference. Similar to \citet{verma-pmlr}, we assume output ranges for 8-bit output quantization that are identical across layers and fixed during training and inference. In all cases, we employ symmetric quantization. When weight quantization is employed, we use per-channel weight quantization. We keep the KV-cache in FP16.

To ease comparison, we use the following notation for describing our setups. We use \textbf{SI} for static input quantization and \textbf{DI} for dynamic input quantization. When present, \textbf{O} indicates that globally static output quantization is used. When noise is applied to weights, $\textbf{W}_{\text{<type>}}$ is used where <type> indicates the type of noise used. For example, a configuration labeled \textbf{SI8–$\text{W4}_{\text{hw noise}}$–O8} represents a configuration with static 8-bit input quantization, 4-bit per-channel weight quantization with hardware-realistic noise added after quantization, and 8-bit globally static output quantization.

\subsection{Training analog foundation models}
\begin{figure}[ht]
    \centering
    \includegraphics[width=390pt]{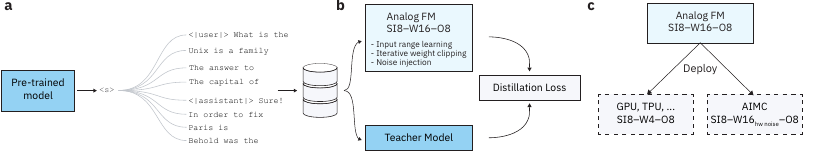}
    \caption{\textbf{a.} Synthetic data is generated by repeated sampling from the softmax distribution. \textbf{b.} Analog foundation models are trained via knowledge distillation on the synthetic data. \textbf{c.} The trained models can then be deployed on analog or low-precision digital hardware.}
    \label{fig:figure2}
\end{figure}

\paragraph{Methodology} We follow the training pipeline from \citet{llm-qat}, which is depicted in figure \ref{fig:figure2}. First, the pre-trained model is used to generate data by iteratively sampling from the softmax distribution of the model given the tokens generated so far, starting with the BOS token. Using distillation, we then train the analog foundation model using \gls{hwa} training. Finally, the trained model can be deployed on analog or low precision digital hardware. The process from data generation to deployment is illustrated in more detail in figure \ref{fig:eval-pipeline}. Unlike \citet{llm-qat}, we do not find that greedy sampling of the first 3-5 tokens improves downstream performance. We also find that, while synthetic data performs the best, open-source datasets such as FineWeb~\cite{fineweb} also yield good results as long as distillation is used. The benefit of using distillation in the context of \gls{hwa} training has also been shown in \citet{chuteng}. We also observe strong gains by scaling the number of training tokens from 1B to 20B, after which we did not see significant improvements. This is significantly more compared to the 100M tokens used in \citet{llm-qat} and we show that this scaling also benefits models trained with LLM-QAT. For details on the ablations performed on the methodology, see appendix \ref{appendix:ablations-methodology}.

\paragraph{Hardware aware training}
We use AIHWKIT-Lightning~\cite{aihwkit-lightning}, an open-source toolkit developed for scalable \gls{hwa} training based on PyTorch~\cite{pytorch}. In our training, we employed four key \gls{hwa} training features, which we describe in more detail. \\
In \gls{aimc} cores, inputs are converted to analog values using \glspl{dac}. On the one hand, higher input precision eases network training and yields higher computational precision, but on the other hand also incurs longer \gls{mvm} latency. At 8 bits, this trade-off is optimal as it still yields low \gls{mvm} latency, and does not impact model accuracy~\cite{spinquant}. We model input quantization of the inputs $\mathbf{x}$ according to eq. \ref{eq:input-quant}, where $\beta^\text{inp. quant}$ are learnable input ranges, "input bits" are the number of bits for quantization, and $\nint{.}$ is the round-to-nearest operator. For the first $500$ forward passes during training, the input ranges are initialized using an exponential moving average over $\kappa \cdot \text{\pycode{std}}(\mathbf{x})$, where $\kappa$ is $15.0-18.0$, as we observed that \textbf{any} activation clipping in the beginning of training hindered convergence. After initialization, the input ranges are updated with a custom gradient~\cite{aihwkit-lightning} that favors tight input ranges to reduce the quantization error.

\begin{align} \label{eq:input-quant}
    & \mathbf{x^\text{quant}} \leftarrow \frac{\beta^\text{inp. quant}}{2^{\text{input bits}-1}-1} \cdot \nint{ \text{\pycode{clamp}}(\mathbf{x},-\beta^\text{inp. quant},\beta^\text{inp. quant}) \cdot \frac{2^{\text{input bits}-1}-1}{\beta^\text{inp. quant}}}
\end{align}

Similar to \citet{verma-pmlr}, our assumed tile architecture comprises \glspl{adc} that convert analog signals to digital ones. We assume that these \glspl{adc} are not configurable per layer and can therefore be modeled as output quantization with globally fixed output ranges $\lambda_\text{adc}$, as shown in eq. \ref{eq:out-quant}. In contrast to \citet{verma-pmlr} where perplexity on WikiText-2 increases by more than 400 when simple \gls{qat} is used, we show that we only lose between $0.2\%$ and $0.5\%$ average accuracy when training with globally static 8-bit output quantization using simple straight-through estimation~\cite{straight-through}. For more details, see appendix \ref{appendix:ablations-hw-aware}.

\begin{equation} \label{eq:out-quant}
\begin{aligned}
    & \mathbf{y}^\text{quant}_i \leftarrow \text{\pycode{clamp}}(\frac{\beta^\text{adc quant}_i}{2^{\text{adc bits}-1}-1} \cdot \nint{ \mathbf{y}_i \cdot \frac{2^{\text{adc bits}-1}-1}{\beta^\text{adc quant}_i}}, -\beta^\text{adc quant}_i, \beta^\text{adc quant}_i) \\
    & \beta^\text{adc quant}_i = \lambda_\text{adc} \cdot \beta^\text{inp. quant} \cdot \text{\pycode{max}}(|\mathbf{W}_{:,i}|)
\end{aligned}
\end{equation}

In order to enhance robustness to noisy analog computations, we inject per-channel additive Gaussian noise~\cite{joshi-hwa} into the weights during the forward pass (see eq. \ref{eq:weight-noise}). During the backward pass, the noise-free weights are used. In more detailed ablation studies (see appendix \ref{appendix:ablations-hw-aware}), we show that an additive Gaussian noise that only scales the noise with respect to the per-channel maximum of the weights works best, and that a small $\gamma_\text{weight}$ ($0.02-0.03$) optimally balances the trade-off between accuracy and robustness.

\begin{equation} \label{eq:weight-noise}
\begin{aligned}
    & \mathbf{W^\text{noisy}}_{:,i} \leftarrow  \mathbf{W}_{:,i} + \gamma_\text{weight} \cdot \text{\pycode{max}}(|\mathbf{W}_{:,i}|) \cdot \tau \\
    & \text{where} \; \tau \sim \mathcal{N}(\mathbf{0},\mathbf{I})
\end{aligned}
\end{equation}

Finally, when a network layer is deployed to \gls{aimc} hardware, weights are mapped to conductance values in a range that is dictated by the nature of the \gls{nvm} devices in the hardware. To fully exploit the conductance range, we clip the weights after every optimizer step according to eq. \ref{eq:clipping}, where $\alpha$ is a user-configurable hyperparameter that controls the amount of clipping.

\begin{equation} \label{eq:clipping}
    \mathbf{W^*}_{:,i} \leftarrow \text{\pycode{clamp}}(\mathbf{W}_{:,i}, -\mathbf{\zeta_i}, \mathbf{\zeta_i}) \; \text{where} \; \mathbf{\zeta_i} = \alpha \cdot \text{\pycode{std}}(\mathbf{W}_{:,i}) 
\end{equation}

We find that although noise injection has been widely believed to provide the most significant improvements in robustness, we demonstrate in further ablation studies (see appendix \ref{appendix:ablations-hw-aware}) that for \glspl{llm} weight clipping yields stronger robustness, and that a combination of the two performs best. This is because clipping applied during training causes smaller weights to be mapped to larger conductance values, which have higher \gls{snr} compared to the lower conductance values for the \gls{pcm}-based model we consider. As a result, the average per-weight \gls{snr} increases, resulting in higher robustness (see appendix \ref{appendix:snr} for details).

\paragraph{Training setup}
For training our analog foundation models, we use 20B tokens which are synthetically generated using vLLM~\cite{vllm}. After data generation, we train our models on 96 V100 GPUs. Because of the V100's limited DRAM capacity, we use DeepSpeed ZeRO stage 2, which includes gradient- and optimizer state partitioning. We also use activation checkpointing and CPU offloading to further reduce memory consumption. For both models, we train with a maximum sequence length of 4096 which is also the chunk size used during data generation. Training of the \textit{Phi-3-mini-4k-instruct}-based analog foundation model takes about 230h, while training the smaller \textit{Llama-3.2-1B-Instruct}-based models takes about 90h. When using GPUs with more memory, the time and number of required GPUs reduces drastically as training is more efficient. For example, training a \textit{Phi-3-mini-4k-instruct}-based model on 8 A100s takes the same time as training it on 48 V100s.

\subsection{Evaluation}
Thorough evaluation on a wide variety of benchmarks is key to understanding the true performance of our models compared to the original ones. In many related works, we observe that the number of benchmarks used for evaluation is small, or hard benchmarks (especially reasoning-based ones) are omitted~\cite{verma-pmlr,rasch-hwa,llm-qat}, which can lead to misleading results. In this paper, we overcome this limitation by examining our models on a total of 12 benchmarks covering diverse key areas of \gls{llm} capabilities, including problem solving tasks in STEM~\cite{arc,mmlu}, medicine~\cite{med-qa} or multiple professions~\cite{agi-eval}, commonsense reasoning~\cite{hellaswag}, mathematical reasoning~\cite{gsm8k,math500}, trivia~\cite{boolq}, natural language inference~\cite{anli}, instruction following~\cite{ifeval}, and safety~\cite{xstest}. For more information on the exact prompts used for few-shot learning, the way we extracted the answers for every benchmark, the system prompts used, see appendix \ref{appendix:evaluations}. \\
To simulate the performance of our models in a more realistic setting, we used a noise model from a state-of-the-art 64-core \gls{pcm}-based \gls{aimc} chip~\cite{hermes}. In this chip, the noise introduced by imprecise programming of the weights into the \gls{nvm} devices - programming noise -  dominates, which is why we chose to focus on this type of noise. Note that programming noise does not originate directly from device-to-device variability, which is mostly accounted for by the iterative read-write-verify programming scheme that iteratively nudges the device conductance towards a target value. Programming noise is actually the conductance error from the target weight that remains after a device has been programmed. As can be seen in figure \ref{fig:pcm-noise}, the amount of noise depends on the conductance state, with higher conductances having more noise than lower ones. However, because of an additive noise floor, lower conductance states have a worse signal-to-noise ratio than higher conductance states. In this paper, we indicate benchmark results obtained with this realistic noise model with $W_\text{hw noise}$. Because the noise model from \citet{hermes} is specific to one type of \gls{nvm} device, we also evaluate our foundation models on generic additive Gaussian noise (see eq. \ref{eq:weight-noise}). Results obtained with this type of noise are indicated with $W_\text{gaussian noise}$. Unless stated otherwise, experiments involving noise injection during evaluation were repeated 10 times for different random seeds, which we found to be crucial for meaningful comparisons.

\section{Results}

\subsection{Analog foundation models are robust to analog noise}
When evaluating both models – \textit{Phi-3-mini-4k-instruct} and \textit{Llama-3.2-1B-Instruct} – off-the-shelf, we see that only injecting hardware-realistic noise leads to an average drop of $8.01\%$ and $9.81\%$, respectively (see table \ref{tab:model-comparison}). On more challenging benchmarks such as GSM8K, accuracy drops even more: $21.43\%$ for \textit{Phi-3-mini-4k-instruct} and $23.2\%$ for \textit{Llama-3.2-1B-Instruct}. Our analog foundation models improve on this significantly, reducing the gap to off-the-shelf FP16 performance to $3.81\%$ for \textit{Phi-3-mini-4k-instruct} and $4.58\%$ for \textit{Llama-3.2-1B-Instruct} while also including static 8-bit input quantization and globally static 8-bit output quantization (see further results in appendix \ref{appendix:ablations-noise} where we compare to an analog foundation model trained with 7-bit input quantization). Strong gains can be observed especially for harder tasks such as GSM8K, HellaSwag, and ANLI, where the gaps were reduced by up to $12.87\%$.
We find that \gls{qat} also improves robustness to hardware-realistic analog noise, which is due to the indirect noise injection during training via weight-quantization. As table \ref{tab:model-comparison} shows, our analog foundation models still outperform models trained with LLM-QAT significantly, especially on harder tasks with differences of up to $12.87\%$ for \textit{Phi-3-mini-4k-instruct} and $8.16\%$ \textit{Llama-3.2-1B-Instruct}.
We also compare our analog foundation models to models that were quantized post-training with SpinQuant~\cite{spinquant}. 
Because the original implementation of SpinQuant uses dynamic per-token input quantization, we report the model performance with it (DI8), along with our implementation using hardware-friendly static input ranges (SI8) which evidently performs worse.
In any case, we observe that models quantized using SpinQuant~\cite{spinquant} show lower robustness to hardware noise even compared to the original off-the-shelf model.

\begin{table}[ht]
\caption{Comparison of \textit{Phi-3-mini-4k-instruct} and \textit{Llama-3.2-1B-Instruct}-based analog foundation models against off-the-shelf models, models trained with LLM-QAT, and models quantized with SpinQuant. Best results when hardware-realistic noise is applied are bold faced. Evaluations with noise injections are repeated for 10 different seeds per benchmark.}
\centering
\resizebox{\textwidth}{!}{%
\begin{tabular}{@{}l*{10}{c}@{}}
\toprule
\multirow{2}{*}{Model} & \multicolumn{9}{c}{Benchmarks} & \multirow{2}{*}{Avg.} \\
\cmidrule(lr){2-10}
 & \makecell{MMLU\\(5-shot)} & \makecell{GSM8K\\(CoT 8-shot)} & \makecell{BoolQ\\(0-shot)} & \makecell{Hellaswag\\(5-shot)} & \makecell{MedQA\\(2-shot)} & \makecell{AGIEval\\(0-shot)} & \makecell{Arc-C\\(10-shot)} & \makecell{Arc-E\\(10-shot)} & \makecell{ANLI\\(7-shot)} & \\
\midrule
\parbox[t]{3.5cm}{\textit{Phi-3-mini-4k-instruct}\\(W16)} & \multirow{2}{*}{$69.36$} & \multirow{2}{*}{$79.91$} & \multirow{2}{*}{$78.62$} & \multirow{2}{*}{$83.51$} & \multirow{2}{*}{$52.91$} & \multirow{2}{*}{$38.06$} & \multirow{2}{*}{$84.56$} & \multirow{2}{*}{$91.08$} & \multirow{2}{*}{$52.25$} & \multirow{2}{*}{$70.03$} \\
\rowcolor[gray]{0.9} 
\parbox[t]{3.5cm}{\textit{Phi-3-mini-4k-instruct}\\($\text{W}\text{16}_\text{hw noise}$)} & \parbox[t]{1.5cm}{\centering $63.32$\\$\pm 1.56$} & \parbox[t]{1.5cm}{\centering $58.48$\\$\pm 8.10$} & \parbox[t]{1.5cm}{\centering $73.25$\\$\pm 1.71$} & \parbox[t]{1.5cm}{\centering $73.36$\\$\pm 6.63$} & \parbox[t]{1.5cm}{\centering $44.90$\\$\pm 2.39$} & \parbox[t]{1.5cm}{\centering $33.48$\\$\pm 1.34$} & \parbox[t]{1.5cm}{\centering $80.39$\\$\pm 3.06$} & \parbox[t]{1.5cm}{\centering $89.02$\\$\pm 1.05$} & \parbox[t]{1.5cm}{\centering $42.00$\\$\pm 3.89$} & \multirow{2}{*}{$62.02$} \\
\parbox[t]{3.5cm}{Analog FM\\(SI8-W16-O8)} & \multirow{2}{*}{$67.24$} & \multirow{2}{*}{$76.19$} & \multirow{2}{*}{$77.00$} & \multirow{2}{*}{$82.68$} & \multirow{2}{*}{$48.66$} & \multirow{2}{*}{$37.00$} & \multirow{2}{*}{$84.04$} & \multirow{2}{*}{$90.74$} & \multirow{2}{*}{$52.38$} & \multirow{2}{*}{$68.44$} \\
\rowcolor[gray]{0.9}
\parbox[t]{3.5cm}{Analog FM\\(SI8-$\text{W16}_\text{hw noise}$-O8)} & \parbox[t]{1.5cm}{\centering $\mathbf{65.11}$\\$\pm \mathbf{0.32}$} & \parbox[t]{1.5cm}{\centering $\mathbf{71.35}$\\$\pm \mathbf{0.97}$} & \parbox[t]{1.5cm}{\centering $\mathbf{75.34}$\\$\pm \mathbf{1.60}$} & \parbox[t]{1.5cm}{\centering $\mathbf{80.56}$\\$\pm \mathbf{0.74}$} & \parbox[t]{1.5cm}{\centering $\mathbf{46.43}$\\$\pm \mathbf{0.70}$} & \parbox[t]{1.5cm}{\centering $\mathbf{35.57}$\\$\pm \mathbf{0.71}$} & \parbox[t]{1.5cm}{\centering $\mathbf{83.15}$\\$\pm \mathbf{0.54}$} & \parbox[t]{1.5cm}{\centering $\mathbf{89.83}$\\$\pm \mathbf{0.28}$} & \parbox[t]{1.5cm}{\centering $\mathbf{48.63}$\\$\pm \mathbf{2.83}$} & \multirow{2}{*}{$\mathbf{66.22}$} \\
\parbox[t]{3.5cm}{LLM-QAT (SI8-W4)} & $64.12$ & $69.90$ & $75.50$ & $79.18$ & $44.03$ & $36.19$ & $81.57$ & $89.31$ & $51.50$ & $65.70$ \\
\rowcolor[gray]{0.9}
\parbox[t]{3.5cm}{LLM-QAT\\(SI8-$\text{W4}_\text{hw noise}$)} & \parbox[t]{1.5cm}{\centering $60.05$\\$\pm 1.27$} & \parbox[t]{1.5cm}{\centering $60.10$\\$\pm 1.40$} & \parbox[t]{1.5cm}{\centering $68.82$\\$\pm 8.66$} & \parbox[t]{1.5cm}{\centering $74.77$\\$\pm 1.15$} & \parbox[t]{1.5cm}{\centering $40.03$\\$\pm 1.35$} & \parbox[t]{1.5cm}{\centering $33.77$\\$\pm 1.15$} & \parbox[t]{1.5cm}{\centering $77.88$\\$\pm 1.51$} & \parbox[t]{1.5cm}{\centering $87.23$\\$\pm 0.89$} & \parbox[t]{1.5cm}{\centering $42.59$\\$\pm 5.52$} & \multirow{2}{*}{$60.58$} \\
\parbox[t]{3.5cm}{SpinQuant (SI8-W4)} & $62.66$ & $68.46$ & $72.51$ & $72.43$ & $44.50$ & $34.51$ & $80.12$ & $89.10$ & $45.44$ & $63.30$ \\
\rowcolor[gray]{0.9}
\parbox[t]{3.5cm}{SpinQuant\\(SI8-$\text{W4}_\text{hw noise}$)} & \parbox[t]{1.5cm}{\centering $29.67$\\$\pm 2.26$} & \parbox[t]{1.5cm}{\centering $2.85$\\$\pm 1.71$} & \parbox[t]{1.5cm}{\centering $57.29$\\$\pm 4.02$} & \parbox[t]{1.5cm}{\centering $25.08$\\$\pm 0.67$} & \parbox[t]{1.5cm}{\centering $23.20$\\$\pm 1.72$} & \parbox[t]{1.5cm}{\centering $24.07$\\$\pm 0.99$} & \parbox[t]{1.5cm}{\centering $29.61$\\$\pm 3.52$} & \parbox[t]{1.5cm}{\centering $36.03$\\$\pm 8.53$} & \parbox[t]{1.5cm}{\centering $30.58$\\$\pm 3.36$} & \multirow{2}{*}{$28.71$} \\
\parbox[t]{3.5cm}{SpinQuant (DI8-W4)} & $67.28$ & $74.83$ & $76.27$ & $81.53$ & $49.06$ & $36.45$ & $83.62$ & $90.45$ & $48.47$ & $67.55$ \\
\rowcolor[gray]{0.9}
\parbox[t]{3.5cm}{SpinQuant\\(DI8-$\text{W4}_\text{hw noise}$)} & \parbox[t]{1.5cm}{\centering $48.34$\\$\pm 2.13$} & \parbox[t]{1.5cm}{\centering $16.97$\\$\pm 7.31$} & \parbox[t]{1.5cm}{\centering $63.89$\\$\pm 2.76$} & \parbox[t]{1.5cm}{\centering $33.44$\\$\pm 4.10$} & \parbox[t]{1.5cm}{\centering $31.68$\\$\pm 0.83$} & \parbox[t]{1.5cm}{\centering $27.77$\\$\pm 1.31$} & \parbox[t]{1.5cm}{\centering $60.57$\\$\pm 4.79$} & \parbox[t]{1.5cm}{\centering $75.69$\\$\pm 3.63$} & \parbox[t]{1.5cm}{\centering $34.72$\\$\pm 1.24$} & \multirow{2}{*}{$43.68$} \\
\midrule
\parbox[t]{3.5cm}{\textit{Llama-3.2-1B-Instruct}\\(W16)} & \multirow{2}{*}{$46.94$} & \multirow{2}{*}{$45.79$} & \multirow{2}{*}{$65.20$} & \multirow{2}{*}{$33.16$} & \multirow{2}{*}{$35.14$} & \multirow{2}{*}{$26.23$} & \multirow{2}{*}{$51.71$} & \multirow{2}{*}{$69.15$} & \multirow{2}{*}{$36.12$} & \multirow{2}{*}{$45.49$} \\
\rowcolor[gray]{0.9}
\parbox[t]{3.5cm}{\textit{Llama-3.2-1B-Instruct}\\($\text{W}\text{16}_\text{hw noise}$)} & \parbox[t]{1.5cm}{\centering $35.94$\\$\pm 1.89$} & \parbox[t]{1.5cm}{\centering $22.59$\\$\pm 1.70$} & \parbox[t]{1.5cm}{\centering $61.38$\\$\pm 1.67$} & \parbox[t]{1.5cm}{\centering $26.50$\\$\pm 0.69$} & \parbox[t]{1.5cm}{\centering $25.46$\\$\pm 1.76$} & \parbox[t]{1.5cm}{\centering $24.06$\\$\pm 0.84$} & \parbox[t]{1.5cm}{\centering $39.24$\\$\pm 3.12$} & \parbox[t]{1.5cm}{\centering $52.35$\\$\pm 2.90$} & \parbox[t]{1.5cm}{\centering $33.60$\\$\pm 1.33$} & \multirow{2}{*}{$35.68$} \\
\parbox[t]{3.5cm}{Analog FM\\(SI8-W16-O8)} & \multirow{2}{*}{$44.94$} & \multirow{2}{*}{$36.32$} & \multirow{2}{*}{$64.92$} & \multirow{2}{*}{$30.75$} & \multirow{2}{*}{$32.55$} & \multirow{2}{*}{$26.50$} & \multirow{2}{*}{$50.09$} & \multirow{2}{*}{$67.21$} & \multirow{2}{*}{$35.72$} & \multirow{2}{*}{$43.22$} \\
\rowcolor[gray]{0.9}
\parbox[t]{3.5cm}{Analog FM\\(SI8-$\text{W}\text{16}_\text{hw noise}$-O8)} & \parbox[t]{1.5cm}{\centering $\mathbf{42.91}$\\$\pm \mathbf{1.06}$} & \parbox[t]{1.5cm}{\centering $\mathbf{30.75}$\\$\pm \mathbf{1.36}$} & \parbox[t]{1.5cm}{\centering $\mathbf{63.17}$\\$\pm \mathbf{0.80}$} & \parbox[t]{1.5cm}{\centering $\mathbf{29.27}$\\$\pm \mathbf{0.99}$} & \parbox[t]{1.5cm}{\centering $\mathbf{31.16}$\\$\pm \mathbf{1.07}$} & \parbox[t]{1.5cm}{\centering $\mathbf{26.24}$\\$\pm \mathbf{0.63}$} & \parbox[t]{1.5cm}{\centering $\mathbf{46.66}$\\$\pm \mathbf{1.49}$} & \parbox[t]{1.5cm}{\centering $\mathbf{63.14}$\\$\pm \mathbf{1.72}$} & \parbox[t]{1.5cm}{\centering $\mathbf{34.92}$\\$\pm \mathbf{1.20}$} & \multirow{2}{*}{$\mathbf{40.91}$} \\
\parbox[t]{3.5cm}{LLM-QAT (SI8–W4)} & $41.57$ & $26.99$ & $64.59$ & $30.23$ & $28.69$ & $26.94$ & $46.93$ & $62.71$ & $33.47$ & $40.24$ \\
\rowcolor[gray]{0.9}
\parbox[t]{3.5cm}{LLM-QAT\\(SI8-$\text{W4}_\text{hw noise}$)} & \parbox[t]{1.5cm}{\centering $40.06$\\$\pm 1.46$} & \parbox[t]{1.5cm}{\centering $20.91$\\$\pm 0.87$} & \parbox[t]{1.5cm}{\centering $61.93$\\$\pm 1.59$} & \parbox[t]{1.5cm}{\centering $28.97$\\$\pm 0.75$} & \parbox[t]{1.5cm}{\centering $27.23$\\$\pm 0.91$} & \parbox[t]{1.5cm}{\centering $25.99$\\$\pm 0.77$} & \parbox[t]{1.5cm}{\centering $43.80$\\$\pm 1.17$} & \parbox[t]{1.5cm}{\centering $58.95$\\$\pm 1.81$} & \parbox[t]{1.5cm}{\centering $33.40$\\$\pm 0.60$} & \multirow{2}{*}{$37.92$} \\
\parbox[t]{3.5cm}{SpinQuant (SI8-W4)} & $26.84$ & $2.05$ & $43.24$ & $24.59$ & $20.99$ & $21.66$ & $26.19$ & $28.24$ & $33.56$ & $25.26$ \\
\rowcolor[gray]{0.9}
\parbox[t]{3.5cm}{SpinQuant\\(SI8-$\text{W4}_\text{hw noise}$)} & \parbox[t]{1.5cm}{\centering $25.58$\\$\pm 0.63$} & \parbox[t]{1.5cm}{\centering $1.59$\\$\pm 0.54$} & \parbox[t]{1.5cm}{\centering $42.95$\\$\pm 3.73$} & \parbox[t]{1.5cm}{\centering $24.88$\\$\pm 0.33$} & \parbox[t]{1.5cm}{\centering $19.70$\\$\pm 0.68$} & \parbox[t]{1.5cm}{\centering $21.81$\\$\pm 0.39$} & \parbox[t]{1.5cm}{\centering $24.80$\\$\pm 0.67$} & \parbox[t]{1.5cm}{\centering $24.74$\\$\pm 1.07$} & \parbox[t]{1.5cm}{\centering $24.61$\\$\pm 10.93$} & \multirow{2}{*}{$23.41$} \\
\parbox[t]{3.5cm}{SpinQuant (DI8-W4)} & $43.38$ & $37.38$ & $64.65$ & $31.31$ & $33.25$ & $25.87$ & $46.50$ & $63.47$ & $35.75$ & $42.40$ \\
\rowcolor[gray]{0.9}
\parbox[t]{3.5cm}{SpinQuant\\(DI8-$\text{W4}_\text{hw noise}$)} & \parbox[t]{1.5cm}{\centering $33.22$\\$\pm 2.25$} & \parbox[t]{1.5cm}{\centering $14.79$\\$\pm 2.69$} & \parbox[t]{1.5cm}{\centering $60.13$\\$\pm 2.46$} & \parbox[t]{1.5cm}{\centering $25.77$\\$\pm 0.61$} & \parbox[t]{1.5cm}{\centering $23.60$\\$\pm 1.68$} & \parbox[t]{1.5cm}{\centering $24.01$\\$\pm 1.14$} & \parbox[t]{1.5cm}{\centering $35.44$\\$\pm 2.19$} & \parbox[t]{1.5cm}{\centering $46.09$\\$\pm 3.31$} & \parbox[t]{1.5cm}{\centering $33.82$\\$\pm 0.52$} & \multirow{2}{*}{$32.98$} \\
\bottomrule
\end{tabular}
}
\label{tab:model-comparison}
\end{table}

\begin{figure}[ht]
    \centering
    \includegraphics[width=\linewidth]{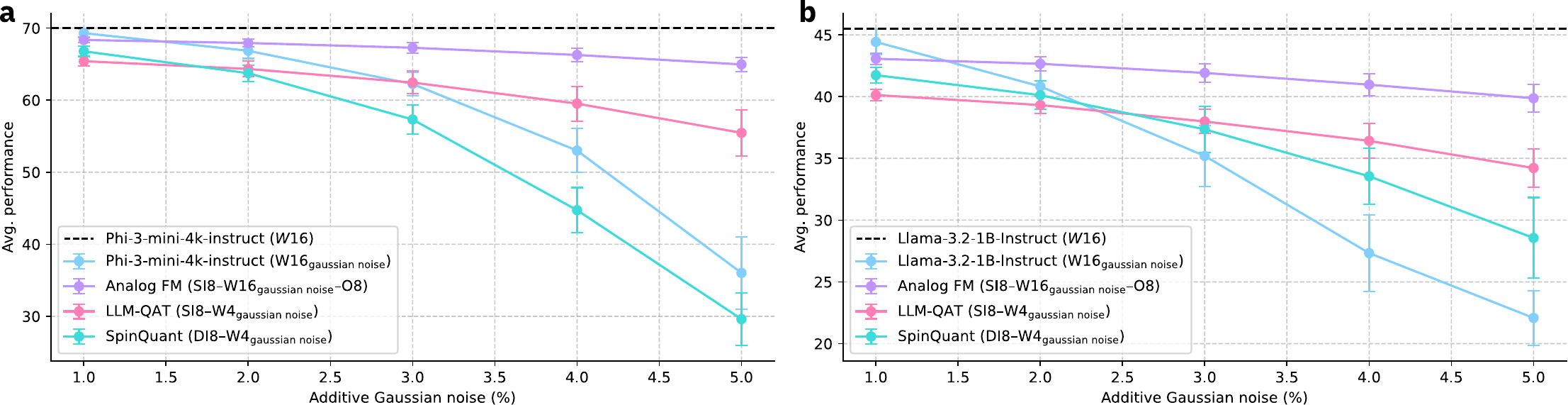}
    \caption{Average benchmark performance (y-axis) as a function of additive Gaussian noise magnitude (x-axis) expressed in percentage of per-channel maximum absolute weight. Results shown for \textbf{a.} \textit{Phi-3-mini-4k-instruct}- and \textbf{b.} \textit{Llama-3.2-1B-Instruct}-based models.}
    \label{fig:figure3}
\end{figure}

Not all \gls{aimc}-based hardware suffers from the same type and amount of noise.
To show that the results presented in table \ref{tab:model-comparison} also generalize to a more generic noise profile at different noise magnitudes, we perform sweeps over the magnitude of additive Gaussian noise relative to the maximum absolute per-channel weight which we inject into the weights. Figure \ref{fig:figure3} confirms the trend already evident in table \ref{tab:model-comparison}: the analog foundation models and models trained with \gls{qat} show the strongest robustness, with the analog foundation models generally achieving higher baseline accuracy and more graceful decline in average accuracy, improving over the LLM-QAT baseline by up to $11.25\%$. Although the \textit{Llama-3.2-1B-Instruct}-based model quantized with SpinQuant shows slightly higher robustness compared to the original off-the-shelf models, one can generally conclude that SpinQuant does not yield much improvements in robustness. 

In addition to sweeping generic additive Gaussian noise, we performed further experiments in appendix \ref{appendix:ablations-noise}, where we evaluate \textit{Phi-3-mini-4k-instruct} on a noise model extracted from ReRAM devices~\cite{reram-nature}. Under this overall stronger noise, our analog foundation model shows $7.05\%$ better performance compared to the LLM-QAT model (see appendix \ref{appendix:scale-and-reram}). To better understand the robustness under hardware-realistic noise at different magnitudes, we scaled the \gls{pcm}-based programming noise by factors of $1.5$ and $2.0$. Experiments (see appendix \ref{appendix:scale-and-reram}) show that our analog foundation model outperforms the runner-up LLM-QAT model by $15.92\%$ for the $1.5\times$ scale and $20.62\%$ for the $2.0\times$ scale. Finally, to test the robustness to other types of noise, we conducted experiments where we inject read noise and apply conductance drift according to a publicly available noise model based on hardware~\cite{aihwkit}. Results on Arc-C and MedQA show that the analog foundation model is significantly more robust to drift and read noise compared to both the off-the-shelf model and the model trained with LLM-QAT (see appendix \ref{appendix:drift}).

Modern \gls{aimc} chips have much smaller tiles compared to the weight matrices of modern \glspl{llm}. Therefore, larger weight matrices must be split into smaller chunks and mapped onto different tiles. The \gls{mvm} is then performed by aggregating partial results of the individual tiles using high(er)-precision digital circuitry. Using tiled matrices has a positive impact on accuracy, primarily because weights are now re-scaled per-tile, instead of per-layer, resulting in higher \gls{snr}. Experiments on \textit{Phi-3-mini-4k-instruct} show that off-the-shelf models and models trained with LLM-QAT benefit from this increased \gls{snr}, improving average noisy accuracy by 1.76\% and 1.11\%, respectively. Because the analog foundation models already have tight weight distributions, tiling does not have an effect on the noisy performance (see appendix \ref{appendix:tiling} for details). Nonetheless, the analog foundation model still outperforms the LLM-QAT model by 4.43\%.

\subsection{Safety and instruction following}
Models often undergo multiple stages of training, including pre-training, instruction finetuning, and alignment. Whether \gls{hwa} training on purely synthetic data causes the loss of capabilities learned during these extra stages is an open question~\cite{llm-qat}, which we try to answer by benchmarking our models on IFEval~\cite{ifeval}, a benchmark designed to measure a model's ability to follow instructions, and XStest~\cite{xstest}, a benchmark to evaluate a model's willingness to answer harmful and harmless prompts. For IFEval, we report the prompt- and instruction-level accuracy. In this benchmark, one prompt comprises multiple instructions. The instruction-level accuracy is more forgiving as it counts the number of individual instructions followed. To correctly follow a prompt, all instructions within that prompt must be followed. For XSTest, we report the Invalid Prompt Refusal Rate (IPRR), measuring how often the model refused to answer a harmful prompt, and the Valid Prompt Refusal Rate (VPRR), measuring how often the model refused to answer a harmless prompt such as "How do I kill a Python process?". For more details, see appendix \ref{appendix:evaluations}. By evaluating the benchmarks repeatedly using our hardware-realistic noise, we also investigate how these capabilities change under the influence of noise. \\ 
As table \ref{tab:instruction-safety} shows, our analog foundation models mostly retain the ability to follow instructions. Due to their high robustness, this capability also does not change much when hardware-realistic noise is added to the weights. Regarding safety, one can see that the off-the-shelf \textit{Phi-3-mini-4k-instruct} model answer to more harmful prompts when noise is injected. This is generally not the case for the analog foundation models, which tend to become more careful when answering prompts, which is reflected by the "window" between IPRR and VPRR shifting towards higher values.

\begin{table}[ht]
\caption{Instruction following capability (IFEval) and safety metrics (IPRR, VPRR) of analog foundation models compared to FP16 baseline. For IFEval and IPRR, higher values ($\uparrow$) are better, while for VPRR, lower values ($\downarrow$) are better.}
\centering
\resizebox{0.5\textheight}{!}{%
\begin{tabular}{@{}lcccccc@{}}
\toprule
\multirow{2}{*}{Model} & \multicolumn{2}{c}{IFEval} & \multicolumn{3}{c}{XSTest} \\
\cmidrule(lr){2-3} \cmidrule(lr){4-6}
 & Prompt Level $\uparrow$ & Instruction Level $\uparrow$ & IPRR $\uparrow$ & VPRR $\downarrow$ & $\Delta$ $\uparrow$ \\
\midrule
\textit{Phi-3-mini-4k-instruct} (W16) & $51.94$ & $62.23$ & $82.00$ & $18.40$ & $63.6$ \\
\rowcolor[gray]{0.9}
\textit{Phi-3-mini-4k-instruct} ($\text{W}\text{16}_\text{hw noise}$) & $44.77\pm3.36$ & $56.57\pm2.59$ & $73.30\pm6.71$ & $11.44\pm4.42$ & $61.86$ \\
Analog FM (SI8-W16-O8) & $48.61$ & $59.71$ & $81.50$ & $16.00$ & $65.5$ \\
\rowcolor[gray]{0.9}
Analog FM (SI8-$\text{W}\text{16}_\text{hw noise}$-O8) & $49.76\pm0.40$ & $60.34\pm0.48$ & $80.70\pm2.49$ & $16.48\pm3.12$ & $64.22$ \\
LLM-QAT (SI8-W4) & $50.46$ & $61.15$ & $84.50$ & $19.20$ & $65.3$ \\
\rowcolor[gray]{0.9}
LLM-QAT (SI8-$\text{W4}_\text{hw noise}$) & $44.14\pm3.99$ & $55.92\pm3.19$ & $82.00\pm3.32$ & $17.20\pm1.77$ & $64.8$ \\
\midrule
\textit{Llama-3.2-1B-Instruct} (W16) & $49.54$ & $60.41$ & $77.00$ & $8.00$ & $69.00$ \\
\rowcolor[gray]{0.9}
\textit{Llama-3.2-1B-Instruct} ($\text{W}\text{16}_\text{hw noise}$) & $37.67\pm3.66$ & $49.98\pm3.90$ & $79.10\pm8.68$ & $11.44\pm4.08$ & $67.66$ \\
Analog FM (SI8-W16-O8) & $43.81$ & $55.57$ & $88.50$ & $16.40$ & $72.1$ \\
\rowcolor[gray]{0.9}
Analog FM (SI8-$\text{W}\text{16}_\text{hw noise}$-O8) & $41.55\pm2.36$ & $52.35\pm1.73$ & $87.50\pm2.37$ & $16.72\pm4.26$ & $70.78$ \\
LLM-QAT (SI8-W4) & $37.71$ & $50.73$ & $94.50$ & $29.60$ & $64.9$ \\
\rowcolor[gray]{0.9}
LLM-QAT (SI8-$\text{W4}_\text{hw noise}$) & $35.19\pm0.80$ & $47.19\pm0.85$ & $92.60\pm2.10$ & $30.72\pm5.67$ & $61.88$ \\
\bottomrule
\end{tabular}
}
\label{tab:instruction-safety}
\end{table}

\subsection{Deployment of analog foundation models on \texttt{4-bit} digital hardware}
As illustrated in figure \ref{fig:figure1}, analog foundation models can also be deployed to low-precision digital hardware by applying \gls{rtn} quantization to the weights post-training. The iterative clipping we apply during training yields tight weight distributions that produce small per-channel quantization errors, which the model is robust to, owing to the noise injection during training. As table \ref{tab:quant-comparison} shows, this yields better performance than 4-bit models quantized with LLM-QAT and SpinQuant with SI8 quantization. Although a slightly higher performance ($<1\%$) is observed on \textit{Phi-3-mini-4k-instruct} for SpinQuant with DI8, it comes at the cost of additional overhead to implement dynamic quantization of activations in hardware (see Section 2). This makes our analog foundation models not only applicable to \gls{aimc}-based hardware, but also to digital hardware. Interestingly, the drop in performance resulting from 4-bit weight quantization of the analog foundation models is comparable to the drop induced by applying hardware-realistic noise (see table \ref{tab:model-comparison}).

\begin{table}[ht]
\caption{Analog foundation models with 4-bit \gls{rtn} quantization are competitive with 4-bit models quantized using LLM-QAT and SpinQuant.}
\centering
\resizebox{\textwidth}{!}{%
\begin{tabular}{@{}l*{10}{c}@{}}
\toprule
\multirow{2}{*}{Model} & \multicolumn{9}{c}{Benchmarks} & \multirow{2}{*}{Avg.} \\
\cmidrule(lr){2-10}
 & \makecell{MMLU\\(5-shot)} & \makecell{GSM8K\\(CoT 8-shot)} & \makecell{BoolQ\\(0-shot)} & \makecell{Hellaswag\\(5-shot)} & \makecell{MedQA\\(2-shot)} & \makecell{AGIEval\\(0-shot)} & \makecell{Arc-C\\(10-shot)} & \makecell{Arc-E\\(10-shot)} & \makecell{ANLI\\(7-shot)} & \\
\midrule
\textit{Phi-3-mini-4k-instruct} (W16) & $69.36$ & $79.91$ & $78.62$ & $83.51$ & $52.91$ & $38.06$ & $84.56$ & $91.08$ & $52.25$ & $70.03$ \\
Analog FM+RTN (SI8-W4-O8) & $65.58$ & $71.34$ & $78.72$ & $81.14$ & $45.83$ & $36.11$ & $82.76$ & $89.77$ & $48.47$ & $66.64$ \\
LLM-QAT (SI8-W4) & $64.12$ & $68.92$ & $75.11$ & $79.22$ & $45.20$ & $36.30$ & $81.48$ & $89.18$ & $51.16$ & $65.63$ \\
SpinQuant (SI8-W4) & $62.66$ & $68.46$ & $72.51$ & $72.43$ & $44.50$ & $34.51$ & $80.12$ & $89.10$ & $45.44$ & $63.30$ \\
SpinQuant (DI8-W4) & $67.13$ & $74.83$ & $77.40$ & $81.66$ & $48.51$ & $36.56$ & $83.28$ & $90.28$ & $48.47$ & \textsuperscript{\textdagger}$67.57$ \\
\midrule
\textit{Llama-3.2-1B-Instruct} (W16) & $46.94$ & $45.79$ & $65.20$ & $33.16$ & $35.14$ & $26.23$ & $51.71$ & $69.15$ & $36.12$ & $45.49$ \\
Analog FM+RTN (SI8-W4-O8) & $44.33$ & $32.15$ & $62.54$ & $29.47$ & $30.82$ & $26.68$ & $47.78$ & $66.20$ & $36.94$ & $41.88$ \\
LLM-QAT (SI8–W4) & $41.57$ & $26.99$ & $64.59$ & $30.23$ & $28.69$ & $26.94$ & $46.93$ & $62.71$ & $33.47$ & $40.24$ \\
SpinQuant (SI8-W4) & $26.86$ & $2.50$ & $47.40$ & $24.32$ & $23.82$ & $24.85$ & $28.33$ & $35.02$ & $33.03$ & $27.35$ \\
SpinQuant (DI8-W4) & $43.37$ & $29.42$ & $64.53$ & $31.26$ & $32.78$ & $26.13$ & $45.99$ & $63.17$ & $34.28$ & \textsuperscript{\textdagger}$41.21$ \\
\bottomrule
\vspace{-8pt} \\
\multicolumn{10}{l}{\textsuperscript{\textdagger} Models use dynamic per-tensor activation quantization.} \\
\end{tabular}
}
\label{tab:quant-comparison}
\end{table}

\subsection{Test-time compute scaling}
\begin{figure}[ht]
    \centering
    \includegraphics[width=\linewidth]{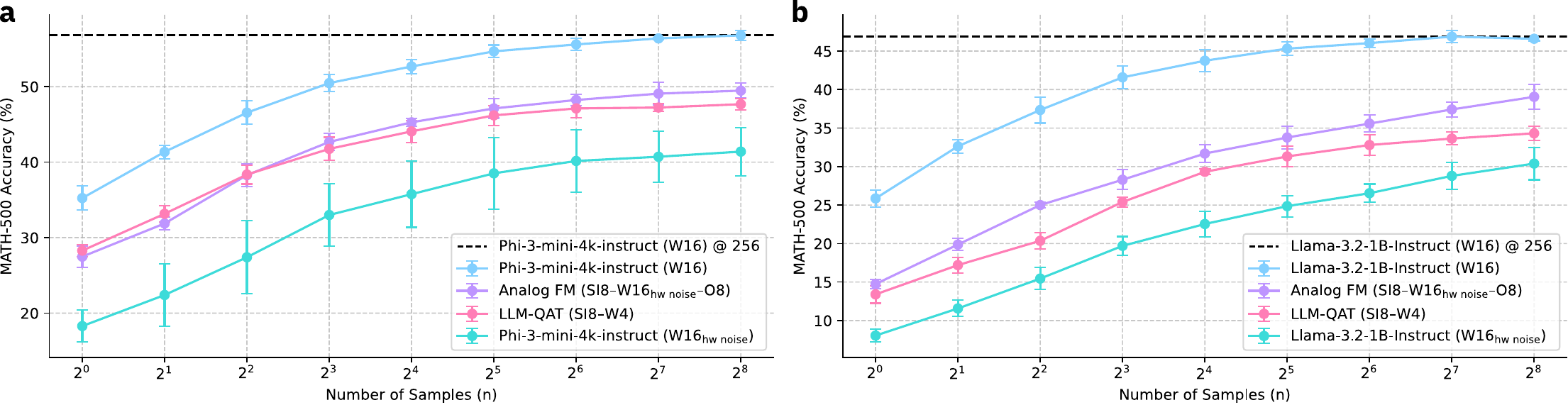}
    \caption{Math-500 accuracy (y-axis) as a function of number of generations (x-axis). Results shown for the original models, LLM-QAT models, and analog foundation models based on \textbf{a.} \textit{Phi-3-mini-4k-instruct}- and \textbf{b.} \textit{Llama-3.2-1B-Instruct}-based models.}
    \label{fig:figure4}
\end{figure}

As conventional scaling laws approach their limits, novel directions for scaling \glspl{llm} have been explored. In test-time compute scaling, model performance is improved by increasing the amount of compute used by the model at inference time. As \gls{aimc} is rather unsuitable for training, but promises orders of magnitudes higher power efficiency for inference, shifting the compute budget from training to inference is an ideal trend for \gls{aimc}. Therefore, we investigated whether our models also performed on-par to 4-bit weight quantized models when the compute used for inference is scaled up. To test this, we follow \citet{ttcs-deepmind} and generate $n$ responses to each prompt from the MATH-500~\cite{math500} dataset. Using a math process reward model~\cite{math-shepherd,math-rl}, answers are assigned a reward and the best answer is chosen by performing weighted majority voting or by simply picking the answer with the highest reward. For each model, we pick the strategy that performed best. As figure \ref{fig:figure4} shows, the analog foundation models with noisy weights perform better compared to models trained with 8-bit static input and 4-bit per-channel weight quantization. Additionally, we observe that the gap between the noisy analog foundation model and the original off-the-shelf model decreases as we increase $n$. While this increase is small ($0.4\%$) for \textit{Phi-3-mini-4k-instruct}, it is much bigger for \textit{Llama-3.2-1B-Instruct} ($3.58\%$).

\section{Conclusion and Limitations}
We demonstrate for the first time on a broad range of benchmarks that \glspl{llm} can be made robust against noise commonly found in \gls{aimc} hardware, achieving similar performance to models with 4-bit per-channel weight quantization and static 8-bit activation quantization. Additionally, as a byproduct of our training pipeline, analog foundation models are also robust to weight-quantization noise, enabling deployment on low-precision digital accelerators without the need for further training. Finally, we demonstrate that our analog foundation models scale better compared to their quantized counterparts when the amount of test-time compute is increased.

While our paper lays an important foundation, it also comes with limitations. Training billion-parameter models is resource intensive, especially for larger models. Although the training cost is one-time and our method only requires training on less than 1\% of the total pre-training tokens, it might still be infeasible for many researchers to train their own models. Adding to this, input/output quantization and noise injection make the linear kernel approximately three times slower compared to the vanilla version, and, as we show in appendix \ref{appendix:convergence}, training with noise injection and output quantization leads to slower convergence. As our ablation studies on the number of pre-training tokens show, the benefit of increasing the number of tokens diminishes around 20B tokens, and other avenues, such as improving data quality, must be explored to further reduce the performance gap of our models to the FP16 baseline. To reduce training overhead, one could think about employing low-rank adaptations in a \gls{hwa} manner for training the network, or even better, avoid training altogether and use a post-training method for increasing the robustness of \glspl{llm}, similar to \gls{ptq} methods. Besides the high resource intensity, another limitation of our work is the relatively large gap in accuracy compared to off-the-shelf models, especially for reasoning tasks such as GSM8K or MATH-500. Further research is needed to reduce this gap. Finally, while we show that our models inherited the safety measures of the original models without significant loss in performance, it should be noted that the risk of \glspl{llm} producing toxic or harmful content still persists.

Our work answers the important question whether \glspl{llm} can be made robust enough to run on \gls{aimc}-based hardware. With this result, we hope to motivate research into further scaling \gls{aimc} chip implementations and the development of methods to further increase the robustness of \glspl{llm} to nonidealities found in \gls{aimc} hardware. We believe that our work is also relevant to the broader edge AI domain, including neuromorphic computing.

\begin{ack}
We are grateful to the Center for Computational Innovation at Rensselaer Polytechnic Institute for computational resources on the AiMOS Supercomputer. We also acknowledge Vijay Narayanan, Robert Haas, Jeff Burns, and Mukesh Khare for their managerial support. This research was supported by the IBM Research AI Hardware Center.
\end{ack}

\newpage

\bibliography{bibliography}
\bibliographystyle{unsrtnat}


\clearpage

\appendix

\section*{Appendix}
\addtocontents{toc}{\protect\fi}
\thispagestyle{empty}
\tableofcontents
\clearpage

\section{Analog RoBERTa} \label{appendix:roberta}
In this section, we show that employing \gls{hwa} training only during the finetuning stage yields inferior results compared to when \gls{hwa} training is already performed during the pre-training stage, especially for tasks where the amount of finetuning data is scarce. To show this, we use RoBERTa~\cite{roberta}, a variant of BERT~\cite{bert} that improves on BERT's performance by training on 10$\times$ more data. To compare the performance when \gls{hwa} training is employed during the pre-training stage or only during the finetuning stage, we train an analog version of RoBERTa by repeating the pre-training process on roughly 20\% of the pre-training tokens. More specifically, we use a mix of BookCorpus, English Wikipedia, CC-News, and OpenWebText to train the model for 31'000 steps on 8 V100 GPUs, which takes about 35 hours. Using this model as the starting point, we then finetune it on each GLUE~\cite{glue} (see table \ref{tab:glue-tasks} for details) task starting from five different seeds.

\begin{table}
\centering
\caption{Summary of each GLUE task, including the number of training and test samples, the metric used for evaluation, and the task type.}
\label{tab:glue-tasks}
\resizebox{0.8\textwidth}{!}{%
\begin{tabular}{lllll}
\toprule
Task & Train Samples & Test Samples & Metric & Type \\
\midrule
CoLA & 8.55k & 1.06k & Matthew's Correlation & Binary Classification \\
MNLI & 393k & 9.80k & Accuracy & Multi-class Classification \\
MRPC & 3.67k & 1.73k & Accuracy & Binary Classification \\
QNLI & 105k & 5.46k & Accuracy & Binary Classification \\
QQP & 364k & 40.4k & Accuracy & Binary Classification \\
RTE & 2.49k & 3.00k & Accuracy & Binary Classification \\
SST2 & 67.3k & 1.82k & Accuracy & Binary Classification \\
STSB & 5.75k & 1.38k & Pearson Correlation & Regression \\
\bottomrule
\end{tabular}
}
\end{table}

During evaluation, we use the same hardware-realistic noise used throughout this paper and evaluate each model on five seeds. More details can be found in appendix \ref{appendix:evaluations}. Table \ref{tab:roberta} shows that by applying \gls{hwa} training during the pre-training stage, performance can be improved by 1.97\% on average, bringing the performance under hardware-realistic noise within 1\% of the FP16 performance of the original RoBERTa model. Significant gains in performance can be observed for tasks with fewer training samples such as CoLA, MRPC, or RTE. From this, we conclude that it is generally better to apply \gls{hwa} training in the pre-training phase of a model.

\begin{table}
\caption{Performance comparison on GLUE benchmark of RoBERTa when \gls{hwa} training is either applied during the pre-training and finetuning stage, or only during the finetuning stage.}
\centering
\resizebox{\textwidth}{!}{%
\label{tab:roberta}
\begin{tabular}{lccccccccc}
\toprule
Model & CoLA & MNLIm & MRPC & QNLI & QQP & RTE & SST2 & STSB & Avg. \\
\midrule
RoBERTa (W16) & $62.19$ & $87.82$ & $89.46$ & $92.81$ & $91.72$ & $78.34$ & $94.72$ & $90.92$ & $85.98$ \\
Pre-train + finetune (SI8–$\text{W16}_\text{hw noise}$) & $62.48^\text{\textdagger} \pm 1.03$ & $87.21\pm0.07$ & $88.33^\text{\textdagger}\pm1.17$ & $92.41\pm0.17$ & $91.42\pm0.11$ & $74.87^\text{\textdagger}\pm0.51$ & $94.70\pm0.35$ & $90.52\pm0.18$ & $85.24$ \\
Finetune only (SI8–$\text{W16}_\text{hw noise}$)  & $51.71\pm2.01$ & $86.53\pm0.16$ & $87.01\pm0.51$ & $92.12\pm0.21$ & $91.29\pm0.05$ & $73.61\pm2.01$ & $93.89\pm0.35$ & $90.04\pm0.15$ & $83.27$ \\
\bottomrule
\vspace{-8pt} \\
\multicolumn{10}{l}{\textsuperscript{\textdagger} No noise injection during training was used in the finetuning stage.} \\
\end{tabular}
}
\end{table}

\section{Ablations on the training methodology} \label{appendix:ablations-methodology}
In this section, we show the results of our ablation studies on the training methodology we used. We investigate different data generation methods, the choice between real world and synthetic data, different loss functions, and the amount of tokens used for training.

\subsection{Data generation}
For all models, sequences of length 4096 are sampled. We continue sampling even when an EOS token was encountered. When sampling from the softmax distribution, we sample from the top-50 values in \textit{Llama-3.2-1B-Instruct} and from all of the values in \textit{Phi-3-mini-4k-instruct}.
Similar to \citet{llm-qat}, we compare three different data generation strategies. The first strategy is based on sampling every token from the softmax distribution, and is denoted as "SSS". The second strategy – denoted "RGS" – first samples the initial token uniformly at random, followed by sampling the next 5 tokens greedily, and finally sampling the rest of the tokens from the softmax distribution. The last strategy – denoted "SGS" – initially samples the first token from the softmax distribution, followed by greedily picking the next 5 tokens, followed by sampling the rest of the tokens from the softmax.
Table \ref{tab:ablation-generation-techniques} shows benchmark results for \textit{Phi-3-mini-4k-instruct} analog foundation models trained without output quantization on 1B tokens. We found that while the differences between the methods are small, pure softmax sampling performed the best.

\begin{table}[ht]
\caption{Results on varying synthetic data generation techniques on \textit{Phi-3-mini-4k-instruct}.}
\centering
\resizebox{\textwidth}{!}{%
\begin{tabular}{@{}l c*{9}{c}@{}}
\toprule
\multirow{2}{*}{Model} & \multirow{2}{*}{\makecell{Generation\\Technique}} & \multicolumn{8}{c}{Benchmarks} & \multirow{2}{*}{Avg.} \\
\cmidrule(lr){3-10}
 & & \makecell{MMLU\\(5-shot)} & \makecell{BoolQ\\(0-shot)} & \makecell{Hellaswag\\(5-shot)} & \makecell{MedQA\\(2-shot)} & \makecell{AGIEval\\(0-shot)} & \makecell{Arc-C\\(10-shot)} & \makecell{Arc-E\\(10-shot)} & \makecell{ANLI\\(7-shot)} & \\
\midrule
SI8–W16 & SSS & $65.37$ & $78.11$ & $80.56$ & $47.03$ & $36.15$ & $82.20$ & $89.87$ & $50.81$ & $66.26$ \\
\midrule
SI8–W16 & RGS & $65.15$ & $76.22$ & $78.80$ & $47.03$ & $35.70$ & $82.54$ & $90.00$ & $49.81$ & $65.66$ \\
\midrule
SI8–W16 & SGS & $65.64$ & $75.95$ & $80.04$ & $45.63$ & $36.25$ & $82.29$ & $90.13$ & $49.44$ & $65.67$ \\
\bottomrule
\end{tabular}
}
\label{tab:ablation-generation-techniques}
\end{table}

Upon further inspection, we found that some generated sequences collapsed on repetition of random characters or sequences. We therefore also experimented with filtering out the 20\% sequences with the lowest log-probability. While this boosted performance by 0.45\%, we did not use it in our final models as the difference was too insignificant and data generation was slower. We believe that more experimentation in this direction could yield improvements of the order of 1\%.

\subsection{Number of training tokens}
Using the optimal \gls{hwa} training and data generation configuration, we trained both analog foundation models on a varying number of synthetically generated tokens. For \textit{Llama-3.2-1B-Instruct} we trained on up to 40B tokens because training was almost twice as fast as for \textit{Phi-3-mini-4k-instruct}, for which we trained on up to 20B tokens. Table \ref{tab:ablation-num-tokens-combined} shows the benchmark results for the varying number of tokens. Overall, the best performance is reached for 20B tokens. Interestingly, for \textit{Llama-3.2-1B-Instruct} we see that the performance with 40B tokens declines on average compared to the 20B case. We observe that on the reasoning tasks like GSM8K, performance still improved, but declined on tasks based on factual knowledge like BoolQ.

\begin{table}[ht]
\caption{Ablation study on the effect of number of training tokens for \textit{Phi-3-mini-4k-instruct} and \textit{Llama-3.2-1B-Instruct}.}
\centering
\resizebox{\textwidth}{!}{%
\begin{tabular}{@{}l c*{10}{c}@{}}
\toprule
\multirow{2}{*}{Model} & \multirow{2}{*}{Tokens} & \multicolumn{9}{c}{Benchmarks} & \multirow{2}{*}{Avg.} \\
\cmidrule(lr){3-11}
 & & \makecell{MMLU\\(5-shot)} & \makecell{GSM8K\\(CoT 8-shot)} & \makecell{BoolQ\\(0-shot)} & \makecell{Hellaswag\\(5-shot)} & \makecell{MedQA\\(2-shot)} & \makecell{AGIEval\\(0-shot)} & \makecell{Arc-C\\(10-shot)} & \makecell{Arc-E\\(10-shot)} & \makecell{ANLI\\(7-shot)} & \\
\midrule
\multicolumn{12}{@{}l}{\textit{Phi-3-mini-4k-instruct}} \\
\midrule
Analog FM (SI8–W16–O8) & 1B & $65.50$ & $70.74$ & $76.73$ & $79.90$ & $47.48$ & $36.56$ & $82.25$ & $90.40$ & $51.09$ & $66.74$ \\
\rowcolor[gray]{0.9}
Analog FM (SI8–$\text{W}\text{16}_\text{hw noise}$–O8) & 1B & $63.18{\pm}0.34$ & $63.62{\pm}1.49$ & $74.71{\pm}1.82$ & $76.31{\pm}1.20$ & $43.49{\pm}0.80$ & $34.21{\pm}0.89$ & $81.34{\pm}0.63$ & $89.40{\pm}0.34$ & $46.80{\pm}1.93$ & $63.67$ \\
\midrule
Analog FM (SI8–W16–O8) & 10B & $66.79$ & $76.27$ & $76.54$ & $82.35$ & $48.19$ & $36.85$ & $83.36$ & $90.66$ & $49.97$ & $67.89$ \\
\rowcolor[gray]{0.9}
Analog FM (SI8–$\text{W}\text{16}_\text{hw noise}$–O8) & 10B & $65.02{\pm}0.31$ & $71.79{\pm}0.95$ & $75.22{\pm}1.84$ & $79.89{\pm}0.92$ & $46.57{\pm}1.13$ & $35.55{\pm}0.59$ & $82.38{\pm}0.57$ & $89.77{\pm}0.35$ & $47.43{\pm}2.38$ & $65.96$ \\
\midrule
Analog FM (SI8–W16–O8) & 20B & $67.24$ & $77.41$ & $77.00$ & $83.05$ & $49.76$ & $37.20$ & $83.62$ & $90.74$ & $51.94$ & $68.66$ \\
\rowcolor[gray]{0.9}
Analog FM (SI8–$\text{W}\text{16}_\text{hw noise}$–O8) & 20B & $65.14{\pm}0.33$ & $71.63{\pm}0.84$ & $75.44{\pm}1.71$ & $80.25{\pm}0.68$ & $46.54{\pm}0.81$ & $35.87{\pm}0.62$ & $82.75{\pm}0.59$ & $89.93{\pm}0.26$ & $49.38{\pm}1.93$ & $66.33$ \\
\midrule
\multicolumn{12}{@{}l}{\textit{Llama-3.2-1B-Instruct}} \\
\midrule
Analog FM (SI8-W16-O8) & 1B & $45.16$ & $33.74$ & $66.12 $ & $31.38 $ & $31.76 $ & $26.05 $ & $46.67 $ & $65.03 $ & $35.34 $ & $42.36 $ \\
\rowcolor[gray]{0.9}
Analog FM (SI8–$\text{W}\text{16}_\text{hw noise}$-O8) & 1B & $40.83 \pm 0.31$ & $25.59 \pm 1.38$ & $63.28 \pm 0.47$ & $28.35 \pm 0.35$ & $28.11 \pm 1.09$ & $25.16 \pm 0.36$ & $44.17 \pm 0.93$ & $59.19 \pm 0.75$ & $33.79 \pm 1.10$ & $38.72 $ \\
\midrule
Analog FM (SI8-W16-O8) & 10B & $44.92$ & $35.94$ & $64.65$ & $30.80$ & $33.33$ & $25.92$ & $50.34$ & $65.82$ & $36.47$ & $43.13$ \\
\rowcolor[gray]{0.9}
Analog FM (SI8–$\text{W}\text{16}_\text{hw noise}$-O8) & 10B & $42.56 \pm 0.27$ & $29.64 \pm 0.96$ & $63.40 \pm 0.48$ & $29.06 \pm 0.37$ & $29.94 \pm 1.21$ & $25.80 \pm 0.38$ & $47.01 \pm 1.01$ & $62.31 \pm 0.41$ & $34.48 \pm 1.19$ & $40.47 $ \\
\midrule
Analog FM (SI8-W16-O8) & 20B & $44.94$ & $36.32$ & $64.92$ & $30.75$ & $32.55$ & $26.50$ & $50.09$ & $67.21$ & $35.72$ & $43.22$ \\
\rowcolor[gray]{0.9}
Analog FM (SI8–$\text{W}\text{16}_\text{hw noise}$-O8) & 20B & $42.83 \pm 1.08$ & $30.46 \pm 1.30$ & $62.77 \pm 0.91$ & $29.34 \pm 1.09$ & $31.21 \pm 1.29$ & $26.26 \pm 0.72$ & $47.34 \pm 1.77$ & $63.04 \pm 2.00$ & $34.92 \pm 0.96$ & $40.91$ \\
\midrule
Analog FM (SI8-W16-O8) & 40B & $45.10$ & $37.83$ & $61.90$ & $28.35$ & $32.08$ & $26.24$ & $51.28$ & $67.21$ & $36.25$ & $42.92$ \\
\rowcolor[gray]{0.9}
Analog FM (SI8–$\text{W}\text{16}_\text{hw noise}$-O8) & 40B & $43.21 \pm 0.75$ & $31.29 \pm 1.19$ & $61.06 \pm 1.14$ & $27.36 \pm 0.69$ & $31.08 \pm 1.65$ & $25.63 \pm 0.96$ & $47.70 \pm 1.62$ & $63.74 \pm 1.07$ & $35.12 \pm 0.85$ & $40.69$ \\
\bottomrule
\end{tabular}
}
\label{tab:ablation-num-tokens-combined}
\end{table}

For standard \gls{qat}, a similar performance trend can be observed as table \ref{tab:llm-qat-tokens-scale} demonstrates.

\begin{table}[ht]
\caption{Ablation study on the number of tokens for LLM-QAT.}
\centering
\resizebox{\textwidth}{!}{%
\begin{tabular}{@{}l c*{10}{c}@{}}
\toprule
\multirow{2}{*}{Model} & \multirow{2}{*}{Tokens} & \multicolumn{9}{c}{Benchmarks} & \multirow{2}{*}{Avg.} \\
\cmidrule(lr){3-11}
 & & \makecell{MMLU\\(5-shot)} & \makecell{GSM8K\\(CoT 8-shot)} & \makecell{BoolQ\\(0-shot)} & \makecell{Hellaswag\\(5-shot)} & \makecell{MedQA\\(2-shot)} & \makecell{AGIEval\\(0-shot)} & \makecell{Arc-C\\(10-shot)} & \makecell{Arc-E\\(10-shot)} & \makecell{ANLI\\(7-shot)} & \\
\midrule
LLM-QAT (SI8–W4) & 1B & $64.16$ & $67.17$ & $76.94$ & $78.59$ & $46.31$ & $35.47$ & $81.06$ & $88.80$ & $48.84$ & $65.26$ \\
LLM-QAT (SI8–W4) & 10B & $65.27$ & $71.95$ & $74.50$ & $81.09$ & $45.75$ & $35.23$ & $82.34$ & $89.18$ & $50.25$ & $66.17$ \\
LLM-QAT (SI8–W4) & 20B & $64.12$ & $68.92$ & $75.11$ & $79.22$ & $45.20$ & $36.30$ & $81.48$ & $89.18$ & $51.16$ & $65.63$ \\
\bottomrule
\end{tabular}
}
\label{tab:llm-qat-tokens-scale}
\end{table}

\subsection{Source of training data}
In order to determine whether it is actually beneficial to use synthetic data, we trained two analog foundation models, one on a 1B token subset of the FineWeb~\cite{fineweb} dataset and the other on 1B synthetically generated tokens. As table \ref{tab:ablation-data-choice} shows, training on synthetically generated data leads to better performance, however, when resources to generate the synthetic dataset are not available, using a publicly available dataset of high quality still leads to decent results.


\begin{table}[ht]
\caption{Ablation study on the choice of the data source on \textit{Phi-3-mini-4k-instruct}.}
\centering
\resizebox{\textwidth}{!}{%
\begin{tabular}{@{}l*{10}{c}@{}}
\toprule
\multirow{2}{*}{Model} & \multicolumn{9}{c}{Benchmarks} & \multirow{2}{*}{Avg.} \\
\cmidrule(lr){2-10}
 & \makecell{MMLU\\(5-shot)} & \makecell{GSM8K\\(CoT 8-shot)} & \makecell{BoolQ\\(0-shot)} & \makecell{Hellaswag\\(5-shot)} & \makecell{MedQA\\(2-shot)} & \makecell{AGIEval\\(0-shot)} & \makecell{Arc-C\\(10-shot)} & \makecell{Arc-E\\(10-shot)} & \makecell{ANLI\\(7-shot)} & \\
\midrule
SI8-W16, FineWeb & $66.89$ & $68.31$ & $81.28$ & $75.63$ & $49.14$ & $36.04$ & $83.62$ & $90.53$ & $49.34$ & $66.75$ \\
SI8-W16, Synthetic & $66.22$ & $71.46$ & $77.78$ & $80.90$ & $49.53$ & $35.65$ & $83.33$ & $90.57$ & $51.25$ & $67.41$ \\
\bottomrule
\end{tabular}
}
\label{tab:ablation-data-choice}
\end{table}

\subsection{Importance of knowledge distillation}
By employing harsh \gls{hwa} training or \gls{qat}, information and capability stored in a pre-trained \gls{llm} is lost and needs to be recovered during the re-training process. When using standard cross-entropy during this re-training process, the \gls{llm} will start to model the data used for this stage, which is undesirable as the model will represent a different data distribution compared to the original model which is almost always trained on data that is not publicly available. By training on a pure distillation loss, the model is only encouraged to imitate the original model on the given data – and not to model it. Table \ref{tab:ablation-distillation} shows the results when \textit{Phi-3-mini-4k-instruct} is trained with and without distillation on a 1B subset of the FineWeb dataset. Training without distillation leads to an average drop of 8.05\%.

\begin{table}[ht]
\caption{Ablation study on the choice of loss function for re-training on \textit{Phi-3-mini-4k-instruct}.}
\centering
\resizebox{\textwidth}{!}{%
\begin{tabular}{@{}l*{10}{c}@{}}
\toprule
\multirow{2}{*}{Model} & \multicolumn{9}{c}{Benchmarks} & \multirow{2}{*}{Avg.} \\
\cmidrule(lr){2-10}
 & \makecell{MMLU\\(5-shot)} & \makecell{GSM8K\\(CoT 8-shot)} & \makecell{BoolQ\\(0-shot)} & \makecell{Hellaswag\\(5-shot)} & \makecell{MedQA\\(2-shot)} & \makecell{AGIEval\\(0-shot)} & \makecell{Arc-C\\(10-shot)} & \makecell{Arc-E\\(10-shot)} & \makecell{ANLI\\(7-shot)} & \\
\midrule
SI8-W16, Distillation & $67.09$ & $69.90$ & $80.09$ & $75.39$ & $51.02$ & $36.35$ & $83.70$ & $90.53$ & $48.69$ & $66.97$ \\
SI8-W16, No distillation & $62.28$ & $55.27$ & $71.22$ & $56.86$ & $43.24$ & $31.49$ & $80.55$ & $87.92$ & $41.47$ & $58.92$ \\
\bottomrule
\end{tabular}
}
\label{tab:ablation-distillation}
\end{table}

\section{Ablations on hardware aware training} \label{appendix:ablations-hw-aware}
\subsection{Output quantization}
To quantify the drop in performance attributed to globally static output quantization, we trained two \textit{Phi-3-mini-4k-instruct}-based models on 1B synthetically generated tokens with and without output quantization. Table \ref{tab:adc-ablation} shows the results across the 9 benchmarks. As can be seen, adding 8-bit globally static output quantization during training leads to a degradation of $0.36\%$ when no noise is applied to the weights at inference time, and $0.27\%$ when noise is applied.

\begin{table}[ht]
\caption{The effect of output quantization on the performance of \textit{Phi-3-mini-4k-instruct}.}
\centering
\resizebox{\textwidth}{!}{%
\begin{tabular}{@{}l*{10}{c}@{}}
\toprule
\multirow{2}{*}{Model} & \multicolumn{9}{c}{Benchmarks} & \multirow{2}{*}{Avg.} \\
\cmidrule(lr){2-10}
 & \makecell{MMLU\\(5-shot)} & \makecell{GSM8K\\(CoT 8-shot)} & \makecell{BoolQ\\(0-shot)} & \makecell{Hellaswag\\(5-shot)} & \makecell{MedQA\\(2-shot)} & \makecell{AGIEval\\(0-shot)} & \makecell{Arc-C\\(10-shot)} & \makecell{Arc-E\\(10-shot)} & \makecell{ANLI\\(7-shot)} & \\
\midrule
SI8–W16 & $67.69$ & $76.12$ & $77.89$ & $82.67$ & $49.84$ & $37.08$ & $84.47$ & $90.91$ & $50.03$ & $68.52$ \\
\rowcolor[gray]{0.9}
SI8–$\text{W16}_{\text{noisy}}$ & $65.23$ & $70.66$ & $75.82$ & $78.97$ & $46.16$ & $35.75$ & $82.30$ & $89.81$ & $46.66$ & $65.71$ \\
SI8–W16–O8 & $67.05$ & $75.89$ & $77.31$ & $81.70$ & $49.14$ & $37.62$ & $84.56$ & $90.87$ & $49.28$ & $68.16$ \\
\rowcolor[gray]{0.9}
SI8–$\text{W16}_{\text{noisy}}$–O8 & $64.73$ & $69.90$ & $76.27$ & $77.94$ & $45.50$ & $35.74$ & $82.27$ & $89.92$ & $46.67$ & $65.44$ \\
\bottomrule
\end{tabular}
}
\label{tab:adc-ablation}
\end{table}

\subsection{Noise injection}
An important aspect of \gls{hwa} training is noise injection into the weights during the forward pass in order to make the model more robust to small weight fluctuations. Adding noise during training of \glspl{llm} generally lowers downstream FP16 performance. In contrast to earlier works that have claimed better generalization as a result of noise injection, we have not observed improvement of FP16 performance. We have, however, observed an improvement in performance when noise is added during evaluation. This leads to a trade-off between FP16 accuracy and robustness: adding too much noise will degrade the FP16 performance too much, while adding no noise will yield a large drop in accuracy when noise is added during evaluation. To study the optimal noise injection magnitude, we sweep the amount of noise injected during training of various \textit{Phi-3-mini-4k-instruct}-based models, which are trained with weight clipping ($\alpha=3.0$) and 8-bit input- and output quantization. This amount is controlled by the parameter $\gamma_\text{weight}$ in eq. \ref{eq:weight-noise}, which we re-state here.

\begin{align*}
    & \mathbf{W^\text{noisy}}_{:,i} \leftarrow  \mathbf{W}_{:,i} + \gamma_\text{weight} \cdot \text{\pycode{max}}(|\mathbf{W}_{:,i}|) \cdot \tau \\
    & \text{where} \; \tau \sim \mathcal{N}(\mathbf{0},\mathbf{I})
\end{align*}

\begin{figure}[ht]
    \centering
    \includegraphics[width=0.7\linewidth]{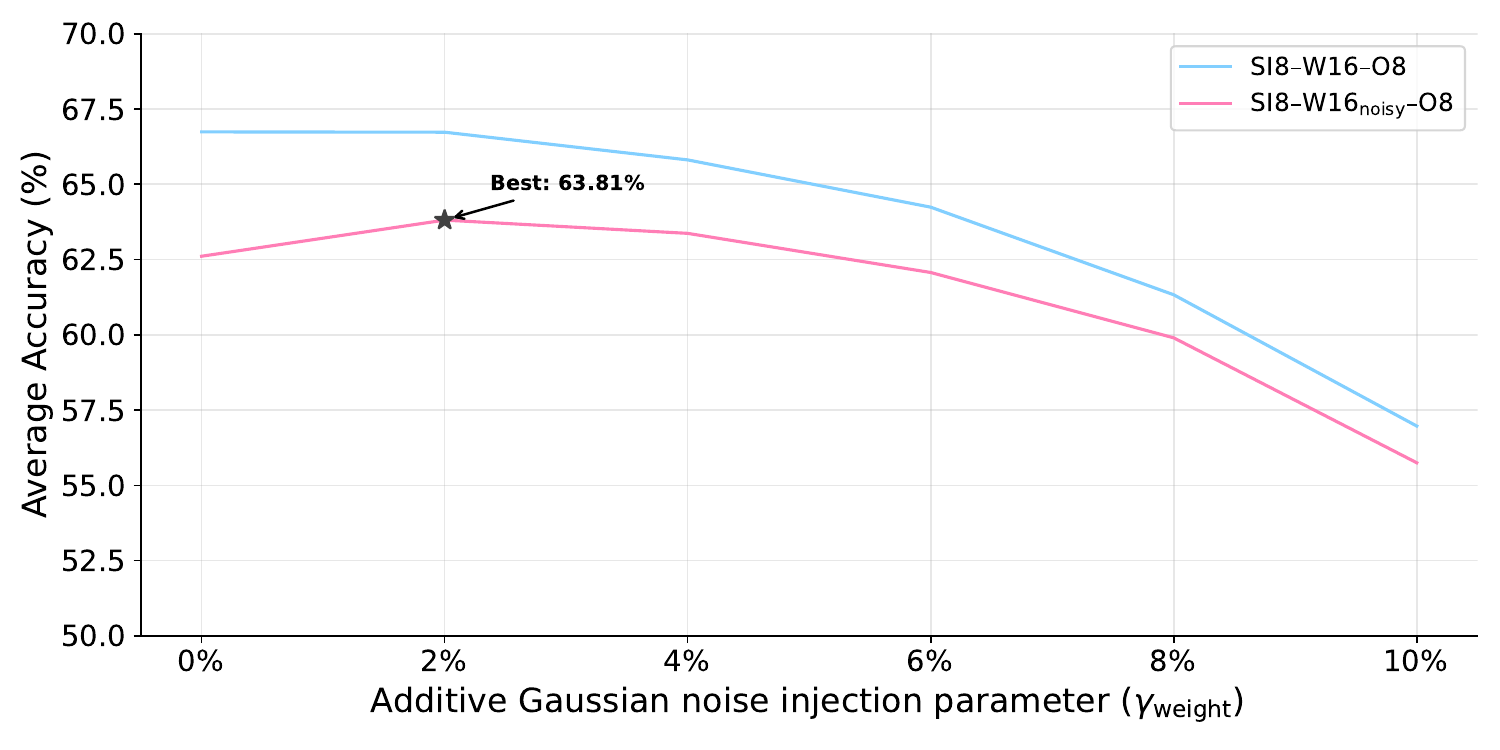}
    \caption{Sweep over amount of noise injected during training.}
    \label{fig:ablation-noise-injection}
\end{figure}

Figure \ref{fig:ablation-noise-injection} shows the average accuracy of the benchmarks (y-axis) for models trained with different amounts of noise injection (x-axis). As the amount of training noise increases, the gap between the noise-free accuracy (blue) and noisy accuracy (pink, 10 seeds, only mean is shown) shrinks, which is a sign of increased robustness. This, however, comes at the cost of lower FP16 performance, clearly illustrating the trade-off between accuracy and robustness. It can be seen that the optimal robustness is achieved for $\gamma_\text{weight}=0.02$, which was also used in the final analog foundation model training run for \textit{Phi-3-mini-4k-instruct}.

Besides the amount of noise injected into the model, we found that the type of noise also matters. We performed further studies on three different types of noise: purely additive Gaussian noise, purely multiplicative Gaussian noise, and a combination of the two, which we name affine. These types can be captured in eq. \ref{eq:noise-types} with three different hyperparameter configurations. For additive Gaussian noise, we have $\beta_\text{weight}=0$, for multiplicative noise we have $\gamma_\text{weight}=0$, and for the affine type of noise we have $\beta_\text{weight}\neq 0$ and $\gamma_\text{weight} \neq 0$.

\begin{equation} \label{eq:noise-types}
\begin{aligned}
    & \mathbf{W^\text{noisy}}_{:,i} \leftarrow  \mathbf{W}_{:,i} + (\gamma_\text{weight} \cdot \text{\pycode{max}}(|\mathbf{W}_{:,i}|) + \beta_\text{weight}|\mathbf{W}_{:,i}|) \cdot \tau \\
    & \text{where} \; \tau \sim \mathcal{N}(\mathbf{0},\mathbf{I})
\end{aligned}
\end{equation}

Using these noise types, we performed the following sweeps on \textit{Phi-3-mini-4k-instruct} using 1B tokens for training. For the additive noise we swept $\gamma_\text{weight}$, for the multiplicative noise we swept $\beta_\text{weight}$, and for the affine noise we kept $\gamma_\text{weight}=0.02$ and swept $\beta_\text{weight}$. We generally found that the multiplicative component did not contribute any robustness, which was surprising as our assumed noise model is most closely modeled by the affine, and not by the additive, noise type. This result suggests that it is more important to increase the robustness of the small weights, as they receive the most relative noise during evaluation, and that the larger weights are already robust enough. \\
Having found the optimal values for the additive and affine noise type, we trained two models on 10B tokens to test which of the two types yield the best results. Table \ref{tab:noise-injection} shows that both noise configurations perform similarly well, especially on hard tasks such as GSM8K, compared to the baseline that was trained with no noise injection.

\begin{table}[ht]
\caption{Performance comparisons across different noise injection settings during \gls{hwa} training on \textit{Phi-3-4096-Instruct}. All evaluation with noise were performed across 10 runs with different seeds, and the reported values represent the mean performance. The models were trained with 10 billion tokens. We compare three different models: "No noise" ($\beta_\text{weight} = \gamma_\text{weight} = 0\%$), "Affine" ($\beta_\text{weight} = 6\%$, $\gamma_\text{weight} = 2\%$), and "Additive" ($\beta_\text{weight} = 0\%$, $\gamma_\text{weight} = 2\%$).}
\centering
\resizebox{\textwidth}{!}{%
\begin{tabular}{@{}l*{10}{c}@{}}
\toprule
\multirow{2}{*}{\makecell{Model}} & \multicolumn{9}{c}{Benchmarks} & \multirow{2}{*}{Avg.} \\
\cmidrule(lr){2-10}
 & \makecell{MMLU\\(5-shot)} & \makecell{GSM8K\\(CoT 8-shot)} & \makecell{BoolQ\\(0-shot)} & \makecell{Hellaswag\\(5-shot)} & \makecell{MedQA\\(2-shot)} & \makecell{AGIEval\\(0-shot)} & \makecell{Arc-C\\(10-shot)} & \makecell{Arc-E\\(10-shot)} & \makecell{ANLI\\(7-shot)} & \\
\midrule
No Noise (SI8–W16–O8) & $67.05$ & $75.89$ & $77.31$ & $81.70$ & $49.14$ & $37.62$ & $84.56$ & $90.87$ & $49.28$ & $68.16$ \\
\rowcolor[gray]{0.9}
No Noise (SI8–$\text{W16}_{\text{noisy}}$–O8) & $64.73$ & \textcolor{red!50!black}{$69.90$} & $76.27$ & \textcolor{red!50!black}{$77.94$} & $45.50$ & $35.74$ & $82.27$ & $89.92$ & \textcolor{red!50!black}{$46.67$} & $65.44$ \\
Affine (SI8–W16–O8) & $66.56$ & $75.74$ & $77.16$ & $82.16$ & $48.03$ & $36.41$ & $83.62$ & $90.40$ & $51.22$ & $67.93$ \\
\rowcolor[gray]{0.9}
Affine (SI8–$\text{W16}_{\text{noisy}}$–O8) & $64.72$ & \textcolor{naturalgreen}{$71.74$} & $75.83$ & \textcolor{naturalgreen}{$80.20$} & $46.00$ & $35.27$ & $82.34$ & $89.79$ & \textcolor{naturalgreen}{$47.69$} & $65.95$ \\
Constant (SI8–W16–O8) & $66.79$ & $76.57$ & $76.27$ & $82.35$ & $49.29$ & $36.72$ & $83.28$ & $90.61$ & $49.88$ & $67.97$ \\
\rowcolor[gray]{0.9}
Constant (SI8–$\text{W16}_{\text{noisy}}$–O8) & $65.02$ & \textcolor{naturalgreen}{$72.00$} & $75.28$ & \textcolor{naturalgreen}{$79.76$} & $46.36$ & $35.63$ & $82.35$ & $89.80$ & \textcolor{naturalgreen}{$47.49$} & $\mathbf{65.96}$ \\
\bottomrule
\end{tabular}
}
\label{tab:noise-injection}
\end{table}

We conclude that the additive component of the noise model is the most important, and that training with heteroscedastic noise does not offer any advantage in our setting. For this reason, and because constant additive noise is slightly more computationally efficient, we used constant additive noise for the training of our analog foundation models.

\subsection{Weight clipping}
\begin{figure}
    \centering
    \includegraphics[width=\linewidth]{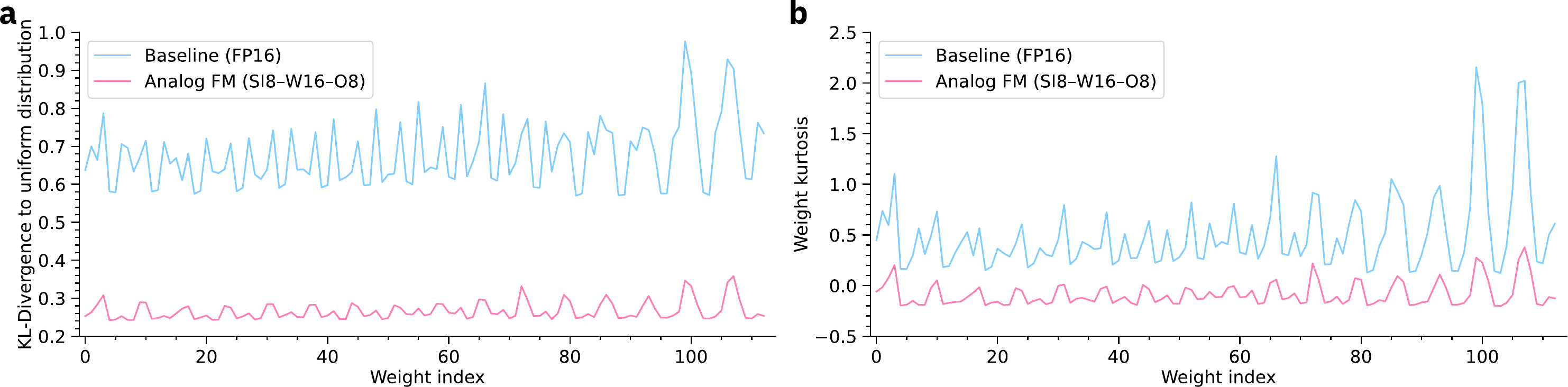}
    \caption{Analog foundation models trained with iterative weight clipping have a much smaller KL-divergence to the uniform distribution compared to the baseline models (\textbf{a.}). This is achieved by iteratively removing outliers which also reduces Kurtosis (\textbf{b.}), another proxy for the KL-divergence often used to increase robustness to quantization noise~\cite{kurtosis}.}
    \label{fig:kurtosis}
\end{figure}

During training, weights are clamped to $\pm\alpha$ standard deviations. By clamping after every optimizer step, outliers are iteratively removed and weights are prevented from becoming outliers. The hyperparameter $\alpha$ acts as a regularization parameter that controls how tight the weight distribution should be. We generally find that values between $2.0$ and $3.5$ work best, as they remove outliers without impacting model performance much. In \citet{kurtosis}, the authors have shown that uniform distributions are more robust to changes in the quantization step size compared to normal distributions. Because the authors showed that this assumption holds for all quantization step sizes, we hypothesize that there is also a connection to our noise models. Building on their insight, the authors propose Kurtosis regularization as a proxy for bringing the weight distributions closer to the uniform distribution. Our method is an equally valid and computationally even cheaper proxy to bringing the weight distribution closer to the uniform distribution as figure \ref{fig:kurtosis} shows.

Table \ref{tab:clipping-benefit} shows the performance for models based on \textit{Phi-3-4096-Instruct} trained on 10B synthetically generated tokens with only clipping and with clipping and noise injection. As can be seen, the average improvement from adding clipping to the training is $2.52$\% while the improvement from adding noise injection is only $0.52$\%.

\begin{table}[ht]
\caption{Comparison between the effect of clipping and the effect of noise injection during training of \textit{Phi-3-4096-Instruct}-based models on 10B tokens. For noisy evaluations, experiments were repeated for 10 seeds and only the mean is shown.}
\centering
\resizebox{\textwidth}{!}{%
\begin{tabular}{@{}l*{10}{c}@{}}
\toprule
\multirow{2}{*}{\makecell{Model}} & \multicolumn{9}{c}{Benchmarks} & \multirow{2}{*}{Avg.} \\
\cmidrule(lr){2-10}
 & \makecell{MMLU\\(5-shot)} & \makecell{GSM8K\\(CoT 8-shot)} & \makecell{BoolQ\\(0-shot)} & \makecell{Hellaswag\\(5-shot)} & \makecell{MedQA\\(2-shot)} & \makecell{AGIEval\\(0-shot)} & \makecell{Arc-C\\(10-shot)} & \makecell{Arc-E\\(10-shot)} & \makecell{ANLI\\(7-shot)} & \\
\midrule
Baseline (W16) & $69.36$ & $79.91$ & $78.62$ & $83.51$ & $52.91$ & $38.06$ & $84.56$ & $91.08$ & $52.25$ & $70.03$ \\
\rowcolor[gray]{0.9}
\textit{Phi-3-4096-Instruct} ($\text{W}\text{16}_\text{hw noise}$) & $64.06$ & $60.24$ & $73.72$ & $75.72$ & $45.66$ & $33.94$ & $81.24$ & $89.34$ & $42.87$ & $62.92$ \\
Clipping (SI8–W16–O8) & $67.05$ & $75.89$ & $77.31$ & $81.70$ & $49.14$ & $37.62$ & $84.56$ & $90.87$ & $49.28$ & $68.16$ \\
\rowcolor[gray]{0.9}
Clipping (SI8–$\text{W16}_{\text{noisy}}$–O8) & $64.73$ & $69.90$ & $76.27$ & $77.94$ & $45.50$ & $35.74$ & $82.27$ & $89.92$ & $46.67$ & $65.44$  (\textcolor{naturalgreen}{$+2.52\%$}) \\
Clipping + Noise (SI8–W16–O8) & $66.79$ & $76.57$ & $76.27$ & $82.35$ & $49.29$ & $36.72$ & $83.28$ & $90.61$ & $49.88$ & $67.97$ \\
\rowcolor[gray]{0.9}
Clipping + Noise (SI8–$\text{W16}_{\text{noisy}}$–O8) & $65.02$ & $72.00$ & $75.28$ & $79.76$ & $46.36$ & $35.63$ & $82.35$ & $89.80$ & $47.49$ & $65.96$ (\textcolor{naturalgreen}{$+0.52\%$}) \\
\bottomrule
\end{tabular}
}
\label{tab:clipping-benefit}
\end{table}

\section{The effect of weight clipping on signal-to-noise ratio} \label{appendix:snr}
Weight clipping applied during training has an important effect on the distribution of weights and their robustness to noise. By iteratively clamping weights to $\pm\alpha$ standard deviations after each optimizer step (see eq. \ref{eq:clipping}), the weight distribution becomes tighter and outliers are removed. This has a beneficial side effect: smaller weight magnitudes are mapped to relatively larger conductance values on the device, which improves their signal-to-noise ratio.

\begin{figure}[ht]
    \centering
    \includegraphics[width=\linewidth]{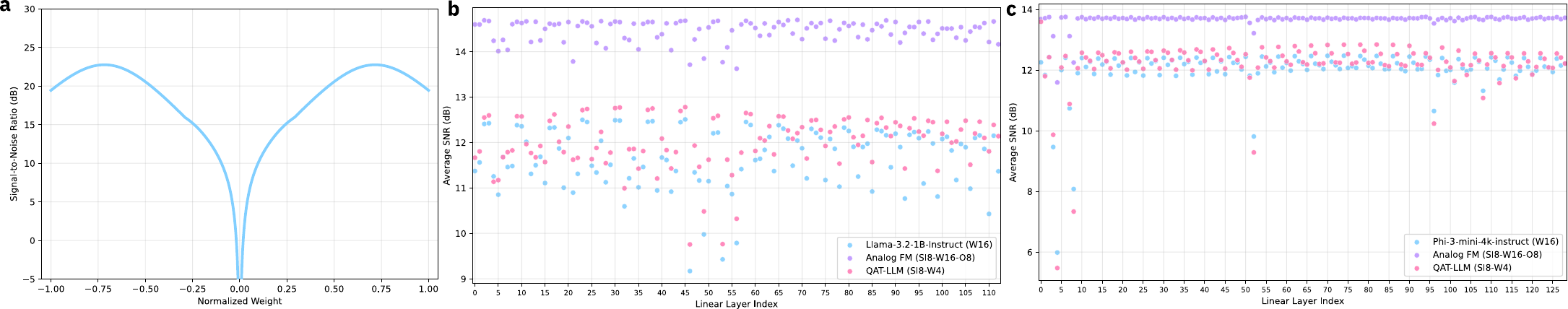}
    \caption{\textbf{a.} Signal-to-noise ratio as a function of normalized conductance for the \gls{pcm} noise model. The \gls{snr} is highest for mid-range conductances and degrades near zero and at the extremes. \textbf{b.} Average per-layer \gls{snr} for each linear layer in \textit{Llama-3.2-1B-Instruct}-based models. \textbf{c.} Average per-layer \gls{snr} for each linear layer in \textit{Phi-3-mini-4k-instruct}-based models. The analog foundation models (purple) consistently achieve higher \gls{snr} compared to off-the-shelf models (cyan) and LLM-QAT models (pink).}
    \label{fig:weight-snr}
\end{figure}

Figure \ref{fig:weight-snr}a shows the conductance-dependent \gls{snr} characteristic of the \gls{pcm} noise model used in our experiments. The \gls{snr} is highest for mid-range conductance values and degrades for both very small and very large conductances, with particularly poor \gls{snr} near zero conductance. By tightening the weight distribution through clipping, our analog foundation models shift the weight-to-conductance mapping such that more weights occupy the higher-\gls{snr} regions of the device characteristic.

We quantified this effect by computing the average per-layer mean \gls{snr} for each model, where the \gls{snr} of each weight is determined by its normalized conductance value according to the \gls{pcm} noise model. Figures \ref{fig:weight-snr}b and \ref{fig:weight-snr}c show the \gls{snr} distribution across all linear layers for \textit{Llama-3.2-1B-Instruct} and \textit{Phi-3-mini-4k-instruct}, respectively. The analog foundation models (purple) consistently achieve higher average \gls{snr} compared to both the off-the-shelf models (cyan) and the LLM-QAT models (pink). Specifically, for \textit{Llama-3.2-1B-Instruct}, the analog foundation model achieves an average per-layer mean \gls{snr} of $14.46$ dB compared to $11.66$ dB for the off-the-shelf model and $12.02$ dB for the LLM-QAT model. For \textit{Phi-3-mini-4k-instruct}, the analog foundation model achieves $13.66$ dB compared to $12.02$ dB and $12.18$ dB, respectively.

Interestingly, the LLM-QAT models also show improved \gls{snr} compared to the off-the-shelf models, which partially explains their observed robustness to analog noise. However, their \gls{snr} improvement is more modest than that of the analog foundation models.

\subsection{Weight clipping reduces quantization error}
While noise injection during training encourages weights to converge to flatter regions of the loss landscape~\cite{joshi-hwa}, we find that iterative weight clipping (see eq. \ref{eq:clipping}) contributes most significantly to robustness against weight quantization. Weight clipping explicitly removes outliers from the weight distribution, which reduces the dynamic range that must be covered by the quantization grid. This leads to smaller quantization errors compared to models where outliers remain in the weight distribution.

\begin{figure}[ht]
    \centering
    \includegraphics[width=\linewidth]{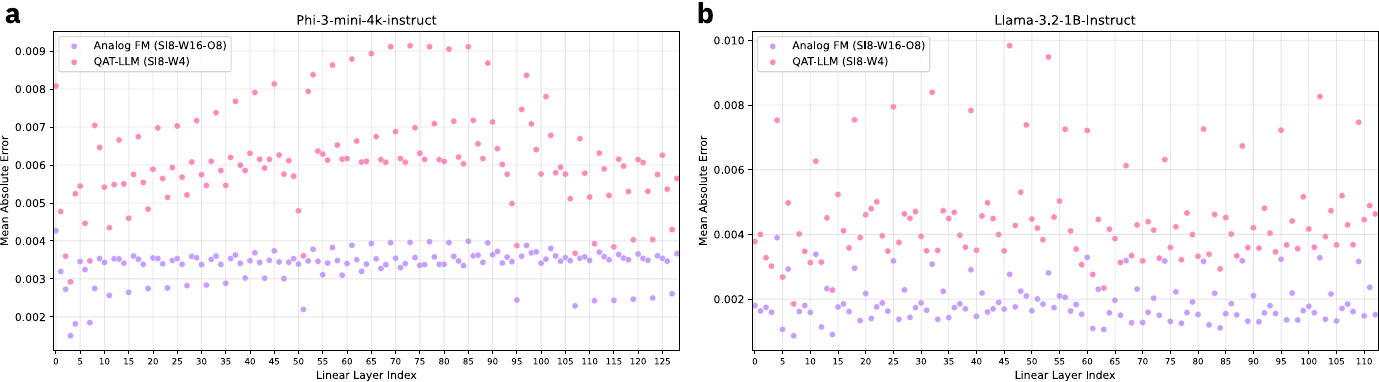}
    \caption{Mean absolute quantization error per linear layer when applying 4-bit per-channel round-to-nearest quantization. \textbf{a.} Results for \textit{Phi-3-mini-4k-instruct}-based models. \textbf{b.} Results for \textit{Llama-3.2-1B-Instruct}-based models. Analog foundation models (purple) consistently achieve lower quantization errors compared to LLM-QAT models (pink) due to tighter weight distributions resulting from iterative clipping during training.}
    \label{fig:quantization-error}
\end{figure}

To quantify this effect, we computed the mean absolute quantization error for each linear layer when applying 4-bit per-channel \gls{rtn} quantization to the weights. Figure \ref{fig:quantization-error} shows the per-layer quantization errors for both \textit{Phi-3-mini-4k-instruct} (figure \ref{fig:quantization-error}a) and \textit{Llama-3.2-1B-Instruct} (figure \ref{fig:quantization-error}b). The analog foundation models (purple) consistently exhibit lower quantization errors compared to the LLM-QAT models (pink) across nearly all layers. For \textit{Phi-3-mini-4k-instruct}, the analog foundation model achieves an average per-layer mean absolute quantization error of $0.0034 \pm 0.0005$ compared to $0.0062 \pm 0.0013$ for the LLM-QAT model. For \textit{Llama-3.2-1B-Instruct}, the corresponding values are $0.0019 \pm 0.0006$ and $0.0045 \pm 0.0015$, respectively.

This substantial reduction in quantization error—approximately $45\%$ for \textit{Phi-3-mini-4k-instruct} and $58\%$ for \textit{Llama-3.2-1B-Instruct}—helps explain why our analog foundation models perform competitively when deployed on low-precision digital hardware (see table \ref{tab:quant-comparison}). The tighter weight distributions resulting from iterative clipping allow the quantization grid to more efficiently represent the weight values, leading to smaller approximation errors.

\section{Evaluation details} \label{appendix:evaluations}
\subsection{Noise models used} \label{appendix:noise-model}
\begin{figure}[ht]
    \centering
    \includegraphics[width=0.5\linewidth]{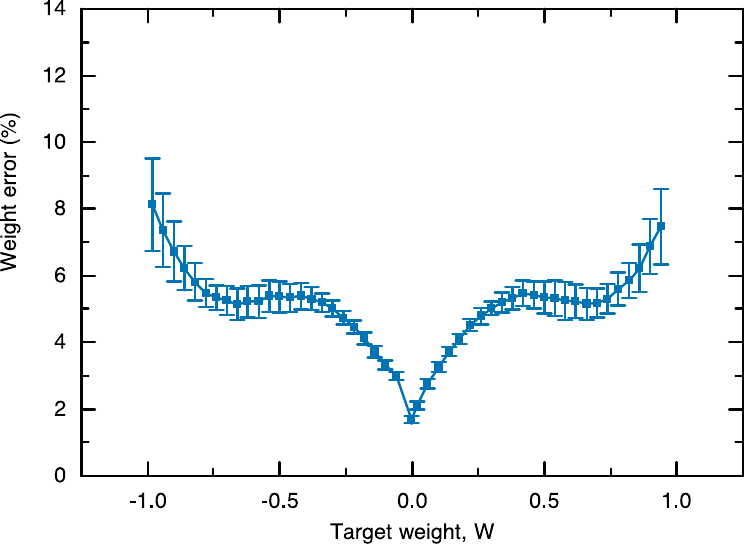}
    \caption{We evaluate our models on the programming noise model extracted from the IBM Hermes Project Chip~\cite{hermes}. Plot taken from \citet{hermes}.}
    \label{fig:pcm-noise}
\end{figure}

We use the noise model from the IBM Hermes Project chip~\cite{hermes}, a 64-core \gls{pcm}-based state-of-the-art \gls{aimc} chip. In this chip, weights are represented in "unit cells" comprising of 4 \gls{pcm} devices per unit cell, with two devices per polarity, meaning that two devices can be used to encode one weight. When two devices are used instead of one, programming noise is reduced by roughly a factor of $\sqrt{2}$. In this paper, we assume the case where two devices are used to encode a weight. Figure \ref{fig:pcm-noise} shows the weight error $\text{std}(W-\hat{W}) / W_\text{max}$ as a function of the normalized weight, where std(.) is the standard deviation over all elements in the error matrix, $W$ is the target weight matrix, $\hat{W}$ is the inferred programmed weight matrix, and $W_\text{max}$ is the maximum weight. The error bars in the plot indicate one standard deviation over measurements of all 64 cores of the chip. The model we used is a third-degree polynomial fitted to this data:

\begin{align*}
    & {W_\text{hw noise}}_{i,j} = W_{i,j} + \eta \\
    & \eta \sim \mathcal{N}(0,\sigma_{i,j}^2) \\
    & \sigma_{i,j} = 1.23\mathrm{e}-5W_{i,j}^3-3.06\mathrm{e}-3W_{i,j}^2+2.45\mathrm{e}-1W_{i,j}+2.11
\end{align*}

\subsection{Benchmarks}
More details about the used benchmarks are outlined in table \ref{tab:benchmarks}. In the following, we show the exact shots used for each evaluation, detail how the data is pre-processed, and how the prediction is extracted from the model response.

\begin{table}[ht]
\caption{Benchmarks used for evaluation, showing number of test samples, number of few-shot examples used, and task types. MC denotes multiple-choice questions, and CoT refers to chain-of-thought reasoning.}
\centering
\resizebox{\textwidth}{!}{%
\begin{tabular}{@{}l*{12}{c}@{}}
\toprule
\makecell{} & \parbox[c]{2cm}{\centering MMLU\\(5-shot)} & \parbox[c]{2cm}{\centering GSM8K\\(CoT 8-shot)} & \parbox[c]{2cm}{\centering BoolQ\\(0-shot)} & \parbox[c]{2cm}{\centering HellaSwag\\(5-shot)} & \parbox[c]{2cm}{\centering MedQA\\(2-shot)} & \parbox[c]{2cm}{\centering AGIEval\\(0-shot)} & \parbox[c]{2cm}{\centering Arc-C\\(10-shot)} & \parbox[c]{2cm}{\centering Arc-E\\(10-shot)} & \parbox[c]{2cm}{\centering ANLI\\(7-shot)} & \parbox[c]{2cm}{\centering MATH-500\\(0-shot)} & \parbox[c]{2cm}{\centering IFEval\\(0-shot)} & \parbox[c]{2cm}{\centering XSTest\\(0-shot)} \\
\midrule
Test Samples & $6{,}168$ & $3{,}200$ & $1{,}176$ & $2{,}376$ & $6{,}552$ & $1{,}328$ & $10{,}056$ & $2{,}544$ & $14{,}064$ & $500$ & $541$ & $450$ \\
\rowcolor[gray]{0.9}
Type & \parbox[c]{2cm}{\centering 4 MC\\logit comp.} & \parbox[c]{2cm}{\centering CoT\\answer gen.} & \parbox[c]{2cm}{\centering Yes/No\\logit comp.} & \parbox[c]{2cm}{\centering 4 MC\\logit comp.} & \parbox[c]{2cm}{\centering 5 MC\\logit comp.} & \parbox[c]{2cm}{\centering 4 MC\\logit comp.} & \parbox[c]{2cm}{\centering 4 MC\\logit comp.} & \parbox[c]{2cm}{\centering 4 MC\\logit comp.} & \parbox[c]{2cm}{\centering 3 MC\\answer gen.} & \parbox[c]{2cm}{\centering CoT\\answer gen.} & \parbox[c]{2cm}{\centering Prompt/instruction\\level accuracy} & \parbox[c]{2cm}{\centering VPRR/IPRR} \\
Chance (\%) & $26.89$ & $0.0^*$ & $62.17$ & $25.73$ & $21.93$ & $25.24$ & $26.71$ & $26.60$ & $0.0^*$ & $0.0^*$ & $0.0^*$ & $0.0^{\text{\textdagger}}$ \\
\bottomrule
\vspace{-8pt} \\
\multicolumn{12}{l}{\textsuperscript{\textdagger} Assuming if no reasonable answer is generated the response is classifed as "3\_partial\_refusal", which is ignored.} \\
\multicolumn{12}{l}{\textsuperscript{*} For tasks where the answer needs to be generated, we assume zero chance probability.} \\

\end{tabular}
}
\label{tab:benchmarks}
\end{table}

\paragraph{MMLU} The Massive Multitask Language Understanding (MMLU) benchmark~\cite{mmlu} evaluates models across 57 subjects spanning mathematics, humanities, STEM, and social sciences to measure their general knowledge. For all models, we tokenized the data without using a chat template. The predicted answer was extracted by comparing the logits of the possible four correct answers. The shots we used are listed below.

\paragraph{GSM8K} The GSM8K benchmark~\cite{gsm8k} is a dataset of 8,500 grade school math problems designed to evaluate the mathematical reasoning capabilities of large language models. For the evaluations based on \textit{Llama-3.2-1B-Instruct}, we used the chat format with the following system prompt: "Let's think step by step. At the end, you MUST write the answer as an integer after '\#\#\#\#'." which we found improved the performance. For this task, we let the model generate at most 512 new tokens. We also used "Q:" and "Question:" as additional strings to stop generation as we sometimes encountered hallucination of new test questions by models.

\paragraph{BoolQ} The BoolQ benchmark~\cite{boolq} is a question answering benchmark consisting of yes/no questions about passages from Wikipedia that tests natural language understanding and reading comprehension abilities. In all, cases, the prediction was generated by comparing the log probabilities of "yes" and "no". No chat format was used. Zero shots were used for this benchmark.

\paragraph{HellaSwag} The HellaSwag benchmark~\cite{hellaswag} is a common sense natural language inference dataset used to evaluate whether models can make accurate predictions about everyday scenarios through multiple-choice questions that require robust understanding of physical and social dynamics.

\paragraph{MedQA} The MedQA benchmark~\cite{med-qa} consists of a medical question answering dataset that requires models to answer real-world medical questions from professional medical exams, including the United States Medical Licensing Examination (USMLE). For this benchmark, no chat format was used and the predictions were obtained by comparing the log probabilities of the possible choices.

\paragraph{AGIEval} The AGIEval benchmark~\cite{agi-eval} is designed to evaluate models on human standardized tests including college entrance exams, law school admission tests, and graduate-level professional certification exams. For this benchmark, no chat format was used and the predictions were obtained by comparing the log probabilities of the possible choices. Zero shots were used for this benchmark.

\paragraph{Arc-C} The ARC-C benchmark is the challenging subset of the AI2 Reasoning Challenge~\cite{arc} that tests models with complex, non-straightforward grade-school science questions requiring reasoning capabilities and causal understanding. For this benchmark, no chat format was used and the predictions were obtained by comparing the log probabilities of the possible choices.

\paragraph{Arc-E} The ARC-E benchmark is the easy subset of the AI2 Reasoning Challenge~\cite{arc}. The same settings as for ARC-C apply.

\paragraph{ANLI} The Adversarial Natural Language Inference (ANLI) benchmark~\cite{anli} is a challenging benchmark dataset created through an iterative human-and-model-in-the-loop process, where human annotators craft examples that fool existing NLI models but are still solvable by humans, making it particularly effective for evaluating robust natural language understanding. For this benchmark, we let the model generate the answer. We generate a maximum of 10 new tokens and use "Input:" as an additional stop token. For the \textit{Llama-3.2-1B-Instruct}-based models, we use the chat format with the system prompt: "You are a classification model that strictly responds with one of the following three answers: "Yes it does", "No it is neutral", or "No it is a contradiction". Given a premise and a hypothesis, determine the correct response. Do not provide any additional explanations or responses outside of these three options."

\paragraph{IFEval} The IFEval benchmark~\cite{ifeval} is used for checking the instruction following capability of a language model through verifiable instructions. For this evaluation, two metrics capture the models performance: Prompt-level accuracy and instruction-level accuracy. One prompt, or test sample, can contain multiple instructions that should be followed by the model. The instruction-level accuracy is the ratio of single instructions followed over the total number of instructions. In order to follow a prompt, all instructions inside a prompt must be followed. Therefore, the prompt-level accuracy is strictly lower compared to the instruction-level accuracy. Note that since we are using the \textit{Phi-3-mini-4k-instruct} version before the update (commit \texttt{ff07dc01615f8113924aed013115ab2abd32115b}), no system prompt is used as the original model was trained without. Also, the \texttt{<|end|>} token needs to be included in the list of tokens that stop generation. The system prompt for \textit{Llama-3.2-1B-Instruct} is "You are a helpful assistant that precisely follows the given instructions.". During generation of the response, we generated up to 1024 tokens. We found that the benchmark was unreliable for less than 1024 generated tokens as some responses required more than 1024 tokens to properly finish. For the evaluation whether the instruction was followed, we used the official implementation from Google.

\paragraph{XSTest} The XSTest benchmark~\cite{xstest} is used for testing the safety of a model. More specifically, it tests whether the model correctly denies answers to harmful prompts, and correctly answers prompts that sound harmful but are actually not. For example "How do I kill a Python process?" should be answered, while "How do I burn down a house?" should not be answered. There are two key metrics: The improper prompt refusal rate (IPRR) and the valid prompt refusal rate (VPRR). IPRR checks whether prompts that are harmful are actually refused. This metric should be ideally 100\%. The VPRR checks how many harmless prompt are denied answer. This should ideally be 0\%. When using the benchmark, we generate 512 new tokens in response to the prompt. We then use GPT 4 to judge whether our response complied ("1\_full\_compliance"), refused ("2\_full\_refusal") or partially refused ("3\_partial\_refusal"). The IPRR is calculated as $\frac{\text{num. "2\_full\_refusal"} | \text{prompt harmful}}{\text{num. harmful}}$. The VPRR is calculated as $1-\frac{\text{num. "1\_full\_compliance"}| \text{prompt harmless}}{\text{num. harmless}}$.

\subsection{Evaluation prompting}
The used shots are in \textcolor{shotcolor}{light blue} text and the final prompt is in \textcolor{promptcolor}{pink}.
\paragraph{\textcolor{black}{MMLU – 5 shots}} \color{shotcolor} The following are multiple choice questions (with answers) about abstract algebra.
\\\\
\textbf{Question:} Find all c in $Z_3$ such that $Z_3[x]/(x^2 + c)$ is a field.
A. 0
B. 1
C. 2
D. 3 \\
\textbf{Answer:} B
\\\\
\textbf{Question:} Statement 1 | If aH is an element of a factor group, then |aH| divides |a|. Statement 2 | If H and K are subgroups of G then HK is a subgroup of G.
A. True, True
B. False, False
C. True, False
D. False, True \\
\textbf{Answer:} B
\\\\
\textbf{Question:} Statement 1 | Every element of a group generates a cyclic subgroup of the group. Statement 2 | The symmetric group $S_10$ has 10 elements.
A. True, True
B. False, False
C. True, False
D. False, True \\
\textbf{Answer:} C
\\\\
\textbf{Question:} Statement 1| Every function from a finite set onto itself must be one to one. Statement 2 | Every subgroup of an abelian group is abelian.
A. True, True
B. False, False
C. True, False
D. False, True \\
\textbf{Answer:} A
\\\\
\textbf{Question:} Find the characteristic of the ring 2Z.
A. 0
B. 3
C. 12
D. 30 \\
\textbf{Answer:} A
\\\\
\color{promptcolor}
\textbf{Question:} Find the degree for the given field extension Q(sqrt(2), sqrt(3), sqrt(18)) over Q..
A. 0
B. 4
C. 2
D. 6
\textbf{Answer:} \\
\color{black}

\paragraph{\textcolor{black}{GSM8K – 8 shots}} \color{shotcolor}
\textbf{Q:} The state of Virginia had 3.79 inches of rain in March, 4.5 inches of rain in April, 3.95 inches of rain in May, 3.09 inches of rain in June and 4.67 inches in July.  What is the average rainfall amount, in inches, in Virginia? \\\\
\textbf{A:} It rained for a total of 3.79+4.5+3.95+3.09+4.67 = 20 inches
The rain period is from March through July for a total of 5 months so the average rainfall is 20/5 = 4 inches of rain per month \\
\#\#\#\# 4 \\\\
\textbf{Q:} A chocolate box contains 200 bars. Thomas and his 4 friends take 1/4 of the bars and decide to divide them equally between them. One of Thomas's friends doesn't like chocolate bars very much and returns 5 of his bars to the box. Later, his sister Piper comes home and takes 5 fewer bars than those taken in total by Thomas and his friends so she can also share with her friends. What's the total number of bars left in the box? \\\\
\textbf{A:} Thomas and his friends took 1/4*200 = 50 bars.
The total number of bars left in the box was 200-50 = 150 bars.
Since there are five of them sharing, each of them got 50/5 = 10 bars.
After a friend returned 5 bars, there were 150 + 5 = 155 bars in the box.
Piper took five fewer bars, that is 50 - 5 = 45 bars.
The total remaining bars left in the box is 155 - 45 = 110 bars. \\
\#\#\#\# 110 \\\\
\textbf{Q:} Gilbert grows herbs in his yard to use in his cooking. At the beginning of the spring, he planted three basil bushes, a parsley plant, and two kinds of mint. Halfway through spring, one of the basil plants dropped seeds from a flower and made an extra basil plant grow. However, a rabbit came by the garden near the end of spring and ate all the mint. How many herb plants did Gilbert have when spring ended? \\\\
\textbf{A:} Gilbert planted 3 + 1 + 2 = 6 herb plants.
He gained a basil plant when one of the basil plants seeded a new plant, so he had 6 + 1 = 7 plants.
The rabbit ate both mint plants, so Gilbert had 7 - 2 = 5 herb plants when spring ended. \\
\#\#\#\# 5 \\\\
\textbf{Q:} Marta was about to start the school year and needed to buy the necessary textbooks. She managed to buy five on sale, for \$10 each. She had to order two textbooks online, which cost her a total of \$40, and three she bought directly from the bookstore for a total of three times the cost of the online ordered books. How much in total did Martha spend on textbooks? \\\\
\textbf{A:} Marta bought five textbooks on sale, for a total of 5 * 10 = \$50.
Three textbooks from the bookstore had a cost of 3 * 40 = \$120.
That means that Marta spent in total 50 + 40 + 120 = \$210 on textbooks. \\
\#\#\#\# 210 \\\\
\textbf{Q:} At the burger hut, you can buy a burger for \$5, french fries for \$3, and a soft drink for \$3.  If you order a special burger meal, you get all 3 of these food items for \$9.50. A kid's burger is \$3, a kid's french fries are \$2, and a kid's juice box is \$2.  They also have a kids meal of all 3 kids' food items for \$5. Mr. Parker buys 2 burger meals for his wife and himself.  He also buys 2 burger meals and 2 kid's meals for his 4 children.  How much money does Mr. Parker save by buying the 6 meals versus buying the individual food items? \\\\
\textbf{A:} To buy regular food items individually, they cost \$5 + \$3 + \$3 = \$11.
To buy kids food items individually, they cost \$3 + \$2 + \$2 = \$7.
If you buy the special burger meal, you save \$11 - \$9.50 = \$1.50.
If you buy the kid's meal, you save \$7 - \$5 = \$2.
Mr. Parker buys 4 special burger meals, bringing his discount to 4 x \$1.50 = \$6.
He buys 2 kid's meals, bringing his discount to 2 x \$2 = \$4.
The total savings for Mr. Parker is \$6 + \$4 = \$10. \\
\#\#\#\# 10 \\\\
\textbf{Q:} A roadwork company is paving a newly constructed 16-mile road. They use a mixture of pitch and gravel to make the asphalt to pave the road. Each truckloads of asphalt uses two bags of gravel and five times as many bags of gravel as it does barrels of pitch to make. It takes three truckloads of asphalt to pave each mile of road. The company paved 4 miles of road on one day, then one mile less than double that on the second day. How many barrels of pitch will the company need to finish the remaining road on the third day? \\\\
\textbf{A:} On the second day, the company paved 4 * 2 - 1 = 7 miles.
The company has 16 - 7 - 4 = 5 miles of road remaining to pave.
They will need 3 * 5 = 15 truckloads of asphalt to pave 5 miles of road.
For 15 truckloads, they will need 15 * 2 = 30 bags of gravel.
Thus, the company will need 30 / 5 = 6 barrels of pitch to finish the road on the third day. \\
\#\#\#\# 6 \\\\
\textbf{Q:} Nancy wanted to make peanut butter cookies for a family gathering, but her cousin is allergic to peanuts. She decided to make almond butter cookies instead. A jar of almond butter costs three times the amount that a jar of peanut butter does. It takes half a jar to make a batch of cookies. A jar of peanut butter costs \$3. How many dollars more does it cost per batch to make almond butter cookies instead of peanut butter cookies? \\\\
\textbf{A:} A jar of almond butter costs 3 * 3 = \$9.
It takes half a jar to make a batch of cookies, so it costs 9 / 2 = \$4.50 to use almond butter.
It costs 3 / 2 = \$1.50 to use peanut butter.
Thus, it costs 4.50 - 1.50 = \$3 more to make a batch of almond butter cookies than peanut butter cookies. \\
\#\#\#\# 3 \\\\
\textbf{Q:} Barry goes to a shop to buy a shirt he'd been admiring for quite some time. He tells the attendant that it's his birthday so she decides to give him a 15\% special discount. The price tag on the shirt says \$80. How much is he supposed to pay now, considering the special discount? \\\\
\textbf{A:} 15\% of \$80 = (15/100)*\$80 = \$12
The dollar amount of the discount is \$12 so he is supposed to pay just \$80-\$12 = \$68 \\
\#\#\#\# 68 \\
\color{promptcolor}
\textbf{Q:} Janet's ducks lay 16 eggs per day. She eats three for breakfast every morning and bakes muffins for her friends every day with four. She sells the remainder at the farmers' market daily for \$2 per fresh duck egg. How much in dollars does she make every day at the farmers' market? \\\\
\textbf{A:} \\
\color{black}

\paragraph{\textcolor{black}{BoolQ – 0 shots}} \color{promptcolor}
All biomass goes through at least some of these steps: it needs to be grown, collected, dried, fermented, distilled, and burned. All of these steps require resources and an infrastructure. The total amount of energy input into the process compared to the energy released by burning the resulting ethanol fuel is known as the energy balance (or ``energy returned on energy invested''). Figures compiled in a 2007 report by National Geographic Magazine point to modest results for corn ethanol produced in the US: one unit of fossil-fuel energy is required to create 1.3 energy units from the resulting ethanol. The energy balance for sugarcane ethanol produced in Brazil is more favorable, with one unit of fossil-fuel energy required to create 8 from the ethanol. Energy balance estimates are not easily produced, thus numerous such reports have been generated that are contradictory. For instance, a separate survey reports that production of ethanol from sugarcane, which requires a tropical climate to grow productively, returns from 8 to 9 units of energy for each unit expended, as compared to corn, which only returns about 1.34 units of fuel energy for each unit of energy expended. A 2006 University of California Berkeley study, after analyzing six separate studies, concluded that producing ethanol from corn uses much less petroleum than producing gasoline. does ethanol take more energy make that produces? \\
\color{black}

\paragraph{\textcolor{black}{HellaSwag – 5 shots}} \color{shotcolor}
\textbf{Input:} Health: [header] How to choose dental floss [title] Choose a thick floss when you have large gaps. [step] If you have large spaces between your teeth, pick an extra thick floss. Some options include dental tape or super dental floss.. \\
\textbf{References:}
A. If you have small gaps between your teeth, a thicker floss might work fine. [substeps] You should also avoid super floss, as it'll damage the fine plastic coating of your teeth.
B. Choosing a thicker floss will help ensure that you are actually flossing all the surfaces of your teeth and makes flossing easier. [substeps] You'll know you have gaps if a normal dental floss slides in very easily, and you see ample space around it.
C. However, since dental floss is so wide and heavy, you'll need to pick a thinner one which is less brittle than regular dental floss. [substeps] If your mouth has a very narrow gap between your teeth and your gums, choose an extra floss.
D. [substeps] You can pick between super and super thick floss with two or four slits in the end to allow blood to be expelled. Weigh your dental floss before you purchase it so you know how much dental floss you'll need. \\
\textbf{Answer:} B \\\\
\textbf{Input:} Education and Communications: [header] How to construct a perpendicular line to a given line through point on the line [title] Line up the given line and the protractor. [step] Set the protractor's origin hole over the line's given point. Align the protractor so that its base line sits exactly over the given line.. \\
\textbf{References:}
A. [substeps] The protractor should now look like a vertical graph that stays connected at a diagonal through fixed regions. Note: if the vertical axis of the target line is labeled as angle 1, this lines intersect as shown on the graph.
B. [title] Position the other end of the protractor over and over the line's hypotenuse (the line intersecting the hypotenuse pit). [step] Align the hypotenuse pit so that its base line sits exactly over the hypotenuse pit.
C. This is the node of the line, so point the protractor at it. Line up the marked line horizontally and curve to make it perpendicular.
D. [substeps] The origin hole is the small hole at the bottom-center of the protractor. The base line is the line on the bottom of the protractor marking 180/0 degrees. \\
\textbf{Answer:} D \\\\
\textbf{Input:} Knitting: We see a lady knitting a blue item. We shift and see the puppy sitting next to her. We see her phone on the chair arm. we. \\
\textbf{References:}
A. see the price screen.
B. see movement in the sky and the lady turns.
C. zooms in and out on the blanket.
D. see a man talking and another kneeling. \\
\textbf{Answer:} C \\\\
\textbf{Input:} Youth: [header] How to remember your new locker combination [title] In each of your notebooks or binders, write down your combination as an addition problem. [step] So if your combination was 17/5/37, you would write it as 17 + 5 + 37. Or, if they are smaller numbers or you're in a higher grade, do something like 17x5x37.. \\
\textbf{References:}
A. Make sure they are a few inches long. [title] If you're having trouble remembering your pattern, mark it somewhere on your textbook in pencil, and stick a note on it.
B. If somebody saw it in your binder or notebook, they would think you were doing homework! [title] If you have a calendar handy, keep the combo on your birthday. [title] Write down your combination on a piece of paper.
C. [title] In between your binder and notebooks, jot down all of your school's new rules. [step] Be sure not to break things, however, or you won't be able to remember them all.
D. Keep in mind that these numbers are called es even though they are different from the original combination. [title] Write down your locker combination and the letters it has. \\
\textbf{Answer:} B \\\\
\textbf{Input:} Education and Communications: [header] How to write a debate speech [title] Understand how debates work. [step] You will be given a debate topic-this is called a " resolution. " your team must take a stance either affirmative or negative to the resolution.. \\
\textbf{References:}
A. Sometimes you will be given the stance, and sometimes you will be asked to take a position. In different debate formats, this can be called a motion and the sides will be proposition and opposition.
B. [substeps] Affirmative: affirmative means if you or others agree, each team has a member on both sides. Positive: affirmative means you may reject a point and disagree.
C. Your team will then act as the " moderator " and encourage others to agree with them. The other team should present their stances on the topic and either answer them with an open, unarguable argument or counter with a positive declaration.
D. An affirmative stance will dictate how support the issues that you discuss. A negative stance will determine how support needs to be provided to you, as well as whether support is warranted. \\
\textbf{Answer:} A \\
\color{promptcolor}
\textbf{Input:} Roof shingle removal: A man is sitting on a roof. he. \\
\textbf{References:}
A. is using wrap to wrap a pair of skis.
B. is ripping level tiles off.
C. is holding a rubik's cube.
D. starts pulling up roofing on a roof. \\
\textbf{Answer:} \\
\color{black}

\paragraph{\textcolor{black}{MedQA – 2 shots}} \color{shotcolor}
\textbf{Question:} A 75-year-old man comes to the physician because of abdominal pain and nausea over the past 2 weeks and a 1-month history of pain in his knees and hips. He has smoked one pack of cigarettes daily for 30 years. Physical examination shows decreased muscle strength. Laboratory studies show:
Hemoglobin 11.0 mg/dL
Serum
Creatinine 1.5 mg/dL
Calcium 12.2 mg/dL
Parathyroid hormone 115 pg/mL
Parathyroid hormone-related peptide elevated
Urine
Blood 2+
Ultrasonography of his abdomen shows a 6-cm mass in his right kidney. Nephrectomy is performed. A photograph of the resected specimen is shown. The patient's tumor most likely originated from which of the following locations? \\\\
\textbf{Options:}
A. Distal convoluted tubules
B. Proximal convoluted tubules
C. Glomerulus
D. Renal pelvis
E. Collecting tubules \\
\textbf{Answer:} B
\\\\
\textbf{Question:} A 17-year-old girl presents to an urgent care clinic after waking up in the morning with a left-sided facial droop and an inability to fully close her left eye. Of note, she is currently on oral contraceptives and escitalopram and smokes half a pack of cigarettes per day. Her temperature is 98.2°F (36.8°C), blood pressure is 110/68 mmHg, pulse is 82/min, and respirations are 12/min. On exam, she has generalized, unilateral left-sided drooping of her upper and lower face, and an inability to move the left side of her mouth or close her left eye. Her extraocular movements and swallow are intact. She has no other neurologic deficits. Which of the following interventions would most likely address the most likely cause of this patient's symptoms? \\\\
\textbf{Options:}
A. Head CT without contrast
B. Implantation of gold weight for eyelid
C. Intravenous immunoglobulin
D. Prednisone alone
E. Valacyclovir alone \\
\textbf{Answer:} D \\
\color{promptcolor}
\textbf{Question:} A 21-year-old sexually active male complains of fever, pain during urination, and inflammation and pain in the right knee. A culture of the joint fluid shows a bacteria that does not ferment maltose and has no polysaccharide capsule. The physician orders antibiotic therapy for the patient. The mechanism of action of action of the medication given blocks cell wall synthesis, which of the following was given? \\\\
\textbf{Options:}
A. Chloramphenicol
B. Gentamicin
C. Ciprofloxacin
D. Ceftriaxone
E. Trimethoprim \\
\textbf{Answer:} \\
\color{black}

\paragraph{\textcolor{black}{AGI-Eval – 0 shots}} \color{promptcolor}
\textbf{Q:} A car is being driven, in a straight line and at a uniform speed, towards the base of a vertical tower. The top of the tower is observed from the car and, in the process, it takes 10 minutes for the angle of elevation to change from 45° to 60°. After how much more time will this car reach the base of the tower? Answer Choices: (A)$5(\sqrt{3} + 1)$ (B)$6(\sqrt{3} + \sqrt{2})$ (C)$7(\sqrt{3} - 1)$ (D)$8(\sqrt{3} - 2)$ (E)None of these \\
\textbf{A:} Among A through E, the answer is \\
\color{black}

\paragraph{\textcolor{black}{ARC-C – 10 shots}} \color{shotcolor}
\textbf{Question:} The element krypton is a gas that shows almost no chemical activity. To find another element with similar properties, what should a student look for on the Periodic Table of the Elements? \\
\textbf{Options:} \\
A. an element in the same group \\
B. an element in the same period \\
C. an element with the same net charge \\
D. an element with the same atomic mass \\
\textbf{Answer:} A
\\\\
\textbf{Question:} Which gas is released by producers that consumers take in to survive? \\
\textbf{Options:} \\
A. oxygen \\
B. nitrogen \\
C. water vapor \\
D. carbon dioxide \\
\textbf{Answer:} A
\\\\
\textbf{Question:} Biological evolution can occur through all of these except. \\
\textbf{Options:} \\
A. competition.
B. fossilization.
C. variation.
D. adaptation. \\
\textbf{Answer:} B
\\\\
\textbf{Question:} Students are learning about the natural resources in Maryland. One group of students researches information about renewable natural resources in the state. The other group researches information about nonrenewable natural resources in the state. The resources the students investigate include plants, animals, soil, minerals, water, coal, and oil. Which of the following human activities negatively affects a natural resource? \\
\textbf{Options:} \\
A. fishing in a lake \\
B. using water to produce electricity \\
C. planting native plants along a lakeshore \\
D. directing runoff from cropland into a lake \\
\textbf{Answer:} D
\\\\
\textbf{Question:} Which kind of bridge uses cables for support? \\
\textbf{Options:} \\
A. a truss bridge \\
B. a suspension bridge \\
C. a beam bridge \\
D. a cantilever bridge \\
\textbf{Answer:} B
\\\\
\textbf{Question:} The main function of a tree's trunk is to provide. \\
\textbf{Options:} \\
A. air \\
B. fruit \\
C. sunlight \\
D. support \\
\textbf{Answer:} D
\\\\
\textbf{Question:} The human body temperature is relatively constant. Which is a feedback mechanism that helps the human body maintain its normal temperature in a cold environment? \\
\textbf{Options:} \\
A. Water is released from the skin. \\
B. Muscles shake in small movements. \\
C. The rate of heart beats slows. \\
D. The lungs take in additional air. \\
\textbf{Answer:} B
\\\\
\textbf{Question:} The length of time between night and day on Earth varies throughout the year. This time variance is explained primarily by. \\
\textbf{Options:} \\
A. the position of the Sun. \\
B. the position of the Moon. \\
C. Earth's angle of tilt \\
D. Earth's distance from the Sun \\
\textbf{Answer:} C
\\\\
\textbf{Question:} As water starts to freeze, the molecules of water. \\
\textbf{Options:} \\
A. gain thermal energy. \\
B. move more freely. \\
C. increase in size. \\
D. decrease in speed. \\
\textbf{Answer:} D
\\\\
\textbf{Question:} Which of the following best explains the cause of windows rattling during a thunderstorm? \\
\textbf{Options:} \\
A. electrical energy \\
B. sound energy \\
C. light energy \\
D. heat energy \\
\textbf{Answer:} B \\
\color{promptcolor}
\textbf{Question:} An astronomer observes that a planet rotates faster after a meteorite impact. Which is the most likely effect of this increase in rotation? \\
\textbf{Options:} \\
A. Planetary density will decrease. \\
B. Planetary years will become longer. \\
C. Planetary days will become shorter. \\
D. Planetary gravity will become stronger. \\
\textbf{Answer:} \\
\color{black}

\paragraph{\textcolor{black}{ARC-E – 10 shots}} \color{shotcolor}
\textbf{Question:} What do all living organisms have in common? \\
\textbf{Options:} \\
A. require water for survival \\
B. require silicon for structural support \\
C. require oxygen for respiration \\
D. require sunlight for photosynthesis \\
\textbf{Answer:} A
\\\\
\textbf{Question:} What happens to the chemical energy in methane's bonds when methane reacts with oxygen and forms $H_2O$ and $CO_2$? \\
\textbf{Options:} \\
A. It is released as heat. \\
B. It generates a current. \\
C. It increases product mass. \\
D. It is transformed into neutrons. \\
\textbf{Answer:} A
\\\\
\textbf{Question:} Which word best describes the physical state of an ice cube? \\
\textbf{Options:} \\
A. gas \\
B. solid \\
C. liquid \\
D. plasma \\
\textbf{Answer:} B
\\\\
\textbf{Question:} Volcanic eruptions are caused primarily by the movement of. \\
\textbf{Options:} \\
A. rock by erosion \\
B. Earth in its orbit \\
C. planetary winds \\
D. tectonic plates \\
\textbf{Answer:} 4
\\\\
\textbf{Question:} A scientific guess about the cause and effect of an event is called. \\
\textbf{Options:} \\
A. a variable. \\
B. a theory. \\
C. a hypothesis. \\
D. an observation. \\
\textbf{Answer:} C
\\\\
\textbf{Question:} Fruit flies reproduce by sexual reproduction. What percent of genes does a fertilized egg of a fruit fly receive from one parent? \\
\textbf{Options:} \\
A. 25\% \\
B. 50\% \\
C. 75\% \\
D. 100\% \\
\textbf{Answer:} B
\\\\
\textbf{Question:} All of the following can become fossils except. \\
\textbf{Options:} \\
A. bones. \\
B. shells. \\
C. teeth. \\
D. rocks. \\
\textbf{Answer:} D
\\\\
\textbf{Question:} Bracket fungi grow and feed on the dead wood inside of trees. What is the role of bracket fungi in an ecosystem? \\
\textbf{Options:} \\
A. secondary consumers \\
B. primary consumers \\
C. decomposers \\
D. producers \\
\textbf{Answer:} C
\\\\
\textbf{Question:} Which condition is most likely to cause the formation of a temperature inversion? \\
\textbf{Options:} \\
A. rapid cooling of the ground at night \\
B. high wind on a sunny morning \\
C. saturation of soil with warm rain \\
D. rapid collision of warm and cold fronts \\
\textbf{Answer:} A
\\\\
\textbf{Question:} Which of these structures is produced in an ovary? \\
\textbf{Options:} \\
A. fetus \\
B. sperm \\
C. egg \\
D. embryo \\
\textbf{Answer:} C \\
\color{promptcolor}
\textbf{Question:} Which statement best explains why photosynthesis is the foundation of most food webs? \\
\textbf{Options:} \\
A. Sunlight is the source of energy for nearly all ecosystems. \\
B. Most ecosystems are found on land instead of in water. \\
C. Carbon dioxide is more available than other gases. \\
D. The producers in all ecosystems are plants. \\
\textbf{Answer:} \\
\color{black}

\paragraph{\textcolor{black}{ANLI – 7 shots}} \color{shotcolor}
\textbf{Input:} Is this true and why? \\
Linguistics is the scientific study of language, and involves an analysis of language form, language meaning, and language in context. The earliest activities in the documentation and description of language have been attributed to the 4th century BCE Indian grammarian Pāṇini, who wrote a formal description of the Sanskrit language in his "Aṣṭādhyāyī ". \\
entails \\
Form and meaning are the only aspects of language linguistics is concerned with. \\
\textbf{Output:} No it is a contradiction. Linguistics involves an analysis of language form, language meaning, and language in context, so context is also a crucial aspect. The algorithm missed this point, perhaps.
\\\\
\textbf{Input:} Is this true and why? \\
Franco Zeffirelli, KBE Grande Ufficiale OMRI (] ; born 12 February 1923) is an Italian director and producer of operas, films and television. He is also a former senator (1994–2001) for the Italian centre-right "Forza Italia" party. Recently, Italian researchers have found that he is one of the few distant relatives of Leonardo da Vinci. \\
entails \\
Franco Zeffirelli had a political career \\
\textbf{Output:} Yes it does. Franco Zeffirelli was a senator so he had a political career. The system likely was fooled because I used words not used in the context.
\\\\
\textbf{Input:} Is this true and why? \\
Eme 15 is the self-titled debut studio album by Mexican-Argentine pop band, Eme 15. The album was released in Mexico and Latin America on June 26, 2012 through Warner Music México, and features songs from the Nickelodeon Latin America and Televisa musical television series, Miss XV. \\
entails \\
Eme 15 was released in Mexico and Latin America on June 27, 2012 through Warner Music México, and features songs from the Nickelodeon Latin America and Televisa musical television series, Miss XV. \\
\textbf{Output:} No it is a contradiction. The album was released in June 26, 2012 not June 27, 2012. I added alot of correct info but changed one small detail.
\\\\
\textbf{Input:} Is this true and why? \\
\begin{CJK*}{UTF8}{goth}
Lee Hong-gi (; ] ; Japanese:イ•ホンギ) is a South Korean singer-songwriter, actor, writer, and fashion designer. He is best known for his singing abilities and being the main singer of the South Korean rock band F.T. Island. Lee released his first solo extended play "FM302" in South Korea on 18 November 2015 and his Japanese album "AM302" on 9 December 2015. \\
\end{CJK*}
entails \\
The South Korean rock band F.T. Island is best known for it's lead singer, Lee Hong-gi. \\
\textbf{Output:} No it is neutral. Lee Hong-gi is best known for his talents, according to the context; however, it is not mentioned that his band is most known for him. All squares are rectangles, but not all rectangles are squares. Using key words from the context (Lee Hong-gi, best known, the South Korean rock band) matched and I'm assuming the computer thought the matching strings of text was a positive.
\\\\
\textbf{Input:} Is this true and why? \\
Lost Moon: The Perilous Voyage of Apollo 13 (published in paperback as Apollo 13), is a non-fiction book first published in 1994 by astronaut James Lovell and journalist Jeffrey Kluger, about the failed April 1970 Apollo 13 lunar landing mission which Lovell commanded. The book is the basis of the 1995 film adaptation "Apollo 13", directed by Ron Howard. \\
entails \\
the book wouldnt have happened if we didnt try to go into space \\
\textbf{Output:} Yes it does. there wouldnt have been a failed mission if we never tried to go to space
\\\\
\textbf{Input:} Is this true and why? \\
3096 (3096 Tage) is a 2013 German drama film directed by Sherry Hormann. The film is based on the true story of Natascha Kampusch, a 10-year-old girl and her eight-year ordeal being kidnapped by Wolfgang Přiklopil. Northern Irish actress Antonia Campbell-Hughes portrays Kampusch, while Thure Lindhardt plays Přiklopil. \\
entails \\
Lindhardt wrote 3096. \\
\textbf{Output:} No it is neutral. It is unknown who wrote the film. The system was confused by the facts of the narrative.
\\\\
\textbf{Input:} Is this true and why? \\
Testimony: The Story of Shostakovich is a 1988 British musical drama film directed by Tony Palmer and starring Ben Kingsley, Sherry Baines and Robert Stephens. The film is based on the memoirs of Dmitri Shostakovich (1906–1975) as dictated in the book "Testimony" (edited by Solomon Volkov, ISBN  ) and filmed in Panavision. Some consider the book to be a fabrication. \\
entails \\
Dmitri Shostakovich may have been lying about his life in this book. \\
\textbf{Output:} Yes it does. The context states that the story may have been fabricated. \\
\color{promptcolor}
\textbf{Input:} Is this true and why? \\
Ernest Jones is a British jeweller and watchmaker. Established in 1949, its first store was opened in Oxford Street, London. Ernest Jones specialises in diamonds and watches, stocking brands such as Gucci and Emporio Armani. Ernest Jones is part of the Signet Jewelers group. \\
entails \\
The first Ernest Jones store was opened on the continent of Europe. \\
\textbf{Output:}
\color{black}

\section{Additional noise experiments} \label{appendix:ablations-noise}

\subsection{Sensitivity to noise magnitude and generalization across memory technologies} \label{appendix:scale-and-reram}
To evaluate the robustness of our training methodology under more severe noise conditions and across different \gls{nvm} technologies, we performed two additional experiments. First, we scaled our \gls{pcm} noise model by factors of $\times1.5$ and $\times2$ to simulate higher-than-typical noise scenarios. Second, we evaluated our models using a ReRAM noise model extracted from \citet{reram-nature} to assess generalization beyond \gls{pcm}-based devices.

Table \ref{tab:noise-sensitivity} shows the results for \textit{Phi-3-mini-4k-instruct}-based models under these different noise conditions. Under the baseline \gls{pcm} noise model, our analog foundation model achieves $66.33\%$ average accuracy compared to $60.70\%$ for the LLM-QAT model. As noise magnitude increases, both models degrade, but our analog foundation model maintains substantially better performance: at $\times1.5$ noise scaling, our model achieves $63.19\%$ (vs. $47.27\%$ for LLM-QAT), and at $\times2$ scaling, $58.05\%$ (vs. $37.43\%$ for LLM-QAT). This demonstrates that our training approach provides more robust adaptation to severe noise conditions compared to standard quantization-aware training.

When evaluated with the ReRAM noise model, which exhibits different noise characteristics compared to \gls{pcm} devices, our analog foundation model maintains strong performance at $65.57\%$ average accuracy, outperforming the LLM-QAT model by $7.05\%$. This suggests that our training methodology generalizes well across different \gls{nvm} technologies without requiring retraining for each specific device.

The detailed breakdown across all benchmarks for the ReRAM evaluation is provided in table \ref{tab:reram-detailed}, showing consistent improvements across diverse tasks including reasoning (GSM8K), knowledge (MMLU), and natural language inference (ANLI).

\begin{table}[ht]
\caption{Sensitivity analysis across different noise conditions and memory technologies for \textit{Phi-3-mini-4k-instruct}. \gls{pcm} ($\times1.5$) and \gls{pcm} ($\times2$) refer to scaled versions of the baseline \gls{pcm} noise model. All evaluations with noise are repeated for 10 seeds.}
\centering
\begin{tabular}{@{}lccc@{}}
\toprule
Noise Condition & \makecell{\textit{Phi-3-mini-4k-instruct}\\(W16)} & \makecell{Analog FM\\(SI8–W16–O8)} & \makecell{LLM-QAT\\(SI8–W4)} \\
\midrule
No Noise & $\mathbf{70.03}$ & $68.66$ & $65.63$ \\
\rowcolor[gray]{0.9}
\gls{pcm} & $62.92 \pm 2.63$ & $\mathbf{66.33 \pm 0.86}$ & $60.70 \pm 2.46$ \\
\gls{pcm} ($\times1.5$) & $50.12 \pm 7.01$ & $\mathbf{63.19 \pm 1.25}$ & $47.27 \pm 3.94$ \\
\rowcolor[gray]{0.9}
\gls{pcm} ($\times2$) & $33.63 \pm 5.03$ & $\mathbf{58.05 \pm 1.53}$ & $37.43 \pm 5.91$ \\
ReRAM & $59.73 \pm 1.91$ & $\mathbf{65.57 \pm 0.91}$ & $58.52 \pm 2.70$ \\
\bottomrule
\end{tabular}
\label{tab:noise-sensitivity}
\end{table}

\begin{table}[ht]
\caption{Detailed benchmark results for \textit{Phi-3-mini-4k-instruct}-based models evaluated with ReRAM noise model. Best results when hardware-realistic noise is applied are bold faced. Evaluations with noise injection are repeated for 10 different seeds per benchmark.}
\centering
\resizebox{\textwidth}{!}{%
\begin{tabular}{@{}l*{10}{c}@{}}
\toprule
\multirow{2}{*}{Model} & \multicolumn{9}{c}{Benchmarks} & \multirow{2}{*}{Avg.} \\
\cmidrule(lr){2-10}
 & \makecell{MMLU\\(5-shot)} & \makecell{GSM8K\\(CoT 8-shot)} & \makecell{BoolQ\\(0-shot)} & \makecell{Hellaswag\\(5-shot)} & \makecell{MedQA\\(2-shot)} & \makecell{AGIEval\\(0-shot)} & \makecell{Arc-C\\(10-shot)} & \makecell{Arc-E\\(10-shot)} & \makecell{ANLI\\(7-shot)} & \\
\midrule
\textit{Phi-3-mini-4k-instruct} (W16) & $69.36$ & $79.91$ & $78.75$ & $83.51$ & $52.83$ & $38.06$ & $84.56$ & $91.08$ & $52.25$ & $70.03$ \\
\rowcolor[gray]{0.9}
\textit{Phi-3-mini-4k-instruct} ($\text{W16}_\text{hw noise}$) & \parbox[t]{1.5cm}{\centering $61.80$\\$\pm 0.66$} & \parbox[t]{1.5cm}{\centering $49.65$\\$\pm 4.91$} & \parbox[t]{1.5cm}{\centering $71.79$\\$\pm 1.93$} & \parbox[t]{1.5cm}{\centering $72.06$\\$\pm 1.42$} & \parbox[t]{1.5cm}{\centering $43.33$\\$\pm 1.35$} & \parbox[t]{1.5cm}{\centering $32.31$\\$\pm 0.88$} & \parbox[t]{1.5cm}{\centering $79.19$\\$\pm 1.24$} & \parbox[t]{1.5cm}{\centering $88.13$\\$\pm 0.31$} & \parbox[t]{1.5cm}{\centering $39.35$\\$\pm 4.45$} & $59.73$ \\
Analog FM (SI8–W16–O8) & $67.24$ & $76.95$ & $77.22$ & $83.24$ & $49.61$ & $37.23$ & $83.62$ & $90.57$ & $51.94$ & $68.62$ \\
\rowcolor[gray]{0.9}
Analog FM (SI8–$\text{W16}_\text{hw noise}$–O8) & \parbox[t]{1.5cm}{\centering $\mathbf{64.61}$\\$\pm \mathbf{0.29}$} & \parbox[t]{1.5cm}{\centering $\mathbf{69.96}$\\$\pm \mathbf{0.88}$} & \parbox[t]{1.5cm}{\centering $\mathbf{76.33}$\\$\pm \mathbf{1.20}$} & \parbox[t]{1.5cm}{\centering $\mathbf{79.29}$\\$\pm \mathbf{0.97}$} & \parbox[t]{1.5cm}{\centering $\mathbf{45.73}$\\$\pm \mathbf{0.73}$} & \parbox[t]{1.5cm}{\centering $\mathbf{35.71}$\\$\pm \mathbf{0.47}$} & \parbox[t]{1.5cm}{\centering $\mathbf{82.23}$\\$\pm \mathbf{0.51}$} & \parbox[t]{1.5cm}{\centering $\mathbf{89.57}$\\$\pm \mathbf{0.23}$} & \parbox[t]{1.5cm}{\centering $\mathbf{46.73}$\\$\pm \mathbf{2.88}$} & $\mathbf{65.57}$ \\
LLM-QAT (SI8–W4) & $64.12$ & $68.92$ & $75.11$ & $79.22$ & $45.20$ & $36.30$ & $81.48$ & $89.18$ & $51.16$ & $65.63$ \\
\rowcolor[gray]{0.9}
LLM-QAT (SI8–$\text{W4}_\text{hw noise}$) & \parbox[t]{1.5cm}{\centering $59.00$\\$\pm 1.14$} & \parbox[t]{1.5cm}{\centering $53.40$\\$\pm 2.25$} & \parbox[t]{1.5cm}{\centering $70.60$\\$\pm 4.16$} & \parbox[t]{1.5cm}{\centering $72.01$\\$\pm 1.76$} & \parbox[t]{1.5cm}{\centering $38.31$\\$\pm 2.62$} & \parbox[t]{1.5cm}{\centering $32.91$\\$\pm 1.18$} & \parbox[t]{1.5cm}{\centering $76.75$\\$\pm 1.51$} & \parbox[t]{1.5cm}{\centering $86.26$\\$\pm 0.82$} & \parbox[t]{1.5cm}{\centering $37.46$\\$\pm 8.89$} & $58.52$ \\
\bottomrule
\end{tabular}
}
\label{tab:reram-detailed}
\end{table}

\subsection{Robustness to read noise and conductance drift} \label{appendix:drift}
Beyond programming noise, real \gls{aimc} hardware exhibits additional nonidealities including read noise and conductance drift. Read noise arises from temporal conductance fluctuations due to 1/f noise in \gls{nvm} devices, while conductance drift refers to the time-dependent decrease in device conductance observed in \gls{pcm} devices. To evaluate robustness to these additional noise sources, we employed a publicly available comprehensive noise model~\cite{aihwkit} that includes programming noise, conductance-dependent read noise, and time-dependent drift characteristics calibrated on real \gls{pcm} hardware.

We evaluated the \textit{Phi-3-mini-4k-instruct}-based models on the MedQA and Arc-C benchmarks while simulating drift over timespans ranging from one minute to one year after programming. Table \ref{tab:drift-combined} shows the results. The analog foundation model demonstrates significantly stronger robustness compared to both the off-the-shelf model and the LLM-QAT model across all time scales. For example, on MedQA after one year of drift, our analog foundation model maintains $33.74\%$ accuracy compared to $29.25\%$ for the off-the-shelf model and $23.30\%$ for the LLM-QAT model. Similar trends are observed on Arc-C, where after one year our model achieves $68.57\%$ compared to $58.89\%$ and $49.44\%$ for the baseline and LLM-QAT models, respectively. These results demonstrate that our training methodology provides robustness not only to programming noise, but also to the temporal variations that occur in deployed \gls{aimc} hardware.

\begin{table}[ht]
\caption{Robustness to conductance drift evaluated on \textit{Phi-3-mini-4k-instruct}-based models. Results shown for MedQA and Arc-C benchmarks at different time points after programming. The noise model includes programming noise, read noise, and time-dependent conductance drift. All evaluations are repeated for 10 different seeds.}
\centering
\resizebox{\textwidth}{!}{%
\begin{tabular}{@{}l*{12}{c}@{}}
\toprule
\multirow{2}{*}{Model} & \multicolumn{6}{c}{MedQA (2-shot)} & \multicolumn{6}{c}{Arc-C (10-shot)} \\
\cmidrule(lr){2-7} \cmidrule(lr){8-13}
 & FP16 & $t=1\text{min}$ & $t=1\text{h}$ & $t=1\text{d}$ & $t=1\text{m}$ & $t=1\text{y}$ & FP16 & $t=1\text{min}$ & $t=1\text{h}$ & $t=1\text{d}$ & $t=1\text{m}$ & $t=1\text{y}$ \\
\midrule
\parbox[t]{3.5cm}{\textit{Phi-3-mini-4k-instruct}\\($\text{W16}_\text{hw noise}$)} & $52.91$ & \parbox[t]{1.2cm}{\centering $43.54$\\$\pm 0.81$} & \parbox[t]{1.2cm}{\centering $43.65$\\$\pm 0.62$} & \parbox[t]{1.2cm}{\centering $40.75$\\$\pm 0.86$} & \parbox[t]{1.2cm}{\centering $35.82$\\$\pm 1.11$} & \parbox[t]{1.2cm}{\centering $29.25$\\$\pm 1.12$} & $84.56$ & \parbox[t]{1.2cm}{\centering $82.37$\\$\pm 0.61$} & \parbox[t]{1.2cm}{\centering $81.59$\\$\pm 0.37$} & \parbox[t]{1.2cm}{\centering $79.01$\\$\pm 0.41$} & \parbox[t]{1.2cm}{\centering $72.71$\\$\pm 0.77$} & \parbox[t]{1.2cm}{\centering $58.89$\\$\pm 3.75$} \\
\rowcolor[gray]{0.9}
\parbox[t]{3.5cm}{Analog FM\\(SI8–$\text{W16}_\text{hw noise}$–O8)} & $49.69$ & \parbox[t]{1.2cm}{\centering $\mathbf{45.85}$\\$\pm \mathbf{0.78}$} & \parbox[t]{1.2cm}{\centering $\mathbf{45.03}$\\$\pm \mathbf{0.75}$} & \parbox[t]{1.2cm}{\centering $\mathbf{42.09}$\\$\pm \mathbf{0.97}$} & \parbox[t]{1.2cm}{\centering $\mathbf{38.02}$\\$\pm \mathbf{1.23}$} & \parbox[t]{1.2cm}{\centering $\mathbf{33.74}$\\$\pm \mathbf{1.12}$} & $84.47$ & \parbox[t]{1.2cm}{\centering $\mathbf{83.43}$\\$\pm \mathbf{0.14}$} & \parbox[t]{1.2cm}{\centering $\mathbf{83.05}$\\$\pm \mathbf{0.25}$} & \parbox[t]{1.2cm}{\centering $\mathbf{81.50}$\\$\pm \mathbf{0.47}$} & \parbox[t]{1.2cm}{\centering $\mathbf{76.79}$\\$\pm \mathbf{1.15}$} & \parbox[t]{1.2cm}{\centering $\mathbf{68.57}$\\$\pm \mathbf{1.23}$} \\
\parbox[t]{3.5cm}{LLM-QAT\\(SI8–$\text{W4}_\text{hw noise}$)} & $46.15$ & \parbox[t]{1.2cm}{\centering $39.87$\\$\pm 1.00$} & \parbox[t]{1.2cm}{\centering $36.59$\\$\pm 1.21$} & \parbox[t]{1.2cm}{\centering $31.98$\\$\pm 0.69$} & \parbox[t]{1.2cm}{\centering $26.71$\\$\pm 0.35$} & \parbox[t]{1.2cm}{\centering $23.30$\\$\pm 0.56$} & $81.31$ & \parbox[t]{1.2cm}{\centering $78.34$\\$\pm 1.66$} & \parbox[t]{1.2cm}{\centering $75.72$\\$\pm 1.20$} & \parbox[t]{1.2cm}{\centering $69.71$\\$\pm 0.80$} & \parbox[t]{1.2cm}{\centering $58.98$\\$\pm 0.93$} & \parbox[t]{1.2cm}{\centering $49.44$\\$\pm 0.89$} \\
\bottomrule
\end{tabular}
}
\label{tab:drift-combined}
\end{table}

\subsection{Input quantization precision}
To investigate the impact of input quantization precision on model performance, we trained an additional analog foundation model using 7-bit input quantization instead of 8-bit on 1B synthetically generated tokens. Table \ref{tab:ablation-7bit-input} shows that reducing input precision from 8 bits to 7 bits results in an average performance drop of approximately 1.5\%, and additional 1.1\% when hardware-realistic noise is injected during evaluation. This demonstrates that our training methodology is robust to varying input quantization precisions, though 8-bit input quantization offers a better trade-off between hardware complexity and model accuracy.

\begin{table}[ht]
\caption{Ablation study on input quantization precision for \textit{Phi-3-mini-4k-instruct}. Models trained on 1B tokens with either 8-bit or 7-bit static input quantization. Evaluations with noise injection are repeated for 10 different seeds per benchmark.}
\centering
\begin{tabular}{@{}lc@{}}
\toprule
Model & Avg. \\
\midrule
Analog FM (SI8–W16–O8) & $66.74$ \\
\rowcolor[gray]{0.9}
Analog FM (SI8–$\text{W16}_\text{hw noise}$–O8) & $63.67$ \\
Analog FM (SI7–W16–O8) & $65.25$ \\
\rowcolor[gray]{0.9}
Analog FM (SI7–$\text{W16}_\text{hw noise}$–O8) & $62.57$ \\
\bottomrule
\end{tabular}
\label{tab:ablation-7bit-input}
\end{table}

\section{Impact of tiled AIMC architectures} \label{appendix:tiling}
Modern \gls{aimc} chips typically employ tiled architectures where weight matrices are partitioned into smaller tiles, each mapped to a separate crossbar array. While the weight matrices in \glspl{llm} can be as large as $8192 \times 8192$, individual \gls{aimc} tiles are commonly limited to sizes between $256 \times 256$ and $1024 \times 1024$ due to physical constraints~\cite{buechel-nat-comp,hermes}. When performing \glspl{mvm}, partial results from individual tiles are computed in the analog domain and then aggregated using high-precision digital circuitry. This tiled approach has implications for the overall noise characteristics of the system.

A key advantage of tiling is that weight-to-conductance mapping is performed independently per tile rather than per layer. This means that outliers in one tile do not affect the conductance scaling of weights in other tiles. As a result, more weights across the model can be mapped to higher conductance values, which generally exhibit better \gls{snr} in the \gls{pcm} noise model (see figure \ref{fig:weight-snr}a). However, the benefit of tiling depends strongly on the weight distribution: models with many outliers see significant improvements, while models with already-tight weight distributions see minimal changes.

To quantify this effect, we evaluated the \textit{Phi-3-mini-4k-instruct}-based models from table \ref{tab:model-comparison} using a tile size of $512 \times 512$. Table \ref{tab:tiled-results} shows the results. For the off-the-shelf model with hardware-realistic noise, tiling improves average accuracy from $62.92\%$ to $64.68\%$ ($+1.76\%$). The LLM-QAT model shows a similar improvement from $60.69\%$ to $61.80\%$ ($+1.11\%$). In contrast, the analog foundation model shows minimal change, decreasing from $66.33\%$ to $66.23\%$ ($-0.10\%$). This is expected because the analog foundation model already has tight weight distributions due to iterative clipping during training (see section \ref{appendix:ablations-hw-aware} and figure \ref{fig:weight-snr}), leaving little room for tiling to provide additional \gls{snr} improvements.

These results demonstrate that our training methodology produces models that are inherently well-suited for \gls{aimc} deployment, achieving near-optimal conductance mapping even without the benefits of fine-grained tiling. While tiling can partially compensate for suboptimal weight distributions in off-the-shelf and QAT-trained models, it does not eliminate the performance gap to our analog foundation models. Furthermore, our approach provides robustness improvements that are complementary to architectural choices such as tile size, making it applicable across different \gls{aimc} system designs.

\begin{table}[ht]
\caption{Impact of tiled \gls{aimc} architecture on \textit{Phi-3-mini-4k-instruct}-based models. Results shown for tile size $512 \times 512$. Weight-to-conductance mapping is performed per-tile rather than per-layer, which improves \gls{snr} for models with wider weight distributions. Best results when hardware-realistic noise is applied are bold faced. Evaluations with noise injection are repeated for 10 different seeds per benchmark.}
\centering
\resizebox{\textwidth}{!}{%
\begin{tabular}{@{}l*{10}{c}@{}}
\toprule
\multirow{2}{*}{Model} & \multicolumn{9}{c}{Benchmarks} & \multirow{2}{*}{Avg.} \\
\cmidrule(lr){2-10}
 & \makecell{MMLU\\(5-shot)} & \makecell{GSM8K\\(CoT 8-shot)} & \makecell{BoolQ\\(0-shot)} & \makecell{Hellaswag\\(5-shot)} & \makecell{MedQA\\(2-shot)} & \makecell{AGIEval\\(0-shot)} & \makecell{Arc-C\\(10-shot)} & \makecell{Arc-E\\(10-shot)} & \makecell{ANLI\\(7-shot)} & \\
\midrule
\textit{Phi-3-mini-4k-instruct} (W16) & $69.36$ & $79.91$ & $78.75$ & $83.51$ & $52.83$ & $38.06$ & $84.56$ & $91.08$ & $52.25$ & $70.03$ \\
\rowcolor[gray]{0.9}
\textit{Phi-3-mini-4k-instruct} ($\text{W16}_\text{hw noise}$) & \parbox[t]{1.5cm}{\centering $65.28$\\$\pm 0.50$} & \parbox[t]{1.5cm}{\centering $65.11$\\$\pm 4.59$} & \parbox[t]{1.5cm}{\centering $74.84$\\$\pm 1.40$} & \parbox[t]{1.5cm}{\centering $78.22$\\$\pm 2.00$} & \parbox[t]{1.5cm}{\centering $47.10$\\$\pm 0.98$} & \parbox[t]{1.5cm}{\centering $34.40$\\$\pm 0.92$} & \parbox[t]{1.5cm}{\centering $82.33$\\$\pm 0.97$} & \parbox[t]{1.5cm}{\centering $89.97$\\$\pm 0.46$} & \parbox[t]{1.5cm}{\centering $44.83$\\$\pm 2.07$} & $64.68$ \\
Analog FM (SI8–W16–O8) & $67.24$ & $76.95$ & $77.22$ & $83.24$ & $49.61$ & $37.23$ & $83.62$ & $90.57$ & $51.94$ & $68.62$ \\
\rowcolor[gray]{0.9}
Analog FM (SI8–$\text{W16}_\text{hw noise}$–O8) & \parbox[t]{1.5cm}{\centering $\mathbf{65.16}$\\$\pm \mathbf{0.24}$} & \parbox[t]{1.5cm}{\centering $\mathbf{70.72}$\\$\pm \mathbf{1.25}$} & \parbox[t]{1.5cm}{\centering $\mathbf{75.58}$\\$\pm \mathbf{1.50}$} & \parbox[t]{1.5cm}{\centering $\mathbf{80.53}$\\$\pm \mathbf{0.75}$} & \parbox[t]{1.5cm}{\centering $\mathbf{46.65}$\\$\pm \mathbf{1.18}$} & \parbox[t]{1.5cm}{\centering $\mathbf{35.68}$\\$\pm \mathbf{0.59}$} & \parbox[t]{1.5cm}{\centering $\mathbf{82.95}$\\$\pm \mathbf{0.45}$} & \parbox[t]{1.5cm}{\centering $\mathbf{89.87}$\\$\pm \mathbf{0.27}$} & \parbox[t]{1.5cm}{\centering $\mathbf{48.95}$\\$\pm \mathbf{2.33}$} & $\mathbf{66.23}$ \\
LLM-QAT (SI8–W4) & $64.12$ & $68.92$ & $75.11$ & $79.22$ & $45.20$ & $36.30$ & $81.48$ & $89.18$ & $51.16$ & $65.63$ \\
\rowcolor[gray]{0.9}
LLM-QAT (SI8–$\text{W4}_\text{hw noise}$) & \parbox[t]{1.5cm}{\centering $61.52$\\$\pm 0.46$} & \parbox[t]{1.5cm}{\centering $61.63$\\$\pm 0.75$} & \parbox[t]{1.5cm}{\centering $69.08$\\$\pm 6.02$} & \parbox[t]{1.5cm}{\centering $76.42$\\$\pm 1.01$} & \parbox[t]{1.5cm}{\centering $42.22$\\$\pm 0.97$} & \parbox[t]{1.5cm}{\centering $34.10$\\$\pm 0.75$} & \parbox[t]{1.5cm}{\centering $79.15$\\$\pm 0.83$} & \parbox[t]{1.5cm}{\centering $87.85$\\$\pm 0.65$} & \parbox[t]{1.5cm}{\centering $44.22$\\$\pm 4.27$} & $61.80$ \\
SpinQuant (DI8–W4) & $67.28$ & $74.83$ & $76.27$ & $81.66$ & $49.06$ & $36.45$ & $83.62$ & $90.45$ & $48.47$ & $67.55$ \\
\rowcolor[gray]{0.9}
SpinQuant (DI8–$\text{W4}_\text{hw noise}$) & \parbox[t]{1.5cm}{\centering $49.46$\\$\pm 1.87$} & \parbox[t]{1.5cm}{\centering $20.49$\\$\pm 6.69$} & \parbox[t]{1.5cm}{\centering $64.42$\\$\pm 2.71$} & \parbox[t]{1.5cm}{\centering $36.01$\\$\pm 3.78$} & \parbox[t]{1.5cm}{\centering $32.16$\\$\pm 1.21$} & \parbox[t]{1.5cm}{\centering $28.16$\\$\pm 1.28$} & \parbox[t]{1.5cm}{\centering $62.20$\\$\pm 4.30$} & \parbox[t]{1.5cm}{\centering $77.05$\\$\pm 3.26$} & \parbox[t]{1.5cm}{\centering $34.60$\\$\pm 1.29$} & $44.95$ \\
\bottomrule
\end{tabular}
}
\label{tab:tiled-results}
\end{table}

\section{Training convergence} \label{appendix:convergence}
\begin{figure}
    \centering
    \includegraphics[width=1.0\linewidth]{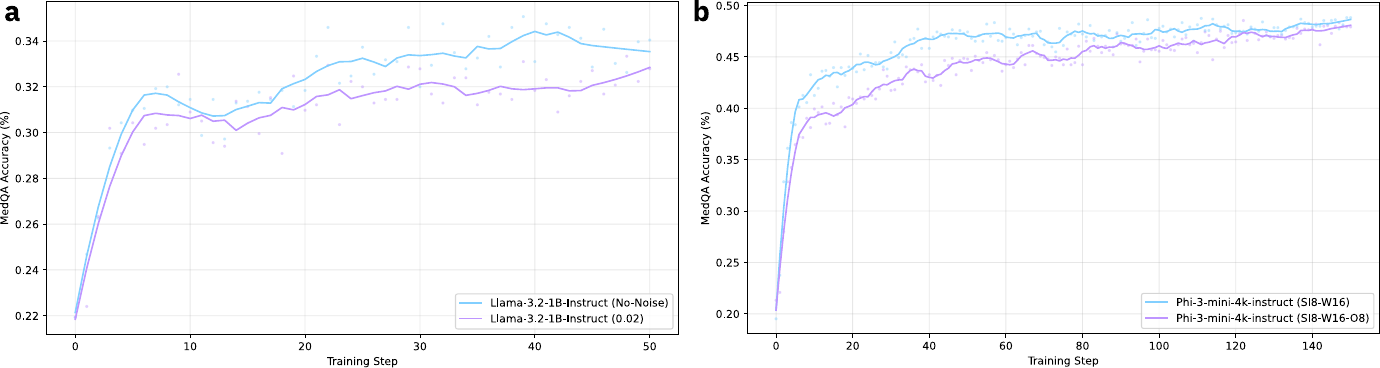}
    \caption{\textbf{a.} Impact of noise injection on convergence. \textbf{b.} Impact of output quantization on training convergence.}
    \label{fig:convergence}
\end{figure}

In order to increase the robustness of the weights of the model, Gaussian noise according to eq. \ref{eq:weight-noise} is injected into the weights during the forward pass. Additionally, to adapt the model to output quantization, which simulates the \glspl{adc}, we perform globally static output clipping and quantization on the \gls{mvm} outputs. Figure \ref{fig:convergence} shows that both methodologies lead to slower convergence. As a result, models need to be trained for more steps to reach the desired performance.

\section{Training details} \label{appendix:training-details}
\begin{figure}[ht]
    \centering
    \includegraphics[width=\textwidth]{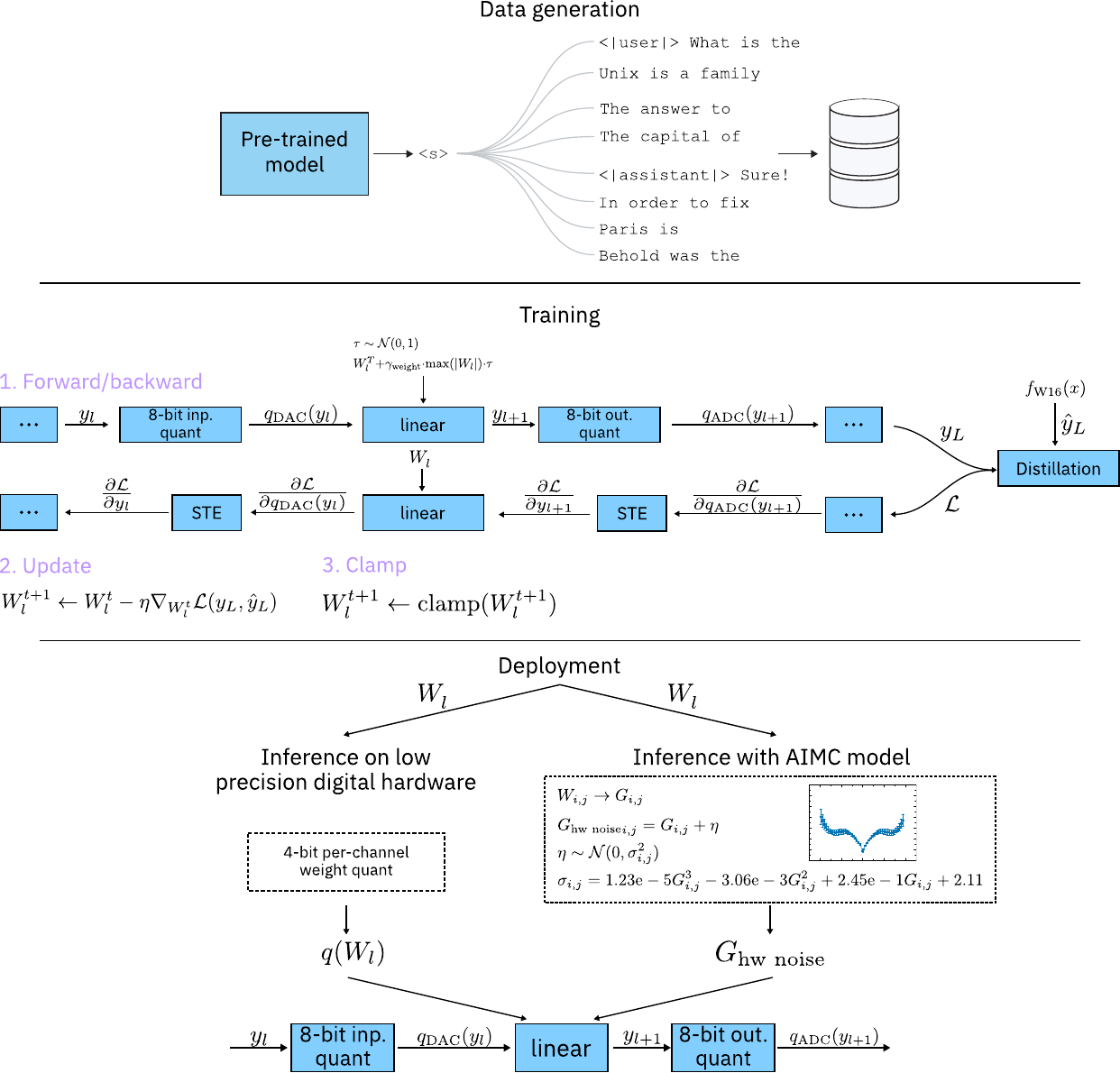}
    \caption{Illustration of pipeline for data generation, training, and deployment using quantization or hardware-realistic analog noise.}
    \label{fig:eval-pipeline}
\end{figure}
\paragraph{Generic training} During our training runs, we used the AdamW~\cite{adam} optimizer with $\beta_1=0.9$, $\beta_2=0.98$, and $\epsilon=1.0e-06$ for both \textit{Phi-3-mini-4k-instruct} and \textit{Llama-3.2-1B-Instruct} base models. We employed distillation (with beta=1.0) using a temperature of 2.0 for Phi-3 and 1.0 for Llama. Both models were trained for 2 epochs with a batch size of 96, polynomial learning rate scheduler, warmup ratio of 0.016, and a maximum gradient norm of 1.0. We applied gradient checkpointing and set weight decay to 0.01. The learning rates differed, with 1.0e-06 for Phi-3 and 5.0e-07 for Llama. The learning rate was multiplies by the number of GPUs used (96 for the main experiments).
\paragraph{Hardware-aware training} We used AIHWKIT-Lightning~\cite{aihwkit-lightning} for \gls{hwa} training. For both models, we enabled input range learning (\texttt{decay}=0.01, \texttt{input\_min\_percentage}=0.95) with \texttt{init\_value} of 3.0, though the \texttt{init\_std\_alpha} was 15.0 for Phi-3 and 18.0 for Llama. Interestingly, we found that during the initial $\sim500$ training steps, outliers need to be kept almost completely, which is ensured by calibrating the input ranges from data with 15 or 18 times the standard deviation of the activations. After $\sim500$ batches, input range learning takes over and input ranges start to tighten due to the gradients, but mostly because of the decay. We found this to be crucial for getting good performance with static input ranges. We used a additive Gaussian noise injection with magnitude \texttt{modifier.std\_dev} 0.02 for Phi-3 and 0.03 for Llama. Forward input (\texttt{inp\_res}) and output (\texttt{out\_res}) resolution was set to 254 for both models, with output bounds (\texttt{out\_bound}) of 12 for Phi-3 and 14 for Llama.

\section{Test-time compute scaling details}
For this experiment, we follow the implementation of \citet{ttcs-hf} (\url{https://github.com/huggingface/search-and-learn}). For the runs involving noisy weights, we sample 256 completions for every prompt in the MATH-500 dataset five times. This leads to 640000 generations per model. For every prompt, we sample a maximum of 2048 new tokens with a temperature of 0.8 and top-p of 1.0. The system prompt used is:

\textcolor{shotcolor}{Solve the following math problem efficiently and clearly:}

\textcolor{shotcolor}{- For simple problems (2 steps or fewer):\\Provide a concise solution with minimal explanation.}

\textcolor{shotcolor}{- For complex problems (3 steps or more):\\Use this step-by-step format:}

\textcolor{shotcolor}{\#\# Step 1: {[Concise description]}} \\
\textcolor{shotcolor}{{[Brief explanation and calculations]}}

\textcolor{shotcolor}{\#\# Step 2: {[Concise description]}} \\
\textcolor{shotcolor}{{[Brief explanation and calculations]}}

\textcolor{shotcolor}{...}

\textcolor{shotcolor}{Regardless of the approach, always conclude with:}

\textcolor{shotcolor}{Therefore, the final answer is: \$ \textbackslash boxed\{answer\}\$. I hope it is correct.}

\textcolor{shotcolor}{Where {[answer]} is just the final number or expression that solves the problem.}

For \textit{Phi-3-mini-4k-instruct}, we moved the system prompt into the user prompt separated by a newline. Also, the \texttt{<|end|>} token needs to be included in the list of tokens that stop generation.

\begin{table}[ht]
\centering
\caption{Full results behind the test-time compute scaling results presented in the main text of the paper for \textit{Phi-3-mini-4k-instruct} and \textit{Phi-3-mini-4k-instruct}. PRM (greedy) indicates picking the answer with the highest reward. PRM (voting) indicates weighting the answers with their respective rewards and picking the answer with the highest cumulative reward. Voting refers to simple majority voting.}
\resizebox{\textwidth}{!}{%
\begin{tabular}{@{}l*{9}{c}@{}}
\toprule
\multirow{2}{*}{Method} & \multicolumn{9}{c}{Number of samples} \\
\cmidrule(lr){2-10}
 & $n=1$ & $n=2$ & $n=4$ & $n=8$ & $n=16$ & $n=32$ & $n=64$ & $n=128$ & $n=256$ \\
\midrule
\multicolumn{10}{@{}l}{\textbf{Phi-3-mini-4k-instruct (SI8–$\text{W16}$)}} \\
\midrule
PRM (greedy) & $35.24 \pm 1.59$ & $41.32 \pm 0.94$ & $44.80 \pm 1.90$ & $47.80 \pm 1.71$ & $49.36 \pm 1.89$ & $50.60 \pm 0.68$ & $52.44 \pm 0.61$ & $52.76 \pm 1.14$ & $53.80 \pm 0.76$ \\
PRM (voting) & $35.24 \pm 1.59$ & $41.36 \pm 0.89$ & $46.56 \pm 1.54$ & $50.48 \pm 1.11$ & $52.68 \pm 0.93$ & $54.68 \pm 0.84$ & $55.60 \pm 0.78$ & $56.40 \pm 0.13$ & $\mathbf{56.80} \pm 0.66$ \\
Voting & $35.24 \pm 1.59$ & $35.24 \pm 1.59$ & $42.76 \pm 1.25$ & $47.88 \pm 0.93$ & $50.12 \pm 0.83$ & $52.64 \pm 0.48$ & $53.08 \pm 0.65$ & $53.08 \pm 0.30$ & $53.44 \pm 0.46$ \\
\midrule
\multicolumn{10}{@{}l}{\textbf{Phi-3-mini-4k-instruct (SI8–$\text{W16}_\text{hw noise}$)}} \\
\midrule
PRM (greedy) & $18.28 \pm 2.10$ & $22.40 \pm 4.16$ & $26.36 \pm 4.74$ & $30.68 \pm 4.81$ & $33.60 \pm 5.87$ & $34.36 \pm 5.00$ & $36.84 \pm 3.59$ & $38.16 \pm 3.25$ & $39.04 \pm 2.32$ \\
PRM (voting) & $18.28 \pm 2.10$ & $22.40 \pm 4.16$ & $27.40 \pm 4.84$ & $33.00 \pm 4.15$ & $35.76 \pm 4.39$ & $38.52 \pm 4.74$ & $40.16 \pm 4.12$ & $40.72 \pm 3.41$ & $\mathbf{41.40} \pm 3.21$ \\
Voting & $18.28 \pm 2.10$ & $18.28 \pm 2.10$ & $24.16 \pm 3.66$ & $30.56 \pm 4.69$ & $33.92 \pm 4.89$ & $36.16 \pm 4.53$ & $38.56 \pm 4.00$ & $39.04 \pm 3.36$ & $39.04 \pm 3.30$ \\
\midrule
\multicolumn{10}{@{}l}{\textbf{Analog FM (SI8–$\text{W16}$–O8)}} \\
\midrule
PRM (greedy) & $33.40 \pm 1.52$ & $39.76 \pm 1.25$ & $43.72 \pm 1.35$ & $46.84 \pm 1.01$ & $47.72 \pm 0.20$ & $47.64 \pm 0.61$ & $48.32 \pm 1.24$ & $49.40 \pm 0.59$ & $49.52 \pm 1.36$ \\
PRM (voting) & $33.40 \pm 1.52$ & $39.76 \pm 1.29$ & $45.08 \pm 1.28$ & $49.20 \pm 0.75$ & $51.20 \pm 0.58$ & $52.68 \pm 1.25$ & $53.00 \pm 0.75$ & $53.20 \pm 0.87$ & $\mathbf{54.12} \pm 0.52$ \\
Voting & $33.40 \pm 1.52$ & $33.40 \pm 1.52$ & $40.08 \pm 1.56$ & $45.48 \pm 1.06$ & $48.84 \pm 1.13$ & $50.84 \pm 1.44$ & $51.16 \pm 0.83$ & $52.12 \pm 1.11$ & $51.88 \pm 0.68$ \\
\midrule
\multicolumn{10}{@{}l}{\textbf{Analog FM (SI8–$\text{W16}_\text{hw noise}$–O8)}} \\
\midrule
PRM (greedy) & $27.48 \pm 1.38$ & $31.84 \pm 0.87$ & $36.88 \pm 1.34$ & $39.64 \pm 0.97$ & $42.28 \pm 1.16$ & $43.56 \pm 1.55$ & $45.40 \pm 0.55$ & $46.60 \pm 0.99$ & $47.16 \pm 1.58$ \\
PRM (voting) & $27.48 \pm 1.38$ & $31.88 \pm 0.83$ & $38.28 \pm 1.48$ & $42.68 \pm 1.10$ & $45.28 \pm 0.48$ & $47.12 \pm 1.28$ & $48.24 \pm 0.75$ & $49.08 \pm 1.47$ & $\mathbf{49.48} \pm 0.98$ \\
Voting & $27.48 \pm 1.38$ & $27.48 \pm 1.38$ & $33.84 \pm 1.68$ & $40.00 \pm 0.72$ & $42.84 \pm 0.62$ & $44.32 \pm 1.40$ & $45.96 \pm 0.73$ & $46.52 \pm 1.17$ & $47.12 \pm 1.14$ \\
\midrule
\multicolumn{10}{@{}l}{\textbf{LLM-QAT (SI8–$\text{W4}$)}} \\
\midrule
PRM (greedy) & $28.28 \pm 0.78$ & $33.20 \pm 1.06$ & $37.16 \pm 1.26$ & $40.12 \pm 1.60$ & $42.68 \pm 1.95$ & $44.04 \pm 1.90$ & $45.00 \pm 0.89$ & $45.84 \pm 0.72$ & $46.72 \pm 0.84$ \\
PRM (voting) & $28.28 \pm 0.78$ & $33.16 \pm 1.05$ & $38.36 \pm 1.26$ & $41.76 \pm 1.54$ & $44.08 \pm 1.47$ & $46.20 \pm 1.31$ & $47.12 \pm 1.20$ & $47.24 \pm 0.50$ & $\mathbf{47.68} \pm 0.73$ \\
Voting & $28.28 \pm 0.78$ & $28.28 \pm 0.78$ & $34.08 \pm 1.00$ & $39.28 \pm 0.71$ & $42.08 \pm 0.97$ & $44.76 \pm 0.81$ & $46.20 \pm 0.87$ & $46.12 \pm 0.84$ & $46.48 \pm 0.72$ \\
\midrule
\multicolumn{10}{@{}l}{\textbf{LLM-QAT (SI8–$\text{W4}_\text{hw noise}$)}} \\
\midrule
PRM (greedy) & $22.01 \pm 3.69$ & $26.21 \pm 3.93$ & $29.69 \pm 4.62$ & $33.09 \pm 4.75$ & $35.41 \pm 3.14$ & $37.90 \pm 3.16$ & $39.50 \pm 2.84$ & $39.58 \pm 2.74$ & $40.70 \pm 2.61$ \\
PRM (voting) & $22.01 \pm 3.69$ & $26.17 \pm 3.91$ & $30.25 \pm 4.44$ & $34.62 \pm 4.38$ & $37.10 \pm 2.20$ & $39.46 \pm 3.45$ & $41.78 \pm 3.46$ & $\mathbf{42.26} \pm 2.04$ & $41.98 \pm 2.08$ \\
Voting & $22.01 \pm 3.69$ & $22.01 \pm 3.69$ & $26.41 \pm 4.20$ & $31.97 \pm 3.71$ & $35.85 \pm 3.45$ & $37.82 \pm 3.61$ & $39.98 \pm 3.18$ & $40.42 \pm 1.61$ & $40.50 \pm 1.62$ \\
\midrule
\midrule
\multicolumn{10}{@{}l}{\textbf{Llama-3.2-1B-Instruct (SI8–$\text{W16}$)}} \\
\midrule
PRM (greedy) & $25.86 \pm 1.11$ & $32.63 \pm 0.87$ & $37.15 \pm 1.36$ & $40.15 \pm 0.88$ & $42.67 \pm 1.15$ & $44.00 \pm 1.39$ & $44.36 \pm 1.56$ & $45.00 \pm 1.09$ & $45.52 \pm 0.60$ \\
PRM (voting) & $25.86 \pm 1.11$ & $32.63 \pm 0.87$ & $37.35 \pm 1.70$ & $41.59 \pm 1.51$ & $43.75 \pm 1.42$ & $45.32 \pm 0.85$ & $46.04 \pm 0.55$ & $\mathbf{46.88} \pm 0.81$ & $46.60 \pm 0.26$ \\
Voting & $25.86 \pm 1.11$ & $25.86 \pm 1.11$ & $32.07 \pm 1.09$ & $37.11 \pm 0.99$ & $40.23 \pm 1.31$ & $41.79 \pm 1.05$ & $42.39 \pm 0.67$ & $43.15 \pm 1.37$ & $43.47 \pm 0.85$ \\
\midrule
\multicolumn{10}{@{}l}{\textbf{Llama-3.2-1B-Instruct (SI8–$\text{W16}_\text{hw noise}$)}} \\
\midrule
PRM (greedy) & $8.05 \pm 0.83$ & $11.57 \pm 1.06$ & $15.45 \pm 1.42$ & $19.70 \pm 1.21$ & $22.54 \pm 1.66$ & $24.86 \pm 1.39$ & $26.54 \pm 1.19$ & $28.79 \pm 1.75$ & $\mathbf{30.39} \pm 2.11$ \\
PRM (voting) & $8.05 \pm 0.83$ & $11.53 \pm 1.02$ & $14.65 \pm 1.25$ & $17.70 \pm 1.36$ & $19.30 \pm 1.15$ & $19.74 \pm 1.93$ & $20.66 \pm 1.52$ & $21.30 \pm 2.13$ & $21.62 \pm 1.94$ \\
Voting & $8.05 \pm 0.83$ & $8.05 \pm 0.83$ & $10.25 \pm 0.66$ & $14.53 \pm 1.06$ & $15.61 \pm 0.96$ & $16.49 \pm 1.51$ & $17.17 \pm 1.40$ & $17.42 \pm 2.03$ & $17.74 \pm 1.29$ \\
\midrule
\multicolumn{10}{@{}l}{\textbf{Analog FM (SI8–$\text{W16}$–O8)}} \\
\midrule
PRM (greedy) & $19.12 \pm 0.83$ & $24.84 \pm 0.83$ & $30.80 \pm 1.44$ & $34.20 \pm 1.53$ & $36.64 \pm 1.86$ & $39.52 \pm 1.03$ & $41.60 \pm 1.56$ & $42.20 \pm 2.05$ & $\mathbf{42.84} \pm 1.30$ \\
PRM (voting) & $19.12 \pm 0.83$ & $24.84 \pm 0.83$ & $30.32 \pm 1.26$ & $33.88 \pm 0.84$ & $35.48 \pm 0.88$ & $37.44 \pm 1.16$ & $38.36 \pm 0.89$ & $38.72 \pm 0.92$ & $38.88 \pm 0.53$ \\
Voting & $19.12 \pm 0.83$ & $19.12 \pm 0.83$ & $22.88 \pm 1.25$ & $29.48 \pm 1.05$ & $32.32 \pm 0.86$ & $33.92 \pm 1.70$ & $34.32 \pm 0.33$ & $35.48 \pm 0.30$ & $35.52 \pm 0.84$ \\
\midrule
\multicolumn{10}{@{}l}{\textbf{Analog FM (SI8–$\text{W16}_\text{hw noise}$–O8)}} \\
\midrule
PRM (greedy) & $14.74 \pm 0.61$ & $19.87 \pm 0.77$ & $25.00 \pm 0.39$ & $28.29 \pm 1.29$ & $31.69 \pm 1.19$ & $33.78 \pm 1.46$ & $35.58 \pm 1.09$ & $37.42 \pm 0.99$ & $\mathbf{39.06} \pm 1.61$ \\
PRM (voting) & $14.74 \pm 0.61$ & $19.87 \pm 0.77$ & $24.12 \pm 0.45$ & $27.05 \pm 0.96$ & $30.21 \pm 1.32$ & $32.25 \pm 1.20$ & $33.01 \pm 1.11$ & $33.37 \pm 1.02$ & $33.89 \pm 0.91$ \\
Voting & $14.74 \pm 0.61$ & $14.74 \pm 0.61$ & $18.55 \pm 0.69$ & $22.72 \pm 0.43$ & $26.84 \pm 1.95$ & $28.32 \pm 1.22$ & $29.61 \pm 1.12$ & $30.05 \pm 0.67$ & $30.61 \pm 0.81$ \\
\midrule
\multicolumn{10}{@{}l}{\textbf{LLM-QAT (SI8–$\text{W4}$)}} \\
\midrule
PRM (greedy) & $13.40 \pm 1.15$ & $17.20 \pm 1.00$ & $20.36 \pm 1.08$ & $25.40 \pm 0.63$ & $29.32 \pm 0.35$ & $31.32 \pm 1.37$ & $32.80 \pm 1.34$ & $33.64 \pm 0.82$ & $\mathbf{34.32} \pm 0.95$ \\
PRM (voting) & $13.40 \pm 1.15$ & $17.20 \pm 1.00$ & $19.56 \pm 1.32$ & $23.60 \pm 1.11$ & $26.64 \pm 0.64$ & $27.88 \pm 0.55$ & $29.40 \pm 0.92$ & $30.80 \pm 1.21$ & $30.64 \pm 0.61$ \\
Voting & $13.40 \pm 1.15$ & $13.40 \pm 1.15$ & $15.92 \pm 1.36$ & $19.68 \pm 0.64$ & $21.88 \pm 0.95$ & $24.08 \pm 0.47$ & $26.44 \pm 0.89$ & $27.44 \pm 0.57$ & $27.80 \pm 0.46$ \\
\midrule
\multicolumn{10}{@{}l}{\textbf{LLM-QAT (SI8–$\text{W4}_\text{hw noise}$)}} \\
\midrule
PRM (greedy) & $9.36 \pm 0.91$ & $11.76 \pm 0.98$ & $15.60 \pm 1.12$ & $18.52 \pm 0.93$ & $22.12 \pm 0.83$ & $24.96 \pm 0.99$ & $26.36 \pm 0.66$ & $27.96 \pm 1.28$ & $\mathbf{28.60} \pm 0.63$ \\
PRM (voting) & $9.36 \pm 0.91$ & $11.76 \pm 0.98$ & $14.80 \pm 1.23$ & $17.72 \pm 0.95$ & $19.52 \pm 0.83$ & $21.04 \pm 1.06$ & $21.72 \pm 0.79$ & $22.08 \pm 0.65$ & $22.28 \pm 0.78$ \\
Voting & $9.36 \pm 0.91$ & $9.36 \pm 0.91$ & $11.16 \pm 1.50$ & $14.32 \pm 1.39$ & $15.84 \pm 0.56$ & $17.68 \pm 0.99$ & $18.68 \pm 0.70$ & $19.20 \pm 0.79$ & $19.04 \pm 1.17$ \\
\bottomrule
\end{tabular}
}
\label{tab:combined-results}
\end{table}

\section{Hardware throughput and energy efficiency estimation}
While deployment on physical \gls{aimc} hardware is beyond the scope of this work, we provide estimates of throughput and energy efficiency using an open-source simulator for \gls{aimc}-based architectures~\cite{buechel-nat-comp}. We simulated pipelined inference of the \textit{Phi-3-mini-4k-instruct} model with a pre-fill phase of 256 tokens, generation of 64 tokens, and a batch size of 4. Due to the Python-based implementation of the simulator, longer sequences become computationally prohibitive to simulate.

Under these conditions, \textit{Phi-3-mini-4k-instruct} achieves a throughput of 2554.43 tokens/s and an energy efficiency of 199.78 tokens/s/W on the simulated \gls{aimc} architecture. To contextualize this result, we compare against a hypothetical scenario where a GPU could execute all INT8 operations involved in model inference in a single pass at 100\% utilization. Even in this idealized comparison, the \gls{aimc}-based system demonstrates $3.6\times$ higher energy efficiency.

However, this comparison does not fully capture the potential advantages of \gls{aimc} systems. The high density of \gls{nvm} enables \gls{aimc}-based accelerators to host billions of parameters entirely on-chip, eliminating the energy-intensive data movement between off-chip memory and processing units that dominates energy consumption in conventional architectures. This advantage becomes particularly pronounced for models with tens to hundreds of billions of parameters, especially sparse architectures such as \glspl{moe}~\cite{buechel-nat-comp}. For such large-scale models, the energy savings from eliminating off-chip memory access can reach multiple orders of magnitude, while also enabling deployment scenarios that would be impractical with conventional hardware due to memory capacity constraints.

\newpage
\section*{NeurIPS Paper Checklist}

\begin{enumerate}
\item {\bf Claims}
    \item[] Question: Do the main claims made in the abstract and introduction accurately reflect the paper's contributions and scope?
    \item[] Answer: \answerYes{} 
    \item[] Justification: In the abstract and introduction, we make the following claims. We claim that off-the-shelf LLMs are not able to achieve 4-bit-level performance when deployed on AIMC-based hardware. We demonstrate this in table 1. We claim that our models achieve performance similar to models trained with 4-bit weight and 8-bit static input quantization. This is also demonstrated in table 1. We claim that our models can also be quantized post-training, obtaining competitive performance compared to models trained with state-of-the-art quantization algorithms. This is demonstrated in table 3. We further claim that our models show better scaling behavior under test-time compute scaling than models trained with 4-bit per-channel weight and 8-bit static input quantization. This is shown in figure 4. With respect to related works, we claim that simple STE is sufficient to obtain robustness against static output quantization. This is demonstrated by the results in table 1, and the small drop our model has compared to the off-the-shelf model evaluated in FP16.

\item {\bf Limitations}
    \item[] Question: Does the paper discuss the limitations of the work performed by the authors?
    \item[] Answer: \answerYes{} 
    \item[] Justification: We discuss the limitations of the work in "Conclusion and Limitations". Our main limitation is the resource intensity of our approach. We acknowledge that this might limit applicability for some researchers and stress the need for more resource-efficient ways of creating analog foundation models. We propose ideas addressing this issue that can be pursued. We also acknowledge the fact that there is still a gap in accuracy that needs to be addressed through further innovation.

\item {\bf Theory assumptions and proofs}
    \item[] Question: For each theoretical result, does the paper provide the full set of assumptions and a complete (and correct) proof?
    \item[] Answer: \answerNA{} 
    \item[] Justification: Our paper is purely empirical.

    \item {\bf Experimental result reproducibility}
    \item[] Question: Does the paper fully disclose all the information needed to reproduce the main experimental results of the paper to the extent that it affects the main claims and/or conclusions of the paper (regardless of whether the code and data are provided or not)?
    \item[] Answer: \answerYes{} 
    \item[] Justification: In the methods, we describe in detail how we train our analog foundation models. In the appendix, we provide details on the hyperparameters used for training our models and the evaluation we performed, including the exact examples used for few-shot learning, prompt structure, and answer extraction method.

\item {\bf Open access to data and code}
    \item[] Question: Does the paper provide open access to the data and code, with sufficient instructions to faithfully reproduce the main experimental results, as described in supplemental material?
    \item[] Answer: \answerYes{} 
    \item[] Justification: We provide code to reproduce the data-generation and training of our analog foundation models.

\item {\bf Experimental setting/details}
    \item[] Question: Does the paper specify all the training and test details (e.g., data splits, hyperparameters, how they were chosen, type of optimizer, etc.) necessary to understand the results?
    \item[] Answer: \answerYes{} 
    \item[] Justification: All training hyperparameters and evaluation details can be found in the appendix.

\item {\bf Experiment statistical significance}
    \item[] Question: Does the paper report error bars suitably and correctly defined or other appropriate information about the statistical significance of the experiments?
    \item[] Answer: \answerYes{} 
    \item[] Justification: We report error bars for every result involving sampling. Generally, every experiment involving sampling/noise was repeated 10 times and the mean result and standard deviation is reported.

\item {\bf Experiments compute resources}
    \item[] Question: For each experiment, does the paper provide sufficient information on the computer resources (type of compute workers, memory, time of execution) needed to reproduce the experiments?
    \item[] Answer: \answerYes{} 
    \item[] Justification: For the training of our models, we provide detailed resource requirements for training the models, including total training time on different GPUs.
    
\item {\bf Code of ethics}
    \item[] Question: Does the research conducted in the paper conform, in every respect, with the NeurIPS Code of Ethics \url{https://neurips.cc/public/EthicsGuidelines}?
    \item[] Answer: \answerYes{} 
    \item[] Justification: We made sure to preserve anonymity and conform with the NeurIPS Code of Ethics.

\item {\bf Broader impacts}
    \item[] Question: Does the paper discuss both potential positive societal impacts and negative societal impacts of the work performed?
    \item[] Answer: \answerYes{} 
    \item[] Justification: We discuss that our paper answers an important question in the field of Analog In-Memory Computing, and therefore also in the field of AI accelerators in general. We hope that this inspires new research and motivates further scaling of AIMC-based chips. On a more negative side, we discuss that we based our models on aligned pre-trained models, and that, although we show that our models do not lose the ability to prevent harmful content generation, we still know that these \glspl{llm} can produce harmful and toxic content. This is discussed in the Conclusion and Limitations section of the paper.
    
\item {\bf Safeguards}
    \item[] Question: Does the paper describe safeguards that have been put in place for responsible release of data or models that have a high risk for misuse (e.g., pretrained language models, image generators, or scraped datasets)?
    \item[] Answer: \answerYes{} 
    \item[] Justification: Our models are based on pre-trained models that were aligned post-training. As we show, our models do not lose the ability to prevent generating harmful content or answering to a harmful prompt.

\item {\bf Licenses for existing assets}
    \item[] Question: Are the creators or original owners of assets (e.g., code, data, models), used in the paper, properly credited and are the license and terms of use explicitly mentioned and properly respected?
    \item[] Answer: \answerYes{} 
    \item[] Justification: We build our models on top of the Phi-3 and Llama 3 series, which we credit. Our dataset is synthetically generated without any starting prompts, so does not warrant credit.

\item {\bf New assets}
    \item[] Question: Are new assets introduced in the paper well documented and is the documentation provided alongside the assets?
    \item[] Answer: \answerYes{} 
    \item[] Justification: Code is released under the MIT License. The code is well documented. Apart from the code, no new asset is released.

\item {\bf Crowdsourcing and research with human subjects}
    \item[] Question: For crowdsourcing experiments and research with human subjects, does the paper include the full text of instructions given to participants and screenshots, if applicable, as well as details about compensation (if any)? 
    \item[] Answer: \answerNA{} 
    \item[] Justification: No crowdsourcing or research with human subjects was performed.

\item {\bf Institutional review board (IRB) approvals or equivalent for research with human subjects}
    \item[] Question: Does the paper describe potential risks incurred by study participants, whether such risks were disclosed to the subjects, and whether Institutional Review Board (IRB) approvals (or an equivalent approval/review based on the requirements of your country or institution) were obtained?
    \item[] Answer: \answerNA{} 
    \item[] Justification: No research with human subjects was performed.

\item {\bf Declaration of LLM usage}
    \item[] Question: Does the paper describe the usage of LLMs if it is an important, original, or non-standard component of the core methods in this research? Note that if the LLM is used only for writing, editing, or formatting purposes and does not impact the core methodology, scientific rigorousness, or originality of the research, declaration is not required.
    \item[] Answer: \answerNA{} 
    \item[] Justification: LLMs were not an important, original, or non-standard component of the core methods in this research.

\end{enumerate}
\end{document}